\documentclass[preprint,12pt]{elsarticle}

\usepackage{amssymb}
\usepackage{amsmath}
\newtheorem{proposition}{Proposition}
\newproof{pf}{Proof}
\providecommand{\qed}{\hfill$\square$}
\usepackage{hyperref}
\usepackage{url}
\usepackage{xurl}
\usepackage{graphicx}
\usepackage{subcaption}
\usepackage[section]{placeins}
\usepackage{booktabs}
\usepackage{siunitx}
\usepackage{xcolor}
\definecolor{red}{rgb}{0,0,0} % clean version: render all revision marks in black
\hypersetup{hidelinks} % clean version: no link borders

\DeclareCaptionFont{revred}{\color{red}}
\newcounter{subsubfigure}
\counterwithin*{subsubfigure}{subfigure}
\renewcommand\thesubsubfigure{\roman{subsubfigure}}
\newenvironment{subsubfigure}[2][b]{%
  \refstepcounter{subsubfigure}%
  \begin{minipage}[#1]{#2}\centering
  \renewcommand\caption[1]{\par\vspace{4pt}{\centering\footnotesize(\thesubsubfigure)~##1\par}}%
}{\end{minipage}}
\newcommand{\rev}[1]{{\color{red}#1}}
\newenvironment{revblock}{\color{red}}{}

\journal{Robotics and Autonomous Systems}

\begin{document}

\begin{frontmatter}

%% Title, authors and addresses

\title{DWPP: Dynamic Window Pure Pursuit Considering Velocity and Acceleration Constraints}

\author[1]{Fumiya Ohnishi\corref{cor1}}
\ead{fumiya-onishi@keio.jp}

\author[2]{Masaki Takahashi}
\ead{takahashi@sd.keio.ac.jp}

\cortext[cor1]{Corresponding author}

\affiliation[1]{organization={School of Science for Open and Environmental Systems, Graduate School of Science and Technology, Keio University},
            addressline={3-14-1 Hiyoshi}, city={Kohoku-ku}, state={Yokohama}, country={Japan}}

\affiliation[2]{organization={Department of System Design Engineering, Keio University},
            addressline={3-14-1 Hiyoshi}, city={Kohoku-ku}, state={Yokohama}, country={Japan}}

\begin{abstract}
Pure pursuit and its variants are widely used for mobile robot path tracking owing to their simplicity and computational efficiency. However, many conventional approaches do not explicitly account for velocity and acceleration constraints, resulting in discrepancies between commanded and actual velocities that result in overshoot and degraded tracking performance. To address this problem, this paper proposes dynamic window pure pursuit (DWPP), which fundamentally reformulates the command velocity computation process to explicitly incorporate velocity and acceleration constraints. Specifically, DWPP formulates command velocity computation in the velocity space (the $v$–$\omega$ plane) and selects the command velocity as the point within the dynamic window that is closest to the line $\omega = \kappa v$. Experimental results demonstrate that DWPP avoids constraint-violating commands and achieves superior path-tracking accuracy compared with conventional pure pursuit methods. The proposed method has been integrated into the official Nav2 repository and is publicly available (\url{https://github.com/ros-navigation/navigation2}).
\end{abstract}

%Graphical abstract
\begin{graphicalabstract}
\includegraphics[width=1.0\linewidth]{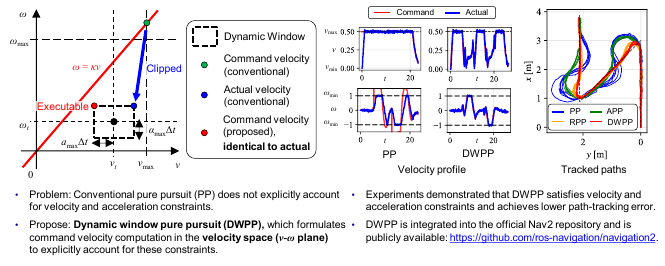}
\end{graphicalabstract}

% Research highlights
\begin{highlights}
\item A novel pure pursuit variant that fundamentally reformulates command velocity computation by explicitly incorporating velocity and acceleration constraints.
\item The core concept of DWPP is to formulate command velocity computation in the velocity space ($v$--$\omega$ plane) and to select the command velocity as the point within the dynamic window closest to the line $\omega = \kappa v$.
\item The performance of DWPP is evaluated through real-robot experiments compared with conventional approaches, demonstrating compliance with velocity and acceleration constraints as well as reduced path-tracking error.
\item The proposed method is integrated into the official Nav2 repository and is publicly available: \url{https://github.com/ros-navigation/navigation2}.
\end{highlights}


\end{frontmatter}

%% \linenumbers

%% main text

\section{Introduction}
\label{sec:intro}

Recently, the demand for automated guided vehicles (AGVs) in environments such as factories and hospitals has been increasing \cite{10091098}. Navigation for these robots typically starts with a global planner, which computes a path on a map to reach the goal. Subsequently, a local planner generates local paths and computes command velocities to track the global path while accounting for the robot's current state and real-time sensor data.

{\sloppy Representative local planners include the dynamic window approach (DWA) \cite{fox2002dynamic}, model predictive control (MPC) \cite{rosmann2015timed}, and pure pursuit (PP) \cite{coulter1992implementation}. Among these planners, DWA and MPC address multi-objective problems such as simultaneous path tracking and obstacle avoidance, whereas PP focuses on a single objective, namely path tracking. Nevertheless, PP remains widely used owing to its simplicity and  computational efficiency, and numerous improvements have continued to be proposed \cite{ahn2021accurate, li2023adaptive, macenski2023regulated, han2025hybrid}.\par}

One of these improved methods, regulated pure pursuit (RPP) \cite{macenski2023regulated}, is the standard implementation in Nav2 \cite{macenski2020marathon}, an autonomous navigation framework for the open-source robotics middleware Robot Operating System~2 (ROS~2) \cite{macenski2022robot}. However, a common limitation of existing PP–based methods, including RPP, is that they cannot explicitly account for a robot’s velocity and acceleration constraints during command velocity computation. Consequently, in conventional PP variants, either an additional layer is introduced to clip the command velocity according to velocity and acceleration limits or the raw command is sent directly to the robot and the realized velocity is clipped by the robot’s hardware constraints. In such cases, a discrepancy occurs between the motion planned by PP and the actual motion executed by the robot, which not only complicates the discussion of controller stability but also can result in overshoot from the reference path owing to the inability to fully realize the planned velocity.

To address these problems, this paper proposes dynamic window pure pursuit (DWPP), which fundamentally reformulates command velocity computation by explicitly embedding velocity and acceleration constraints. The core concept of DWPP is to formulate command velocity computation in the velocity space, namely the $v$--$\omega$ plane, where $v$ and $\omega$ denote the linear and angular velocities, respectively. Specifically, DWPP selects the command velocity as the point within the dynamic window—representing velocity and acceleration constraints in the $v$--$\omega$ plane—that is closest to the line $\omega = \kappa v$, which corresponds to the velocity condition derived from PP. This strategy enables the computation of velocity commands that \rev{best satisfy the path-tracking condition of PP} under the velocity and acceleration constraints.
% \rev{Because DWPP changes only how the command velocity is computed, its computational complexity remains $O(1)$ as in conventional PP variants, and it inherits their extensions such as the adaptive lookahead distance and the regulation heuristics.}
Robot experiments demonstrated that DWPP consistently respects velocity and acceleration limits while achieving superior path-following performance compared with existing PP variants implemented in Nav2. Furthermore, as a practical contribution, the DWPP implementation has been integrated into the Nav2 framework and is publicly available.~\footnote{\url{https://github.com/ros-navigation/navigation2}}

\rev{The remainder of this paper is organized as follows. Section \ref{sec:related_work} formulates the path-tracking problem addressed in this paper and reviews conventional PP variants, model-based and model-predictive methods, and learning-based and hybrid path-tracking methods. Section \ref{sec:dwpp} describes the proposed DWPP methodology in detail. Section \ref{sec:exp} presents real-robot experiments, which demonstrate the practical effectiveness of DWPP and form the core results of this paper, and Section \ref{sec:sim} presents simulation analyses that examine its behavior in more detail under idealized conditions. Section \ref{sec:discussion} provides a discussion, and Section \ref{sec:conclusion} concludes the paper and outlines directions for future work.}

\section{Related work}
\label{sec:related_work}

% \rev{This section first formulates the path-tracking problem addressed in this paper and then reviews three families of path-tracking controllers: the pure-pursuit family, model predictive control, and learning-based methods. Figure \ref{fig1} illustrates the processing flow of the major existing PP methods and the proposed DWPP.}

\rev{This section first formulates the path-tracking problem addressed in this paper and then reviews the pure-pursuit family, model-based and model-predictive control, and learning-based and hybrid control.}

\subsection{\texorpdfstring{\rev{Problem Formulation}}{Problem Formulation}}
\label{subsec:problem_formulation}

\begin{revblock}
This subsection formalizes the robot model, the path-tracking problem, and the input constraints addressed in this paper.

\paragraph{Robot Model}
This paper adopts the unicycle model,
\begin{equation}
  \dot{x} = v \cos\theta, \qquad
  \dot{y} = v \sin\theta, \qquad
  \dot{\theta} = \omega,
  \label{eq:unicycle}
\end{equation}
where $(x, y)$ and $\theta$ denote the position and orientation in the map coordinate system, and $v$ and $\omega$ denote the linear and angular velocities, respectively. The unicycle model is adopted because it represents the body-velocity command space $(v, \omega)$ shared by nonholonomic mobile robots. 

% The controller runs at a fixed control period $\Delta t$. The formulation treats a command that satisfies the input constraints below as realized within one control period. Section~\ref{sec:exp} examines how far the response lag of a real actuator affects this premise.

\paragraph{Path-Tracking Problem}
Consider a path $P = \{\boldsymbol{p}_0, \boldsymbol{p}_1, \ldots, \boldsymbol{p}_n\}$, where each waypoint $\boldsymbol{p}_i = (x_i, y_i) \in \mathbb{R}^2$ is defined in the map coordinate system. The path $P$ is provided externally, for example by a global planner, and generating or modifying it lies outside this problem formulation \cite{ollero1995stability, paden2016survey}. Let $\mathbf{x}_t = [x_t, y_t, \theta_t, v_t, \omega_t]^\top$ denote the robot's state at time $t$, consisting of its position $(x_t, y_t)$, orientation $\theta_t$, and current linear and angular velocities $(v_t, \omega_t)$.  A path-tracking controller is a function $f$ that computes the command velocities from the path and the state at every control period:
\begin{equation}
    (v_{\mathrm{cmd}}, \omega_{\mathrm{cmd}}) = f(P, \mathbf{x}_t).
\label{eq:pp}
\end{equation}
The path-tracking problem addressed in this paper is to design $f$ such that the robot progresses along $P$ while keeping the cross-track error
\begin{equation}
  e_t = \min_{i} \left\| (x_t, y_t) - \boldsymbol{p}_i \right\|
  \label{eq:cross_track_error}
\end{equation}
small, subject to the input constraints introduced next. This problem statement follows the standard formulation of path tracking as progressing along a geometric reference path while minimizing the deviation from it \cite{paden2016survey, snider2009automatic}.

% The formulation provides the controller with the true robot pose. On a real system, the pose comes from a localization module and contains estimation noise, and Section~\ref{sec:sim} analyzes its effect. Tracking performance is evaluated by the mean and maximum of $e_t$, the travel time, and the rate of commands violating the constraints. The formal metric definitions are given in Section~\ref{sec:exp}.

\paragraph{Input Constraints}
The command velocities computed by $f$ must respect the robot's velocity limits,
\begin{equation}
  v_{\min} \le v_{\mathrm{cmd}} \le v_{\max}, \qquad
  \omega_{\min} \le \omega_{\mathrm{cmd}} \le \omega_{\max},
  \label{eq:vel_limits}
\end{equation}
and its acceleration limits, which bound the velocity change achievable within one control period. For forward motion ($v_t \ge 0$), these read
\begin{equation}
  -a^{\mathrm{dcc}}_{\max} \le \frac{v_{\mathrm{cmd}} - v_t}{\Delta t} \le a^{\mathrm{acc}}_{\max}, \qquad
  -\alpha^{\mathrm{dcc}}_{\max} \le \frac{\omega_{\mathrm{cmd}} - \omega_t}{\Delta t} \le \alpha^{\mathrm{acc}}_{\max},
  \label{eq:acc_limits}
\end{equation}
where $v_{\max}$ ($v_{\min}$) and $\omega_{\max}$ ($\omega_{\min}$) are the maximum (minimum) linear and angular velocities, $a^{\mathrm{acc}}_{\max} > 0$ and $a^{\mathrm{dcc}}_{\max} > 0$ are the magnitudes of the maximum linear acceleration and deceleration, and $\alpha^{\mathrm{acc}}_{\max} > 0$ and $\alpha^{\mathrm{dcc}}_{\max} > 0$ are their angular counterparts. These body-velocity command limits are not estimates of the physical actuator limits but operational design parameters that the system integrator chooses, for example, for ride comfort and safety on a passenger-carrying platform \cite{de2023standards}. The configured limits are assumed to lie within the physical capabilities of the platform.

\end{revblock}

\begin{figure}[h]
\centering
\includegraphics[width=1.0\linewidth]{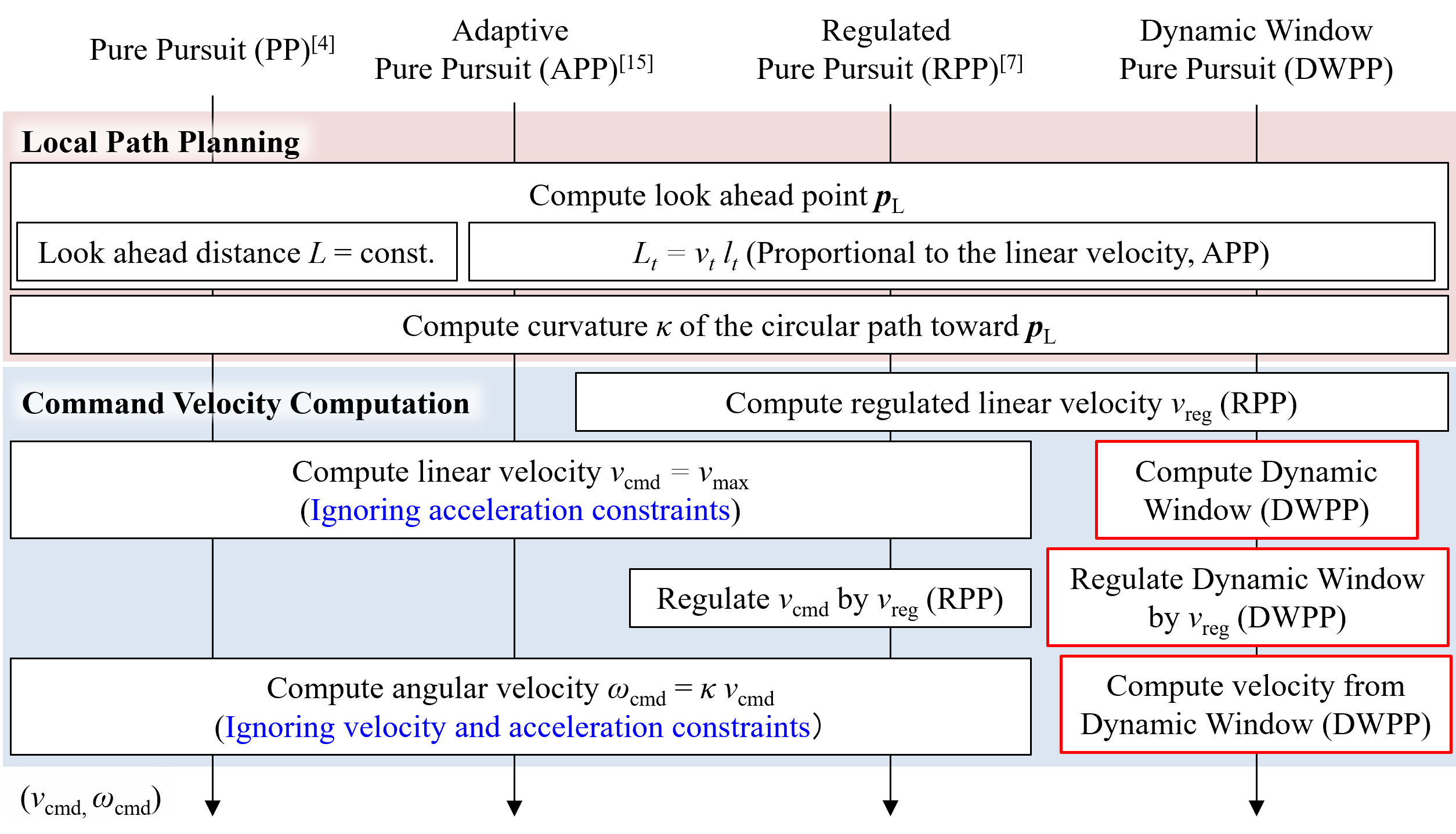} 
% Replace with actual filename
\caption{Processing flow of conventional pure pursuit methods and the proposed method.}
\label{fig1}
\end{figure}

\subsection{\texorpdfstring{\rev{Pure Pursuit}}{Pure Pursuit}}
\label{subsec:pp_family}

\rev{This subsection reviews the pure-pursuit family of path-tracking controllers, on which the proposed method builds.\footnote{\rev{Two distinct methods named ``adaptive pure pursuit'' appear in the literature: the classic velocity-scaled-lookahead APP \cite{campbell2007steering, ohta2016pure} reviewed below, and the more recent, unrelated method of Li et al.~\cite{li2023adaptive} cited among general PP improvements in Section~\ref{sec:intro}.}} Figure \ref{fig1} illustrates the processing flow of the major existing PP methods and the proposed DWPP. }

\subsubsection{Pure Pursuit (PP)}
\label{subsec:pp}

\begin{figure}[h]
\centering
\includegraphics[width=0.3\linewidth]{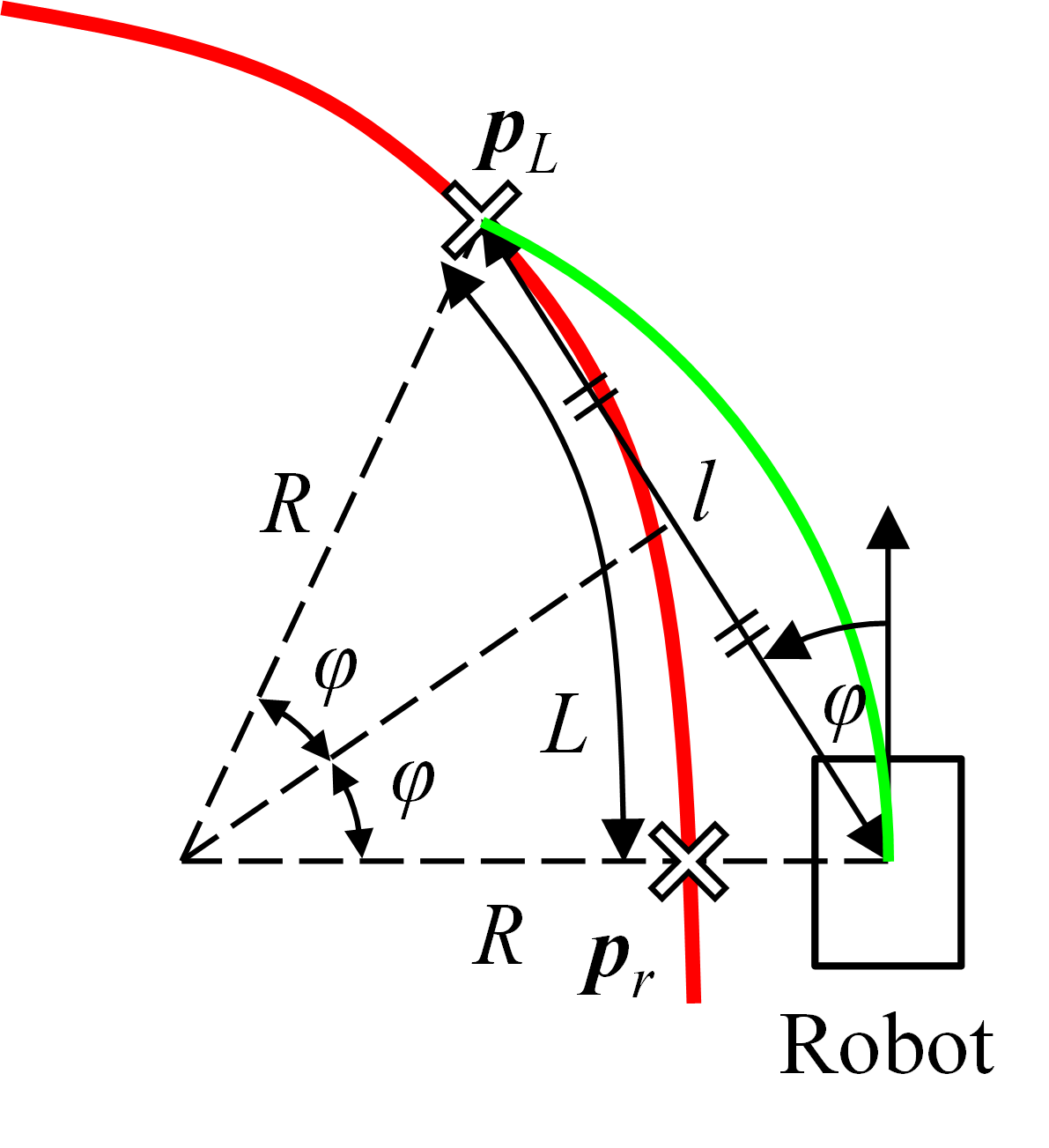} 
% Replace with actual filename
\caption{Geometric relationships between robot position, path, and lookahead position.}
\label{fig2}
\end{figure}

\rev{The PP algorithm \cite{coulter1992implementation} realizes the function $f$ of Eq.~\eqref{eq:pp} as follows.}

First, the algorithm computes the lookahead point $\boldsymbol{p}_L$ on the path, which serves as the reference position for path tracking. Figure \ref{fig2} shows the geometric relationship among the robot’s position, path, and lookahead point $\boldsymbol{p}_L$. Let $L$ be the lookahead distance. Let $\boldsymbol{p}_r = (x_r, y_r) \in P$ denote the point on the path nearest to the current robot position. Then, $\boldsymbol{p}_L$ is determined as follows:
\begin{equation}
  \mathrm{dist}(\boldsymbol{p}_i) =
  \sqrt{(x_r - x_i)^2 + (y_r - y_i)^2}
\end{equation}

\begin{equation}
  \boldsymbol{p}_L = \boldsymbol{p}_i \in P,
  \begin{cases}
    \mathrm{dist}(\boldsymbol{p}_{i-1}) < L \\
    \mathrm{dist}(\boldsymbol{p}_i) \ge L
  \end{cases}
  \label{eq:lookahead_point}
\end{equation}

Next, the curvature $\kappa$ of the arc that connects the current heading direction to the lookahead point is computed. Let $\varphi$ be the heading angle from the robot’s forward direction to $\boldsymbol{p}_L$. Let $l$ be the distance between the robot’s position and $\boldsymbol{p}_L$. From the geometric relationships shown in Fig.~\ref{fig2}, the radius $R$ of the circular path from the robot to $\boldsymbol{p}_L$ can be computed by

\begin{equation}
  R = \frac{l}{2 \sin \varphi}
  \label{eq:pp_radius}
\end{equation}

Hence, the curvature $\kappa$ can be derived by

\begin{equation}
  \kappa = \frac{1}{R} = \frac{2 \sin \varphi}{l}
  \label{eq:pp_curvature}
\end{equation}

Next, the command linear velocity \( v_{\mathrm{cmd}} \) is determined. In conventional PP formulations, \( v_{\mathrm{cmd}} \) is typically set to a constant value, namely, the robot’s maximum linear velocity \( v_{\max} \), as follows \cite{ollero1995stability}:
\begin{equation}
  v_{\mathrm{cmd}} = v_{\max}.
  \label{eq:v_cmd}
\end{equation}

Finally, the command angular velocity $\omega_{\mathrm{cmd}}$ is computed as follows. If the robot moves along a path of curvature $\kappa$ at $v_{\mathrm{cmd}}$, then

\begin{equation}
  \omega_{\mathrm{cmd}} = \kappa \, v_{\mathrm{cmd}}
  \label{eq:pp_angular_velocity}
\end{equation}

PP has the following three limitations:
\begin{itemize}
    \item \textbf{Difficulty in tuning the lookahead distance $L$:} Even if $L$ is tuned carefully, excessively sharp curves may result in overshooting behavior \cite{coulter1992implementation}. A short $L$ enables the robot to easily converge to the path but can introduce oscillations around the path center. A large $L$ can reduce oscillations but slow convergence, resulting in a trade-off.
    \item \textbf{Constant command linear velocity $v_\mathrm{cmd}$:} It does not adapt to the scenario. For example, even if a curve is extremely tight or if an obstacle is too close to the robot, the velocity will not automatically be reduced.
    \item \textbf{Velocity and acceleration constraints are not considered:} The command linear velocity is determined without accounting for acceleration constraints, whereas the command angular velocity is computed directly from the path curvature. Consequently, command velocities that explicitly satisfy both velocity and acceleration constraints are difficult to generate.
\end{itemize}

\subsubsection{Adaptive Pure Pursuit (APP)}

Adaptive pure pursuit (APP)~\cite{campbell2007steering}  addresses the first limitation. It extends the lookahead distance in proportion to the robot’s linear velocity. Specifically, APP sets the lookahead distance $L$ as follows~\cite{ohta2016pure}:

\begin{equation}
  L = \mathrm{clamp}\!\left( v_t l_t,\, L_{\min},\, L_{\max} \right)
  \label{eq:app_lookahead}
\end{equation}
where $l_t$ is a gain that indicates how many seconds ahead to project the velocity, and $L_{\min}$ and $L_{\max}$ are the minimum and maximum allowable lookahead distances, respectively. The function $\mathrm{clamp}(\text{value}, \text{low}, \text{high})$ returns values in $[\text{low}, \text{high}]$; if the value is below low, it is set to low, and if the value is above high, it is set to high. By adaptively adjusting the lookahead distance based on speed, APP reduces overshoot and oscillations at high speeds and enhances path convergence at low speeds, resulting in more stable and accurate path tracking. However, APP still leaves the second and third limitations unresolved.

\subsubsection{Regulated Pure Pursuit (RPP)}

The RPP controller \rev{\cite{macenski2023regulated}} extends the conventional PP method by introducing several heuristics that regulate the command linear velocity according to the path geometry and the surrounding environment. In RPP, the linear velocity is adjusted using three heuristics for the curvature, the proximity to obstacles, and the distance to the goal.

\paragraph{Curvature Heuristic}
The curvature heuristic reduces the linear velocity when the robot follows a path with a small curvature radius, thereby mitigating overshoot on sharp turns. The regulated velocity $v^\mathrm{curv}_{\mathrm{reg}}$ is defined based on \rev{the magnitude $|R|$ of} the curvature radius $R$ of the circular path from the robot to $\boldsymbol{p}_L$ as
\begin{equation}
  v^\mathrm{curv}_{\mathrm{reg}} =
  \begin{cases}
    v & \text{if } \rev{|R|} > R_{\min}, \\
    v \dfrac{\rev{|R|}}{R_{\min}} & \text{if } \rev{|R|} \le R_{\min},
  \end{cases}
  \label{eq:rpp_curvature}
\end{equation}
where $R_{\min}$ denotes the threshold radius below which velocity scaling is applied.

\paragraph{Proximity Heuristic}
The proximity heuristic decreases the velocity when the robot is close to obstacles to enhance safety. Let $d_O$ be the minimum distance to surrounding obstacles. The regulated velocity $v^\mathrm{prox}_{\mathrm{reg}}$ is given by
\begin{equation}
  v^\mathrm{prox}_{\mathrm{reg}} = 
  \begin{cases} 
    v & \text{if } d_O > d_{\mathrm{prox}}, \\
    v g_d \dfrac{d_O}{d_{\mathrm{prox}}} & \text{if } d_O \le d_{\mathrm{prox}},
  \end{cases}
  \label{eq:rpp_proximity}
\end{equation}
where $d_{\mathrm{prox}}$ is the proximity threshold, and $g_d \in (0, 1]$ is a gain parameter that controls the aggressiveness of the deceleration.

\paragraph{Goal Heuristic}
The goal heuristic gradually reduces the velocity as the robot approaches the goal to ensure a smooth and accurate stop. Let $d_G$ denote the remaining distance to the goal along the path, and $d_{\mathrm{goal}}$ be the distance threshold at which deceleration begins. The regulated velocity is computed as
\begin{equation}
  v^\mathrm{goal}_{\mathrm{reg}} = 
  \begin{cases} 
    v & \text{if } d_G > d_{\mathrm{goal}}, \\
    v \dfrac{d_G}{d_{\mathrm{goal}}} & \text{if } d_G \le d_{\mathrm{goal}}.
  \end{cases}
  \label{eq:rpp_goal}
\end{equation}

In the Nav2 implementation, the curvature and proximity heuristics are first applied to the command linear velocity $v_{\mathrm{cmd}}$. The intermediate regulated velocity is computed as
\begin{equation}
  v^{\mathrm{curv,prox}}_{\mathrm{reg}} =
  \rev{\max \left(
    \min \left(
      v^\mathrm{curv}_{\mathrm{reg}},
      v^\mathrm{prox}_{\mathrm{reg}}
    \right),
    v^\mathrm{min}_{\mathrm{reg}}
  \right)},
\end{equation}
where $v^\mathrm{min}_{\mathrm{reg}}$ is a parameter that specifies the minimum allowable regulated velocity.

Subsequently, the goal heuristic is applied to $v^{\mathrm{curv,prox}}_{\mathrm{reg}}$\rev{, i.e., $v$ in Eq.~\eqref{eq:rpp_goal} is replaced by $v^{\mathrm{curv,prox}}_{\mathrm{reg}}$}, and the final regulated velocity is computed as
\begin{equation}
  v_{\mathrm{reg}} =
  \rev{\min \left(
    v^{\mathrm{curv,prox}}_{\mathrm{reg}},\,
    \max \left(
      v^{\mathrm{goal}}_{\mathrm{reg}},
      v^\mathrm{min}_{\mathrm{goal}}
    \right)
  \right)},
\end{equation}
where $v^\mathrm{min}_{\mathrm{goal}}$ denotes the minimum linear velocity allowed when approaching the goal.

If the final regulated velocity $v_{\mathrm{reg}}$ is smaller than the originally computed command velocity $v_{\mathrm{cmd}}$, the command velocity is replaced by $v_{\mathrm{reg}}$, thereby limiting the translational motion.

This regulation strategy effectively reduces overshoot on sharp curves and improves safety in the vicinity of obstacles or people. However, RPP does not explicitly consider velocity and acceleration constraints during the command velocity computation, leaving this limitation unresolved.

\subsection{\texorpdfstring{\rev{Model-Based and Model Predictive Control}}{Model-Based and Model Predictive Control}}
\label{subsec:mpc}

\begin{revblock}
% Model-based and model predictive methods handle motion constraints by representing them explicitly in a model and, in most cases, solving an optimization problem at every control period. They can accommodate a wide range of constraints and objectives. DWPP concentrates on path tracking under velocity and acceleration constraints and solves it at a small computational and tuning cost. This subsection reviews these methods and clarifies the position of DWPP among them.

Model-based path-following methods track a path while accounting for the vehicle characteristics captured by their models. In this broad sense, the PP family is also model based because it builds on the unicycle model. Huskić et al. \cite{huskic2019highspeed} proposed a high-speed path-following controller for skid-steered vehicles that combines a Lyapunov-based feedback law with a longitudinal-speed controller. Built on an experimentally identified, terrain-dependent kinematic model, the controller derives reachable curvatures and admissible speeds from the bounded tread speeds and slows down when the path-following error grows. Their study adopts a model that captures the skid-steer kinematics and the tread-speed limits of a specific platform, aiming at high-speed driving that fully exploits its physical capability. DWPP adopts the general unicycle model of Section~\ref{subsec:problem_formulation} together with the configured velocity and acceleration limits, aiming at constraint-satisfying path tracking that applies uniformly to nonholonomic platforms.

Model predictive control (MPC) enforces constraints, including velocity and acceleration limits, as part of a finite-horizon optimization. Representative examples include the timed-elastic-band approach, which optimizes a time-parameterized local trajectory online \cite{rosmann2015timed}, robust constrained learning-based nonlinear MPC for mobile robots \cite{ostafew2016nmpc}, and the combination of PP with MPC that treats the lookahead distance as an optimization variable \cite{kim2024ppmpc}. MPC-DS adds a control-barrier-function constraint for avoiding dense dynamic obstacles and thereby addresses a broader local-navigation problem than the path tracking of Section~\ref{subsec:problem_formulation} \cite{zhang2025mpcds}. Model predictive path integral control (MPPI) solves the finite-horizon problem with a sampling-based stochastic solver that perturbs a nominal control sequence, forward-simulates the candidate trajectories under a motion model, and updates the sequence according to their evaluated costs \cite{williams2017information}. The MPPI controller distributed with Nav2 implements this procedure with plugin-based critic functions (modular cost terms) and configurable velocity and acceleration limits \cite{macenski2023survey}.

While these predictive approaches cover a broad range of constraints and objectives, DWPP focuses on the single objective of path tracking under only the velocity and acceleration constraints, as formulated in Section~\ref{subsec:problem_formulation}, and solves this problem in closed form. Instead of predicting future states over a horizon, DWPP considers only the next control period and obtains the path-tracking command velocity with an $O(1)$ computation that satisfies the limits by construction, without an iterative solver or sampled trajectory rollouts. The tuning also stays simple. DWPP inherits the few, geometrically interpretable parameters of the PP family, whereas the cost weights of MPC and MPPI become more laborious to tune and harder to interpret as objectives are added. DWPP therefore complements predictive methods and offers a lightweight solution for the common case in which path tracking under configured velocity and acceleration limits is the primary objective.
\end{revblock}

\subsection{\texorpdfstring{\rev{Learning-Based and Hybrid Path-Tracking Control}}{Learning-Based and Hybrid Path-Tracking Control}}
\label{subsec:rl}

\begin{revblock}
Learning-based path-tracking methods use data or interaction to learn either a control policy or parameters that augment an analytical controller. Their strength is adaptability to conditions that are difficult to model. Reinforcement learning (RL) can represent nonlinear mappings from observed states to control actions and can adapt controller behavior across operating conditions represented during training. Hybrid architectures are particularly relevant to path tracking because they retain an interpretable model-based or geometric controller while assigning selected adaptation tasks to a learned policy.

Xia et al. \cite{xia2025adaptive} combined a preview model with a twin delayed deep deterministic policy gradient lateral controller, an adaptive correction module based on predicted waypoints along the path, and a soft actor-critic-based longitudinal speed planner. Han et al. \cite{han2025hybrid} integrated PP with proximal policy optimization and an adaptive lookahead mechanism for autonomous trucks. In the latter approach, PP provides the baseline geometric controller, whereas the learned component improves adaptation to nonlinear vehicle behavior. These studies illustrate the two principal roles of RL in path tracking. One is generating or correcting control actions, and the other is adapting controller parameters such as the lookahead distance.

This adaptability comes with a training and verification burden and without built-in constraint guarantees. RL shifts part of the design effort to the state and action spaces, the reward, and the training scenarios, and a standard unconstrained RL objective does not by itself ensure that every command satisfies prescribed velocity and acceleration limits, so explicit safety mechanisms are required when formal constraint satisfaction is needed \cite{brunke2022safe}. DWPP offers the complementary strengths. It computes commands without training, keeps the computation interpretable, and satisfies the configured velocity and acceleration limits by construction, while RL-based methods excel at adapting to nonlinearities and unmodeled effects.
\end{revblock}

\section{Dynamic Window Pure Pursuit (DWPP)}
\label{sec:dwpp}

The proposed DWPP fundamentally reformulates the command velocity computation process to address the third \rev{limitation listed in Section~\ref{subsec:pp}}, namely, the lack of explicit consideration of velocity and acceleration constraints in conventional command computation and regulation. As illustrated in Fig.~\ref{fig1}, DWPP differs from RPP through the following three processes. First, the dynamic window, defined as the feasible velocity region for the next control step under the robot’s velocity and acceleration constraints, is computed. Second, this dynamic window is regulated using the regulated linear velocity $v_{\mathrm{reg}}$. Third, from the resulting region, the velocity point closest to the line $\omega = \kappa v$ (Eq.~\eqref{eq:pp_angular_velocity}) is selected as the command velocity.

Through these steps, DWPP explicitly accounts for velocity and acceleration constraints when computing the path-tracking command velocity. This section describes the three processes in detail and presents the stability analysis of DWPP. The implementation is publicly available.\footnote{
\url{https://github.com/ros-navigation/navigation2/blob/main/nav2_regulated_pure_pursuit_controller/include/nav2_regulated_pure_pursuit_controller/dynamic_window_pure_pursuit_functions.hpp}
} The overall procedure corresponds to the implementation in the \texttt{computeDynamicWindow\-Velocities} function.

\subsection{Compute the dynamic window}
First, the maximum and minimum feasible linear and angular velocities at the next calculation step, denoted by $v^\mathrm{dw}_{\max}, v^\mathrm{dw}_{\min}, \omega^\mathrm{dw}_{\max}$, and $\omega^\mathrm{dw}_{\min}$ are computed using the following equations:
\begin{align}
    v^\mathrm{dw}_{\max} &= \min(v_{\max}, v_t + a^{\mathrm{acc}}_{\max} \Delta t) \label{eq:dw_v_max} \\
    v^\mathrm{dw}_{\min} &= \max(v_{\min}, v_t - a^\mathrm{dcc}_{\max} \Delta t) \\
    \omega^\mathrm{dw}_{\max} &= \min(\omega_{\max}, \omega_t + \alpha^{\mathrm{acc}}_{\max} \Delta t) \\
    \omega^\mathrm{dw}_{\min} &= \max(\omega_{\min}, \omega_t - \alpha^\mathrm{dcc}_{\max} \Delta t) \label{eq:dw_w_min}
\end{align}
\rev{where the velocity limits, the acceleration limits, and the control period $\Delta t$ are as defined in Section~\ref{subsec:problem_formulation}.} This procedure corresponds to the implementation in the \texttt{computeDynamicWindow} function.

\subsection{Apply regulation to the dynamic window}

Next, the values $v^{\mathrm{dw}}_{\max}$ and $v^{\mathrm{dw}}_{\min}$ are constrained by the regulated command linear velocity $v_{\mathrm{reg}}$ computed by RPP.

Specifically, because the admissible range of the robot’s linear velocity is limited to $[0,\, v_{\mathrm{reg}}]$, the dynamic window is restricted by taking the intersection between the original dynamic window and this velocity range. \rev{If this intersection is empty, the linear-velocity range of the window collapses to the reachable bound closest to $[0,\, v_{\mathrm{reg}}]$.}

This procedure corresponds to the implementation in the \texttt{apply-\\RegulationToDynamicWindow} function.

\begin{figure}[t]
     \centering
     \begin{subfigure}[b]{0.48\linewidth}
         \centering
         \includegraphics[width=\linewidth]{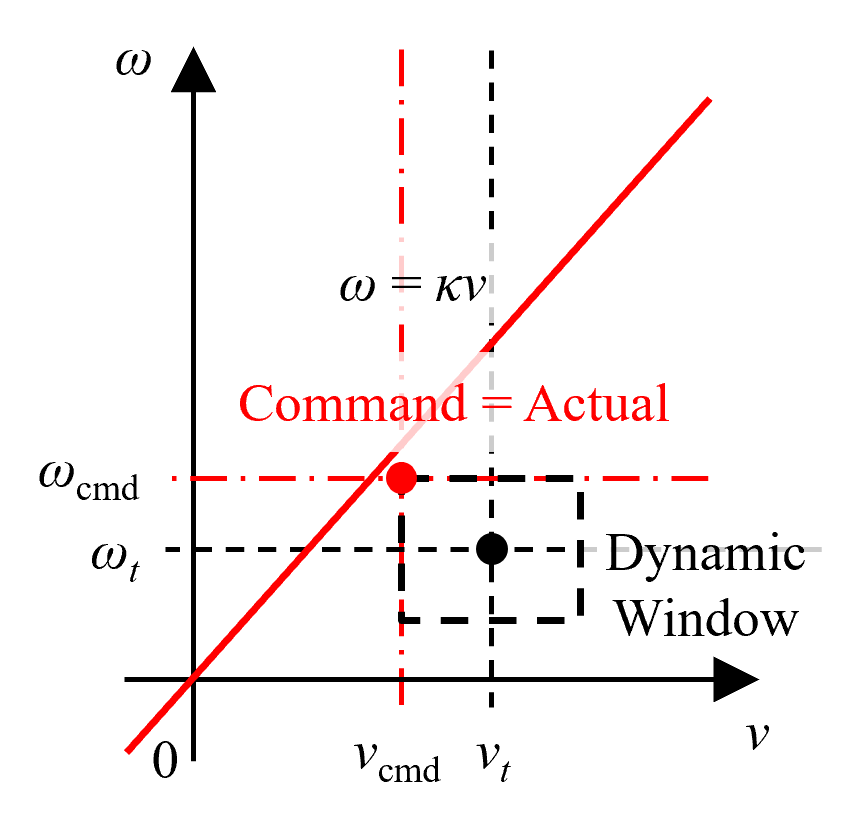}
         \caption{Proposed DWPP}
         \label{fig:vw_dwpp}
     \end{subfigure}
     \hfill
     \begin{subfigure}[b]{0.48\linewidth}
         \centering
         \includegraphics[width=\linewidth]{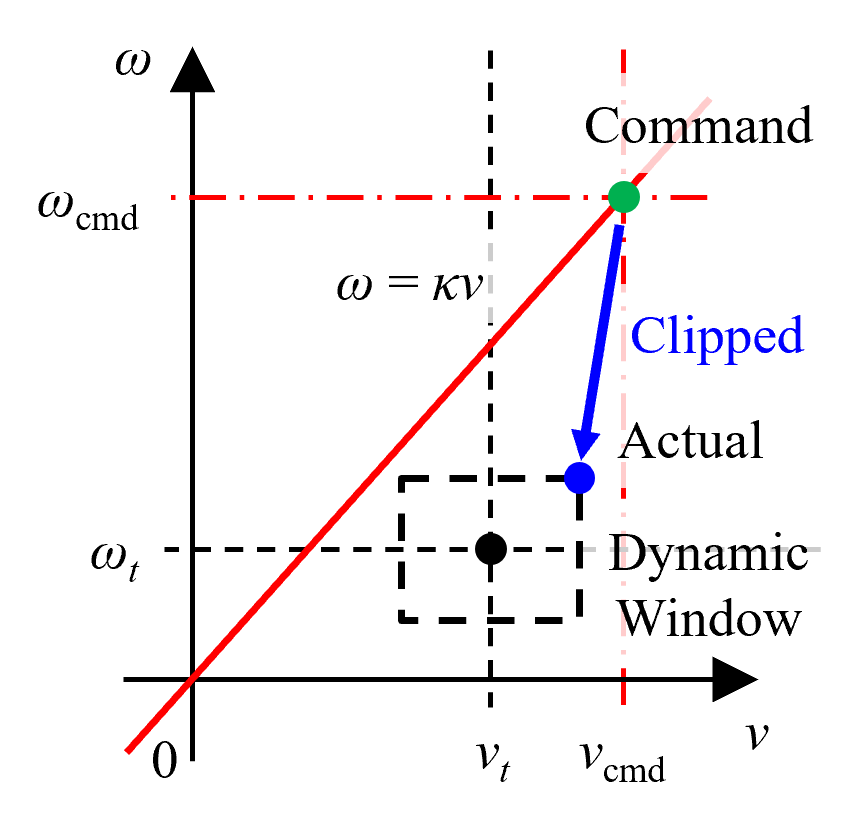}
         \caption{Conventional pure pursuit}
         \label{fig:vw_conventional}
     \end{subfigure}
     \caption{Comparison of command velocity computations in the \( v \)--\( \omega \) space.
(a) DWPP selects an executable velocity within the dynamic window closest to the line $\omega=\kappa v$,
whereas (b) conventional PP may select a velocity outside the dynamic window, which is subsequently clipped by constraints.}
     \label{fig:vw_comparison}
\end{figure}

\subsection{Compute Optimal Velocity within the Dynamic Window}
\label{sec:3}

Third, the optimal velocity command is selected from the dynamic window to best satisfy the path-following condition defined in Eq.~\eqref{eq:pp_angular_velocity}.

Figure~\ref{fig:vw_dwpp} illustrates an overview of the velocity selection process in the \( v \)--\( \omega \) space. Noting that Eq.~\eqref{eq:pp_angular_velocity} represents a straight line passing through the origin in the \( v \)--\( \omega \) space, the command velocity is determined by selecting the point within the dynamic window that is closest to the line. If multiple candidate points achieve the same minimum distance to this line, the one with the largest linear velocity is selected to reduce the time required to reach the goal. This strategy yields a velocity command that is both feasible and most closely aligned with the ideal path-following velocity.

The specific implementation is conducted as follows. Note that the following procedure corresponds to the implementation of the \texttt{computeOptimal\-VelocityWithinDynamicWindow} function.

First, consider the case where $\kappa = 0$. The linear velocity is set to $v^{\mathrm{dw}}_{\max}$.
% for forward motion and to $v^{\mathrm{dw}}_{\min}$ for backward motion.
For the angular velocity, if the line $\omega = \kappa v$ intersects the dynamic window, the intersection point is selected, which yields $\omega = 0$. If no intersection exists, either $\omega^{\mathrm{dw}}_{\max}$ or $\omega^{\mathrm{dw}}_{\min}$ is selected, whichever is closer to zero.

Next, consider the case where $\kappa \neq 0$ and the line $\omega = \kappa v$ intersects the dynamic window. The intersections between the line $\omega = \kappa v$ and the four extended edges of the dynamic window are given by

\begin{revblock}
\begin{equation}
\begin{aligned}
\mathbf{p}_1 &= \left( v^{\mathrm{dw}}_{\min},\ \kappa\, v^{\mathrm{dw}}_{\min} \right), \\
\mathbf{p}_2 &= \left( v^{\mathrm{dw}}_{\max},\ \kappa\, v^{\mathrm{dw}}_{\max} \right), \\
\mathbf{p}_3 &= \left( \frac{\omega^{\mathrm{dw}}_{\min}}{\kappa},\ \omega^{\mathrm{dw}}_{\min} \right), \\
\mathbf{p}_4 &= \left( \frac{\omega^{\mathrm{dw}}_{\max}}{\kappa},\ \omega^{\mathrm{dw}}_{\max} \right).
\end{aligned}
\label{eq:vw_intersection_points}
\end{equation}
\end{revblock}

Among these points, the one that lies inside the dynamic window and has the largest linear velocity is selected. If none of these intersection points lies inside the dynamic window, then the line $\omega = \kappa v$ does not intersect the dynamic window.

Finally, consider the case where $\kappa \neq 0$ and no intersection exists. From convex geometry, we know that when a convex set and line do not intersect, the point on the convex hull that minimizes the distance to the line must lie at a vertex of the hull or on an edge parallel to the line. Therefore, evaluating the four vertices of the dynamic window is sufficient. Specifically, the distances between the line $\omega = \kappa v$ and the following four points are computed, and the point with the shortest distance and the largest linear velocity is selected:

\begin{revblock}
\begin{equation}
\begin{aligned}
\mathbf{c}_1 &= \left( v^{\mathrm{dw}}_{\min},\ \omega^{\mathrm{dw}}_{\min} \right), \\
\mathbf{c}_2 &= \left( v^{\mathrm{dw}}_{\min},\ \omega^{\mathrm{dw}}_{\max} \right), \\
\mathbf{c}_3 &= \left( v^{\mathrm{dw}}_{\max},\ \omega^{\mathrm{dw}}_{\min} \right), \\
\mathbf{c}_4 &= \left( v^{\mathrm{dw}}_{\max},\ \omega^{\mathrm{dw}}_{\max} \right).
\end{aligned}
\end{equation}
\end{revblock}

That is, the computational complexity of the DWPP algorithm is $O(1)$, the same as that of conventional PP variants.

Figure~\ref{fig:vw_comparison} illustrates the command velocity computation in the \( v \)--\( \omega \) space. As shown in Fig.~\ref{fig:vw_conventional}, RPP first computes a command velocity (green dot) that best follows the path without explicitly considering velocity and acceleration constraints. The actual velocity executed by the robot (blue dot) is then obtained by clipping this command according to the constraints. In contrast, as shown in Fig.~\ref{fig:vw_dwpp}, DWPP selects a command velocity (red dot) that is closest to \( \omega = \kappa v \) within the dynamic window, and this velocity is directly executed. In other words, DWPP computes the command velocity that best follows the path while explicitly respecting velocity and acceleration constraints. \rev{For} example, it reduces the linear velocity when tracking a path with large curvature, whereas conventional PP methods tend to increase it, as illustrated in Fig.~\ref{fig:vw_comparison}.

\subsection{Stability Analysis}
\label{sec:stability}

\begin{revblock}
For conventional PP, Ollero and Heredia \cite{ollero1995stability} analyzed the stability for straight and constant-curvature reference paths and derived conditions on the lookahead distance for asymptotically stable tracking, assuming a constant linear velocity and a constant lookahead distance. For RPP, Macenski et al.~\cite{macenski2023regulated} explained that these convergence properties carry over because the linear velocity regulation preserves the PP velocity relation $\omega=\kappa v$. Indeed, as long as the angular velocity is recomputed as $\omega=\kappa v$, the robot moves along the same arc of curvature $\kappa$ for any positive linear velocity, so varying $v$ only rescales the geometric convergence of PP in time. This argument extends to DWPP whenever the line $\omega=\kappa v$ intersects the dynamic window, because DWPP then also selects a feasible command on the line.

However, when the line $\omega = \kappa v$ does not intersect the dynamic window, this discussion is not applicable, because the relation $\omega = \kappa v$ cannot be realized exactly.
% DWPP then selects the feasible command that minimizes the deviation $|\omega - \kappa v|$, typically by reducing the linear velocity and steering the angular velocity toward the required direction.
This non-intersection case is nevertheless a transient constraint-active mode. Specifically, if the curvature command and the regulated velocity stay constant while this deviation lasts, the deviation of the executed command decreases at every control step by at least a positive amount, moving the subsequent dynamic window toward the line, so the intersection is recovered within a finite number of steps (Proposition~\ref{prop:recovery} in \ref{app:stability}, which also shows that the recovery tolerates small step-to-step variations of both quantities). Once the intersection is recovered, DWPP again selects a command on the line $\omega = \kappa v$, and the next dynamic window is built around this command and hence contains it. While the intersection is maintained in this way, the convergence discussion at the beginning of this subsection applies.

\end{revblock}

\section{Experiment}
\label{sec:exp}

\rev{Two real-robot experiments were conducted to verify the effectiveness of the proposed DWPP. The first is a path-tracking experiment on obstacle-free reference paths with increasing corner sharpness (Section~\ref{sec:experiment}), and the second is a confined corridor experiment in an obstacle environment with RPP's proximity heuristic active (Section~\ref{sec:corridor}). This section first describes the settings common to both experiments.}

\subsection{\texorpdfstring{\rev{Common Setting}}{Common Setting}}
\label{sec:exp_common}

\begin{figure}[!tbp]
    \centering
    \includegraphics[width=\linewidth, height=5cm, keepaspectratio]{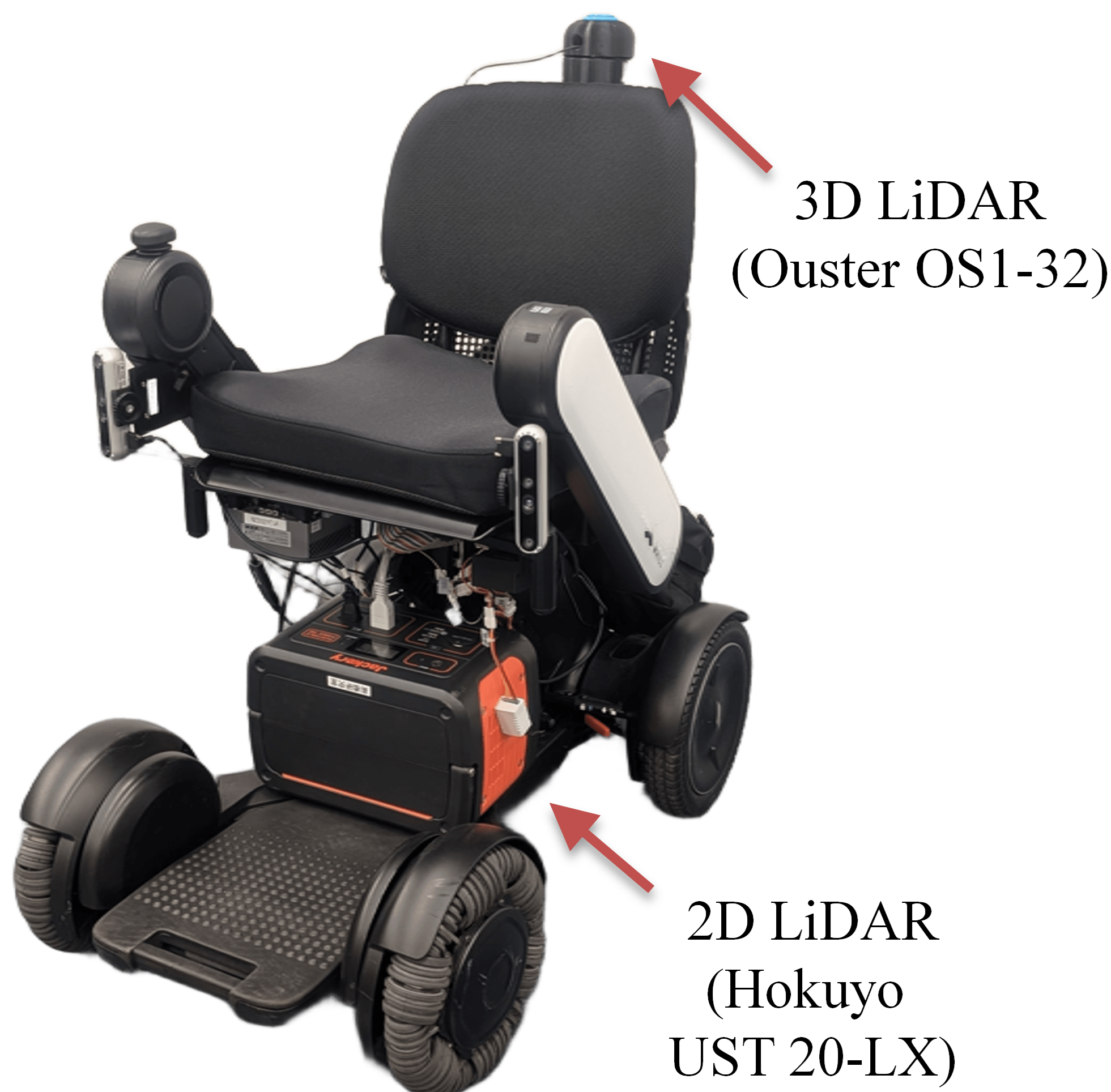}
    \caption{Robot used in the experiment.}
    \label{fig:robot_image}
\end{figure}

The robot used in the experiment was a WHILL Model CR, shown in Fig.~\ref{fig:robot_image}. The robot measured 0.985~m in length and 0.55~m in width. A 2D map for localization was created using the SLAM Toolbox~\cite{macenski2021slam}, and self-localization was performed using \rev{the adaptive Monte Carlo localization (AMCL) package of} Nav2. {\sloppy \rev{The software was implemented on Ubuntu 24.04 using ROS~2 Jazzy, and all computations were executed on a Minisforum UM890 Pro mini-PC equipped with an AMD Ryzen~9~8945HS CPU and 96~GB of RAM. For navigation, the Nav2 framework was employed. In the implementation, the \texttt{RegulatedPurePursuitController} plugin was used in the Nav2 \texttt{controller\_server}, and the controller type was switched among PP, APP, RPP, and DWPP via parameter settings. Downstream of the controller, the command velocity was clipped to the same velocity and acceleration limits by the Nav2 \texttt{velocity\_smoother}, which was active for every controller, and the recorded command velocities are the raw controller outputs before this clipping. Note that DWPP constructs the dynamic window of Eqs.~\eqref{eq:dw_v_max}--\eqref{eq:dw_w_min} using the velocity command of the previous control period as the current velocity $(v_t, \omega_t)$, so that the window is not affected by the noise of the measured velocity. Experimental data were recorded once per control period (30~Hz).}\par}

The performance was evaluated using the following four metrics:
\begin{enumerate}
    \item \textbf{Constraint violation ratio [\%]}: The percentage of control steps in which the computed command velocities $(v_{\mathrm{cmd}}, \omega_{\mathrm{cmd}})$ violated \rev{the velocity and acceleration limits of Eqs.~\eqref{eq:vel_limits} and \eqref{eq:acc_limits}. In this evaluation, the velocity command of the previous control period is used as the current velocity, so that the metric reflects the property of the controller itself, independently of the noise of the measured velocity.}
    \item \textbf{Mean path tracking error [m]}: The average lateral deviation between the actual robot and reference paths, representing the overall tracking accuracy\rev{, computed as $\bar{e} = \frac{1}{N} \sum_{t=1}^{N} e_t$, where $e_t$ is the cross-track error of Eq.~\eqref{eq:cross_track_error} at control step $t$ and $N$ is the number of control steps in a run}.
    \item \textbf{Maximum path tracking error [m]}: The peak lateral deviation from the reference path, which characterizes the magnitude of overshoot\rev{, computed as $e^{\max} = \max_{1 \le t \le N} e_t$}.
    \item \textbf{Travel time [s]}: The total duration required for the robot to reach the goal.
\end{enumerate}

\begin{revblock}
The measurement setup involves three practical aspects specific to the real-robot experiments. First, the realized velocities shown in the velocity profiles are obtained from the wheel-encoder-based odometry of the robot platform, i.e., computed from the motor rotation. Second, the robot pose used both by the controller and in the error metric is the AMCL estimate and therefore contains localization noise. Likewise, the actuation involves a small response lag about 0.007s, so commands are not realized instantaneously. The experiments in this section examine whether the controllers remain effective under such practical conditions. Third, the offset between the LiDAR and the robot base frame is compensated by a static transform taken from the mechanical design, and the localization, the control, and the error evaluation all refer to the base frame through this transform.
\end{revblock}

\begin{figure}[!tbp]
    \centering
    \includegraphics[width=\linewidth, height=5cm, keepaspectratio]{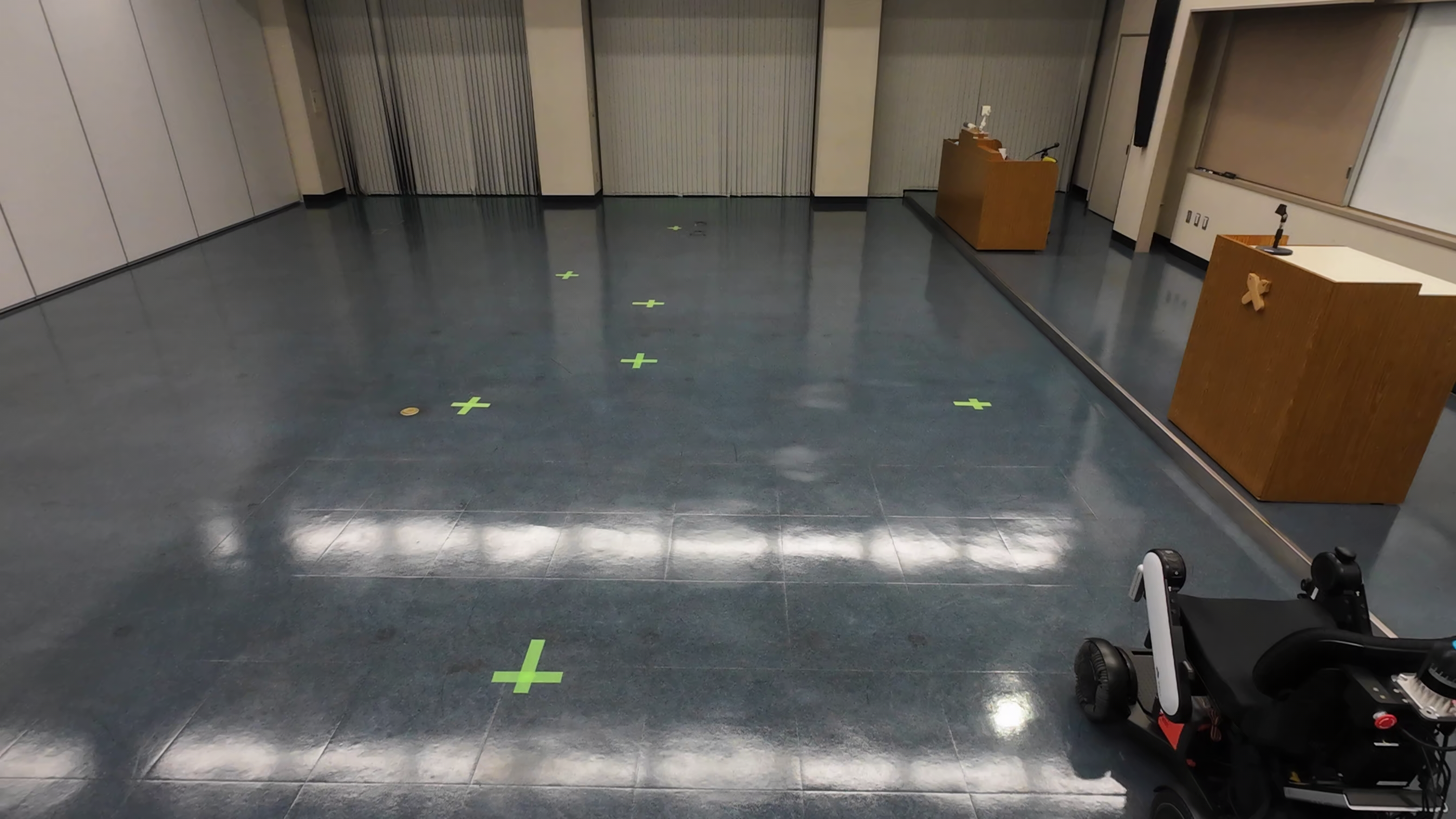}
    \caption{Experimental environment.}
    \label{fig:experiment_environment}
\end{figure}
\begin{figure}[!tbp]
    \centering

    \begin{subfigure}[b]{0.32\linewidth}
        \centering
        \includegraphics[width=\linewidth, height=5cm, keepaspectratio]{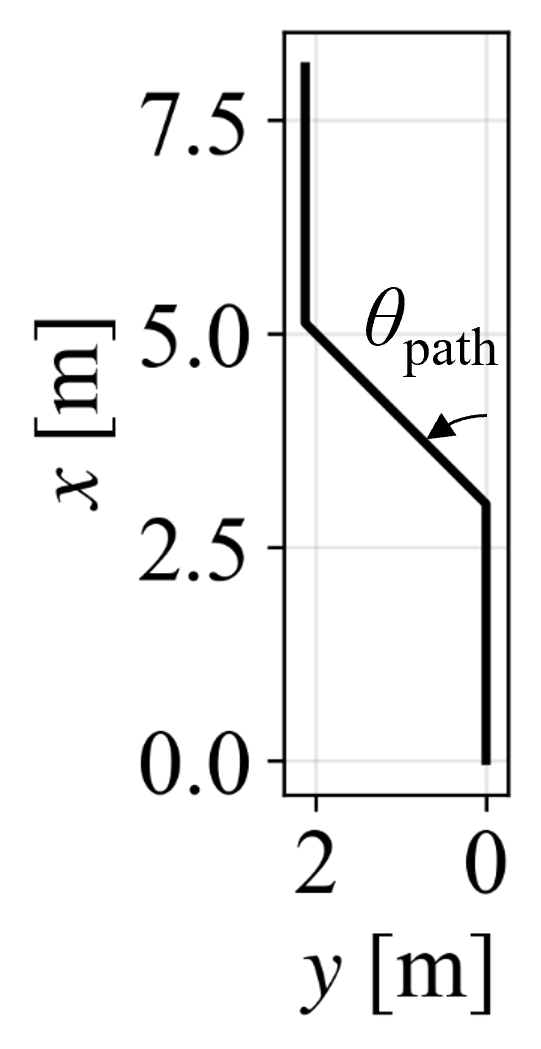}
        \caption{Path A ($\theta_\mathrm{path} = 45 \deg$)}
    \end{subfigure}
    \hfill
    \begin{subfigure}[b]{0.32\linewidth}
        \centering
        \includegraphics[width=\linewidth, height=5cm, keepaspectratio]{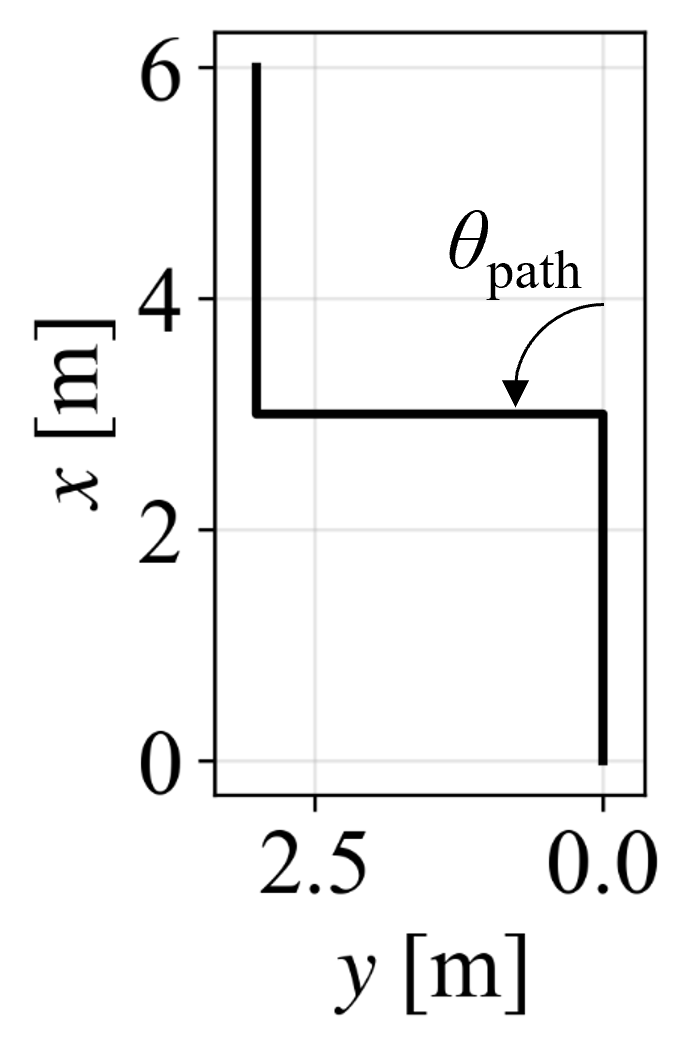}
        \caption{Path B ($\theta_\mathrm{path} = 90 \deg$)}
    \end{subfigure}
    \hfill
    \begin{subfigure}[b]{0.32\linewidth}
        \centering
        \includegraphics[width=\linewidth, height=5cm, keepaspectratio]{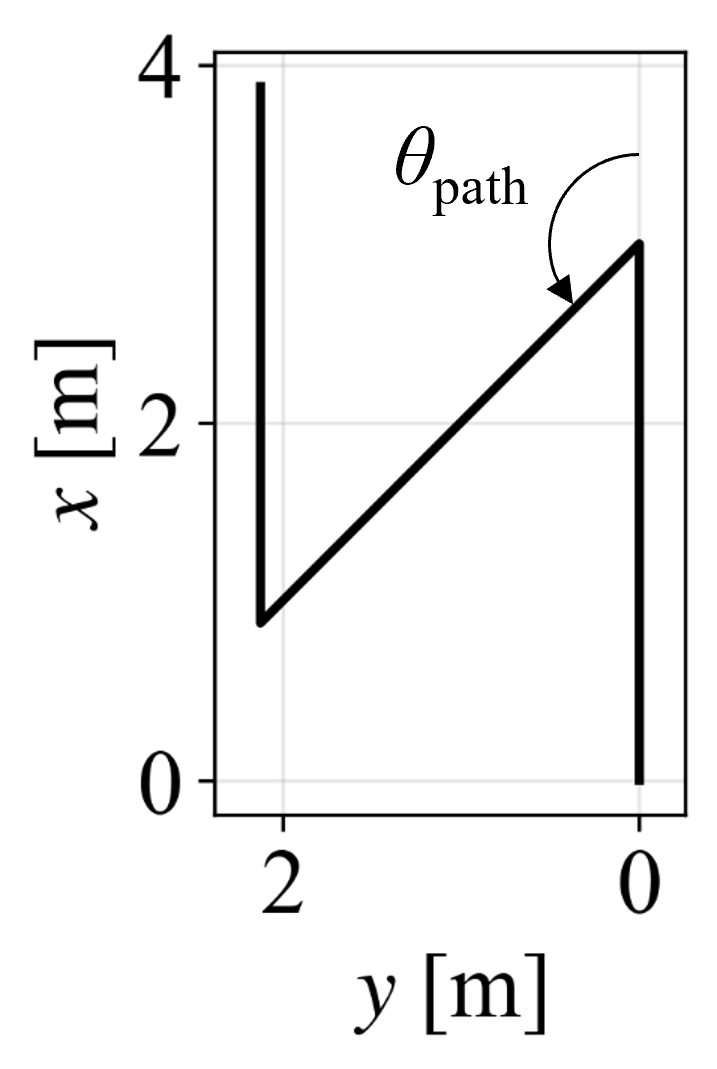}
        \caption{Path C ($\theta_\mathrm{path} = 135 \deg$)}
        \label{fig:pathC}
    \end{subfigure}

    \caption{Reference paths used in the experiment.}
    \label{fig:reference_paths}
\end{figure}

\subsection{Path Tracking Experiment}
\label{sec:experiment}

\subsubsection{Setting}
\label{sec:experiment_setting}
\rev{The path-tracking experiment compares the tracking performance of the controllers on obstacle-free reference paths. The experiments were conducted in an obstacle-free room, as illustrated in Fig.~\ref{fig:experiment_environment}. Because no obstacles were present, the proximity heuristic of RPP and DWPP was disabled in this experiment.}

The changed conditions in this experiment were the global path and local planner. Three paths (Paths A, B, and C) were defined as global paths, that is, reference paths, as shown in Fig.~\ref{fig:reference_paths}. \rev{Each reference path consists of waypoints spaced $0.01$~m apart.} All paths shared the same segment length of 3.0 m, whereas the corner angle $\theta_{\mathrm{path}}$ was set to $45^\circ$, $90^\circ$, and $135^\circ$ for Paths A, B, and C, respectively. To ensure that the robot’s initial pose at the start of tracking exactly matched the reference path, we generated the reference path for each trial based on the robot’s initial pose. \rev{The evaluated local planners were} PP, APP, RPP, the proposed DWPP\rev{, and MPPI \cite{williams2017information}, the sampling-based MPC controller}. \rev{To account for trial-to-trial variability, five trials were conducted} for each combination of the \rev{five} controllers and three paths, resulting in a total of \rev{75} experimental runs. \rev{In addition to the metrics of Section~\ref{sec:exp_common}, the per-cycle computation time of the command velocity computation was measured for every controller. Measured over all trials of this experiment, the per-update corrections of the AMCL pose estimate had root-mean-square magnitudes of about $0.010$~m in position and $0.0063$~rad in heading, arriving at about 7~Hz.}

\begin{table}[t]
\centering
\caption{Parameters \rev{of the PP family} used in the experiment}
\label{tab:exp_parameters}
\begin{tabular}{ll}
\toprule
Parameter & Value \\ \midrule
$v_{\max}, v_{\min}$ & 0.5, 0.0 [m/s] \\
$a^{\mathrm{acc}}_{\max}, a^{\mathrm{dcc}}_{\max}$ & 0.5, 0.5 [m/s$^2$] \\
$\omega_{\max}, \omega_{\min}$ & 1.0, $-1.0$ [rad/s] \\
$\alpha^{\mathrm{acc}}_{\max}, \alpha^{\mathrm{dcc}}_{\max}$ & 1.0, 1.0 [rad/s$^2$] \\
$\Delta t$ & 0.033 [s] \\
$L$ & 0.6 [m] \\
$L_{\min}, L_{\max}$ & 0.3, 0.7 [m] \\
$l_t$ & 1.4 \\
$R_{\min}$ & 0.9 [m] \\
$v^\mathrm{min}_{\mathrm{reg}}$ & 0.25 [m/s] \\ \bottomrule
\end{tabular}
\end{table}

{\sloppy The parameters \rev{of PP family} used in the experiment are summarized in Table~\ref{tab:exp_parameters}. \rev{The velocity and acceleration constraints were set operationally for ride comfort and safety on the passenger-carrying platform, within the measured physical capabilities of the robot (approximately $1.67$~m/s and $1.7$~m/s$^2$ in translation and $1.69$~rad/s and $5.7$~rad/s$^2$ in rotation), as required in Section~\ref{subsec:problem_formulation}.} The constant lookahead distance $L$ for the PP controller was set slightly larger than the stability limit, considering the maximum linear velocity. For the other controllers (APP, RPP, and DWPP), the lookahead distance limits ($L_{\min}, L_{\max}$) were configured such that the distance at maximum velocity exceeded the constant value used in PP while maintaining tracking accuracy at lower speeds through a smaller $L_{\min}$. Parameters specific to RPP ($R_{\min}$, $v^\mathrm{min}_{\mathrm{reg}}$) were set to their default Nav2 values \rev{for a fair comparison}. Additionally, the Nav2 parameter \texttt{use\_rotate\_to\_heading} was set to \texttt{false} to prevent the robot from stopping at cusp points and ensure continuous path tracking. \texttt{use\_collision\_detection} was also set to \texttt{false} to avoid unintended deceleration caused by sensor noise. All other parameters were maintained at their default values provided by the Nav2 framework.\par}

\begin{revblock}
The MPPI baseline used the \texttt{nav2\_mppi\_controller::MPPIController} plugin of the Nav2 \texttt{controller\_server}, with the same velocity and acceleration limits (Table~\ref{tab:exp_parameters}) and the same 30~Hz control frequency as the PP-family controllers, the \texttt{DiffDrive} motion model, a prediction horizon of 4.0~s (120 steps of the 33~ms control period), and 2000 sampled trajectories per cycle. Like DWPP, MPPI used the velocity command of the previous control period as the current velocity from which the sampled trajectories start. Starting from the default critic configuration of Nav2, tuning maximized the path-tracking accuracy of MPPI on the reference paths. The tuned configuration uses trajectory-sampling standard deviations of $(0.20~\mathrm{m/s},\, 0.50~\mathrm{rad/s})$ and, among the enabled critics, strengthens the ones that keep the sampled trajectories close to the path, progressing along it, and aligned with its direction, namely the path-alignment critic (weight 20 with a squared cost) together with the path-following and path-angle critics (weights 10 and 5). The complete parameter file used in this experiment is available online.\footnote{\rev{\url{https://github.com/decwest/dwpp_test_simulation/blob/main/params/path_tracking_experiment.yaml}}}
\end{revblock}

\subsubsection{Results}
\label{sec:experiment_result}

\begin{figure*}[tbp]
    \centering
    
    % ---- 凡例（横長） ----
    \begin{subfigure}[b]{0.80\textwidth}
        \centering
        \includegraphics[width=\textwidth]{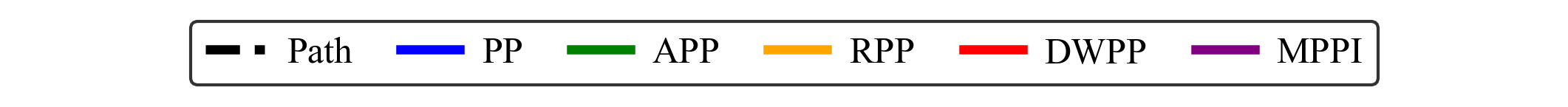}
    \end{subfigure}
    
    % 1. 経路比較
    \begin{subfigure}[b]{0.80\textwidth}
        \centering
        \includegraphics[width=\textwidth, height=6.5cm, keepaspectratio]{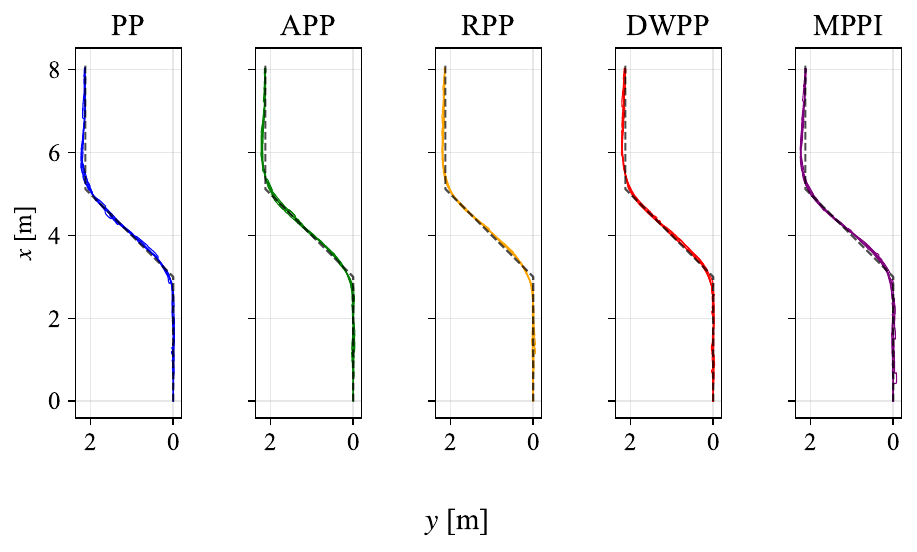}
        \caption{Path comparison}
        \label{fig:path_comparison_pathA}
    \end{subfigure}

    \vspace{0.5em}

    % ---- 凡例（横長） ----
    \begin{subfigure}[b]{0.45\textwidth}
        \centering
        \includegraphics[width=\textwidth]{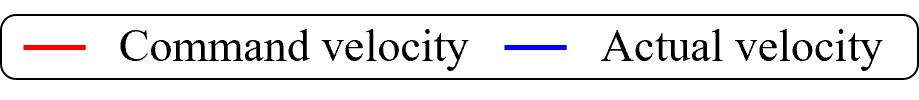}
    \end{subfigure}
    
    \vspace{0.5em}

    % 2. 速度プロファイル群
    \begin{subfigure}[b]{0.45\textwidth}
        \centering
        \includegraphics[width=\textwidth]{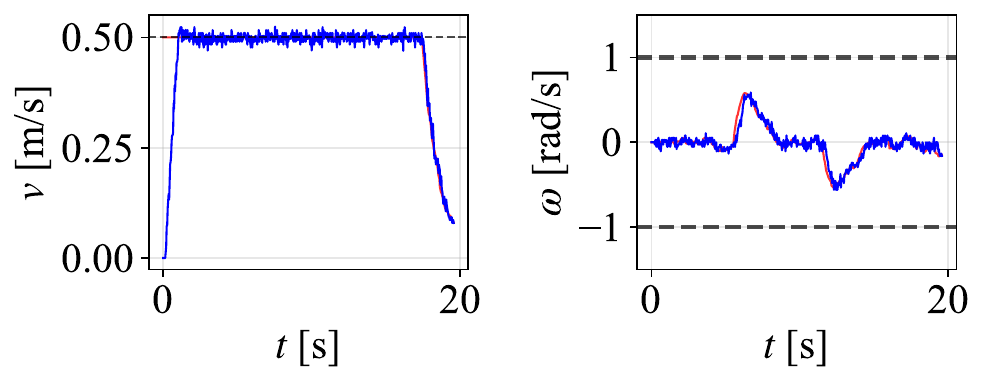}
        \caption{PP}
        \label{fig:vel_pp_pathA}
    \end{subfigure}
    \hfill
    \begin{subfigure}[b]{0.45\textwidth}
        \centering
        \includegraphics[width=\textwidth]{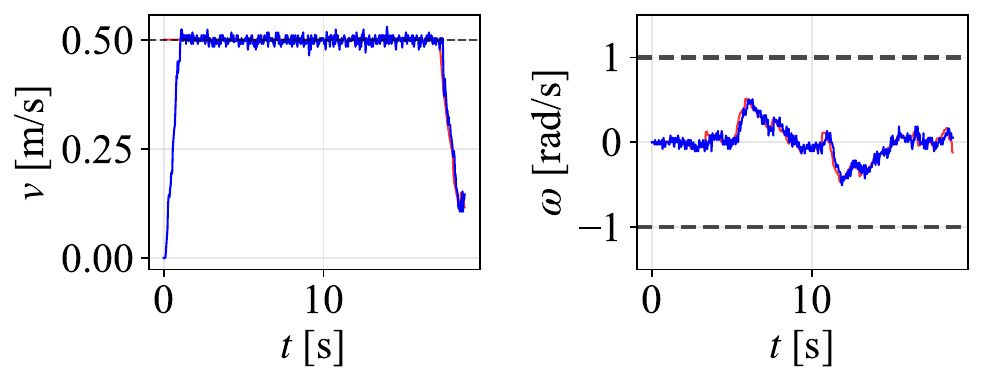}
        \caption{APP}
        \label{fig:vel_app_pathA}
    \end{subfigure}

    \vspace{0.4em}

    \begin{subfigure}[b]{0.45\textwidth}
        \centering
        \includegraphics[width=\textwidth]{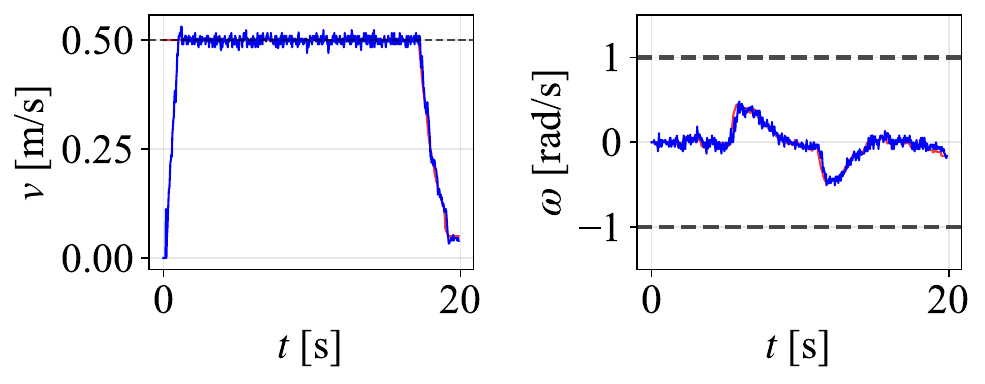}
        \caption{RPP}
        \label{fig:vel_rpp_pathA}
    \end{subfigure}
    \hfill
    \begin{subfigure}[b]{0.45\textwidth}
        \centering
        \includegraphics[width=\textwidth]{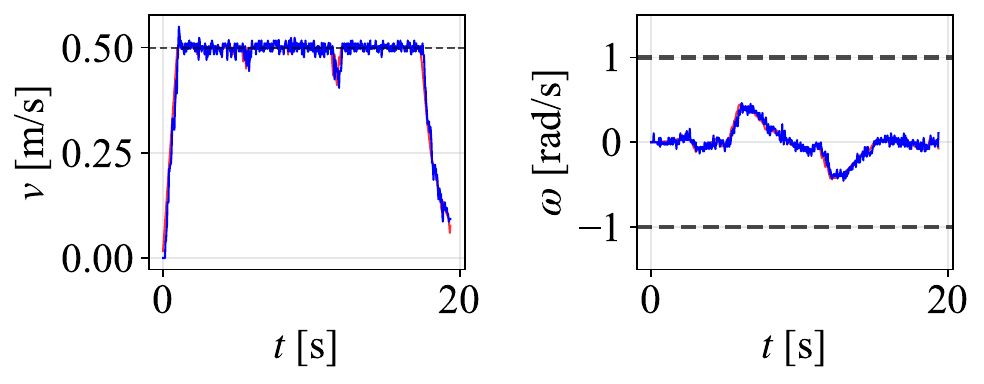}
        \caption{DWPP (Proposed)}
        \label{fig:vel_dwpp_pathA}
    \end{subfigure}

    \vspace{0.4em}

    \begin{subfigure}[b]{0.45\textwidth}
        \centering
        \color{red}
        \captionsetup{font+=revred}
        \includegraphics[width=\textwidth]{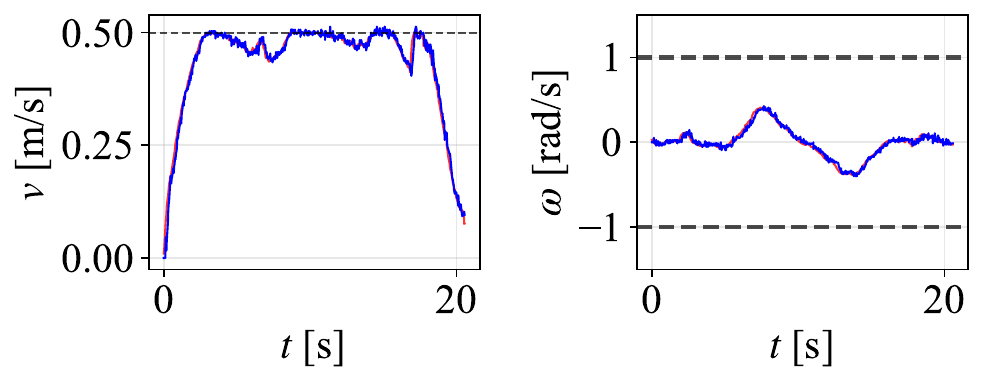}
        \caption{MPPI}
        \label{fig:vel_mppi_pathA}
    \end{subfigure}

    \caption{\rev{Experimental results in Path A: (a) paths tracked in all five trials of each controller, shown in separate panels, and (b)--(f) velocity profiles of a representative trial of each controller, where black dashed lines show the velocity limits.}}
    \label{fig:results_all_pathA}
\end{figure*}

\begin{figure*}[tbp]
    \centering
    
    % ---- 凡例（横長） ----
    \begin{subfigure}[b]{0.80\textwidth}
        \centering
        \includegraphics[width=\textwidth]{path_comparison_label.png}
    \end{subfigure}
    
    % 1. 経路比較
    \begin{subfigure}[b]{\textwidth}
        \centering
        \includegraphics[width=\textwidth, height=6.5cm, keepaspectratio]{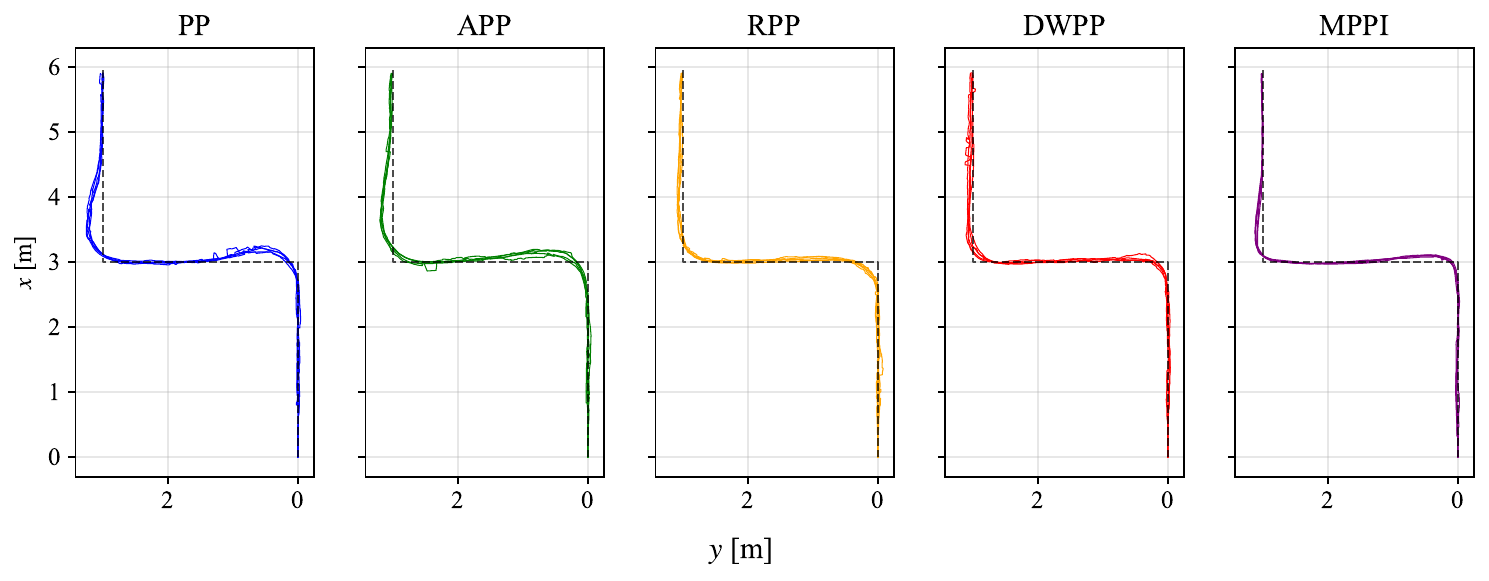}
        \caption{Path comparison}
        \label{fig:path_comparison_pathB}
    \end{subfigure}
    
    \vspace{0.5em}

    % ---- 凡例（横長） ----
    \begin{subfigure}[b]{0.45\textwidth}
        \centering
        \includegraphics[width=\textwidth]{velocity_legend_horizontal.png}
    \end{subfigure}
    
    \vspace{0.5em}

    % 2. 速度プロファイル群
    \begin{subfigure}[b]{0.45\textwidth}
        \centering
        \includegraphics[width=\textwidth]{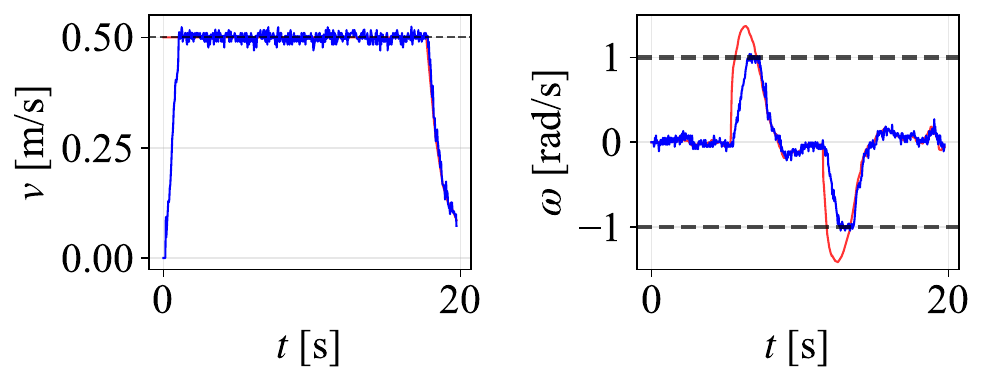}
        \caption{PP}
        \label{fig:vel_pp_pathB}
    \end{subfigure}
    \hfill
    \begin{subfigure}[b]{0.45\textwidth}
        \centering
        \includegraphics[width=\textwidth]{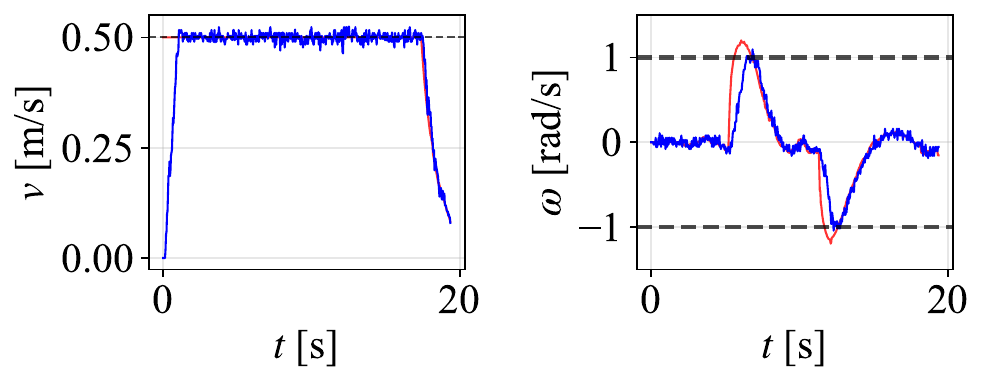}
        \caption{APP}
        \label{fig:vel_app_pathB}
    \end{subfigure}

    \vspace{0.4em}

    \begin{subfigure}[b]{0.45\textwidth}
        \centering
        \includegraphics[width=\textwidth]{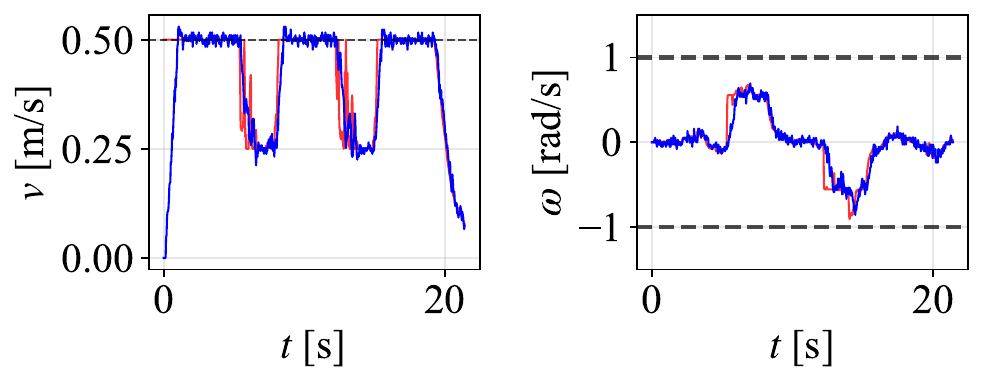}
        \caption{RPP}
        \label{fig:vel_rpp_pathB}
    \end{subfigure}
    \hfill
    \begin{subfigure}[b]{0.45\textwidth}
        \centering
        \includegraphics[width=\textwidth]{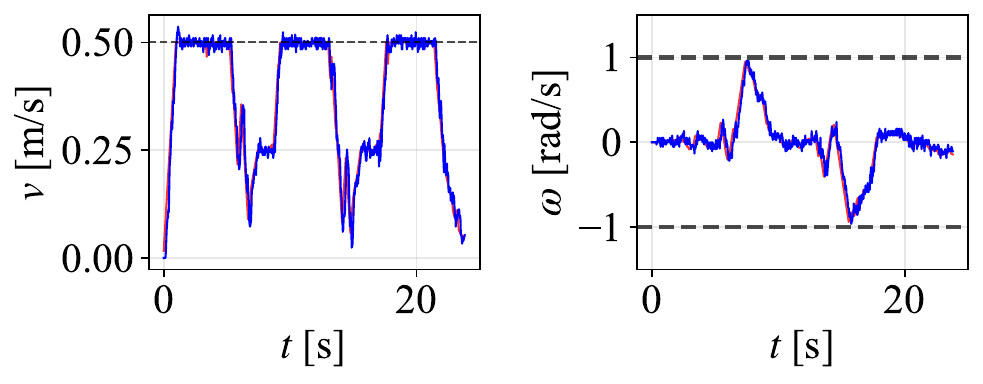}
        \caption{DWPP (Proposed)}
        \label{fig:vel_dwpp_pathB}
    \end{subfigure}

    \vspace{0.4em}

    \begin{subfigure}[b]{0.45\textwidth}
        \centering
        \color{red}
        \captionsetup{font+=revred}
        \includegraphics[width=\textwidth]{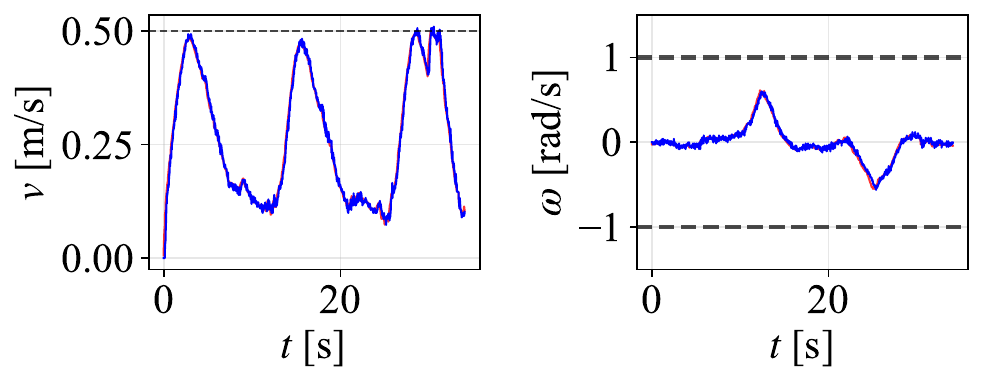}
        \caption{MPPI}
        \label{fig:vel_mppi_pathB}
    \end{subfigure}

    \caption{\rev{Experimental results in Path B: (a) paths tracked in all five trials of each controller, shown in separate panels, and (b)--(f) velocity profiles of a representative trial of each controller, where black dashed lines show the velocity limits.}}
    \label{fig:results_all_pathB}
\end{figure*}

\begin{figure*}[tbp]
    \centering
    
    % ---- 凡例（横長） ----
    \begin{subfigure}[b]{0.80\textwidth}
        \centering
        \includegraphics[width=\textwidth]{path_comparison_label.png}
    \end{subfigure}
    
    % 1. 経路比較
    \begin{subfigure}[b]{0.8\textwidth}
        \centering
        \includegraphics[width=\textwidth, height=6.5cm, keepaspectratio]{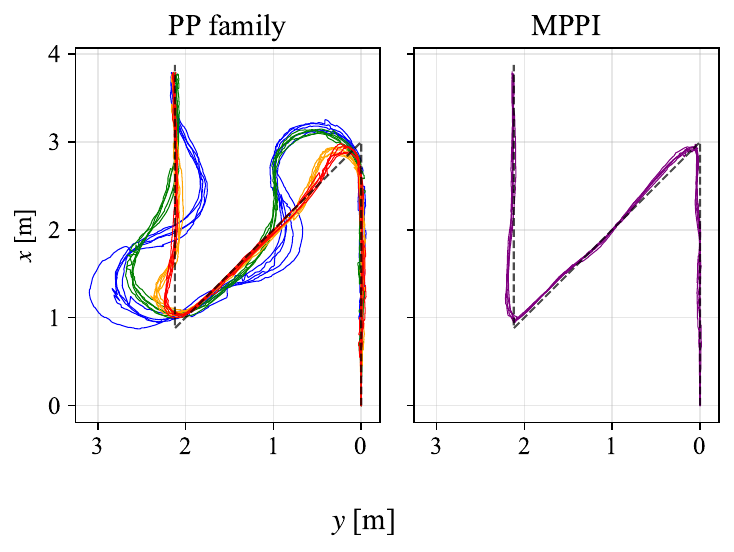}
        \caption{Path comparison}
        \label{fig:path_comparison_pathC}
    \end{subfigure}
    
    \vspace{0.5em}

    % ---- 凡例（横長） ----
    \begin{subfigure}[b]{0.45\textwidth}
        \centering
        \includegraphics[width=\textwidth]{velocity_legend_horizontal.png}
    \end{subfigure}
    
    \vspace{0.5em}

    % 2. 速度プロファイル群（共通の説明はメインキャプションへ）
    \begin{subfigure}[b]{0.45\textwidth}
        \centering
        \includegraphics[width=\textwidth]{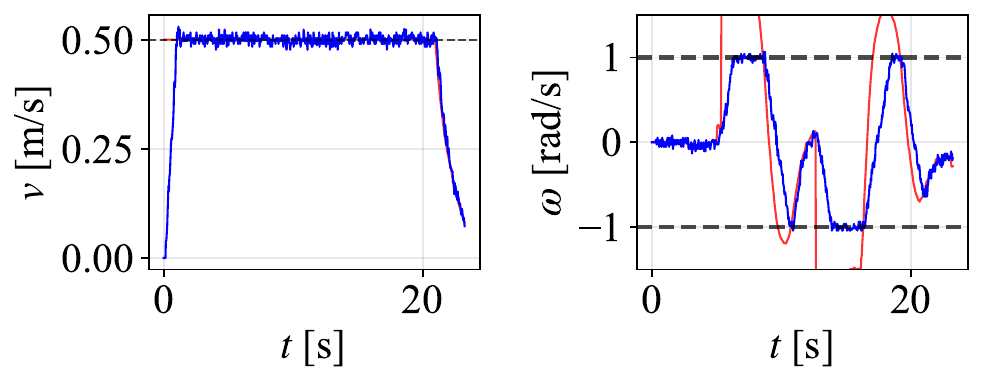}
        \caption{PP}
        \label{fig:vel_pp_pathC}
    \end{subfigure}
    \hfill
    \begin{subfigure}[b]{0.45\textwidth}
        \centering
        \includegraphics[width=\textwidth]{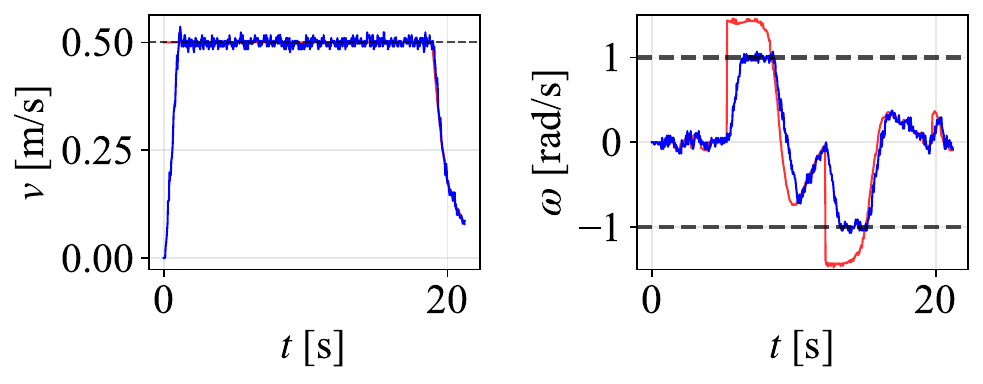}
        \caption{APP}
        \label{fig:vel_app_pathC}
    \end{subfigure}

    \vspace{0.4em}

    \begin{subfigure}[b]{0.45\textwidth}
        \centering
        \includegraphics[width=\textwidth]{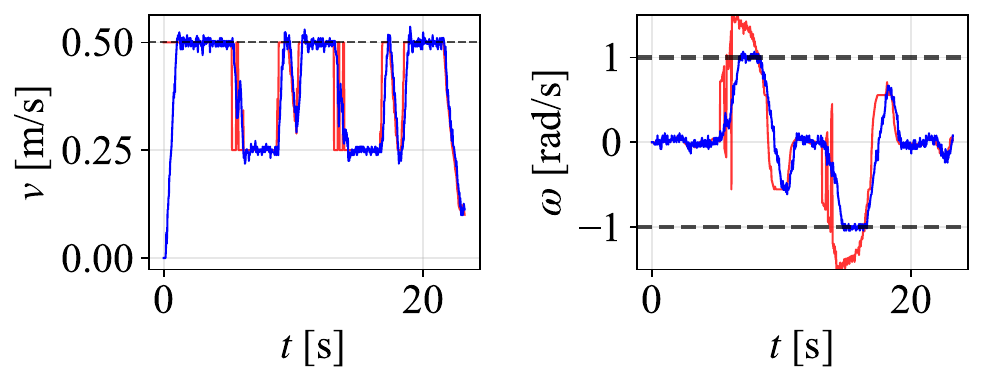}
        \caption{RPP}
        \label{fig:vel_rpp_pathC}
    \end{subfigure}
    \hfill
    \begin{subfigure}[b]{0.45\textwidth}
        \centering
        \includegraphics[width=\textwidth]{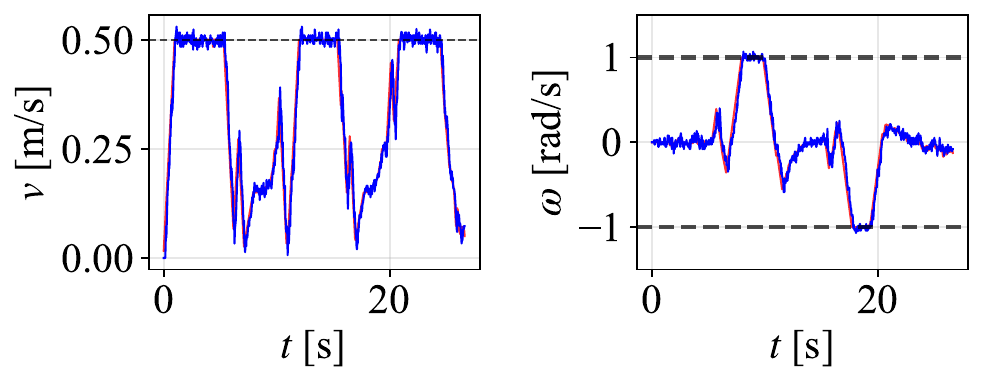}
        \caption{DWPP (Proposed)}
        \label{fig:vel_dwpp_pathC}
    \end{subfigure}

    \vspace{0.4em}

    \begin{subfigure}[b]{0.45\textwidth}
        \centering
        \color{red}
        \captionsetup{font+=revred}
        \includegraphics[width=\textwidth]{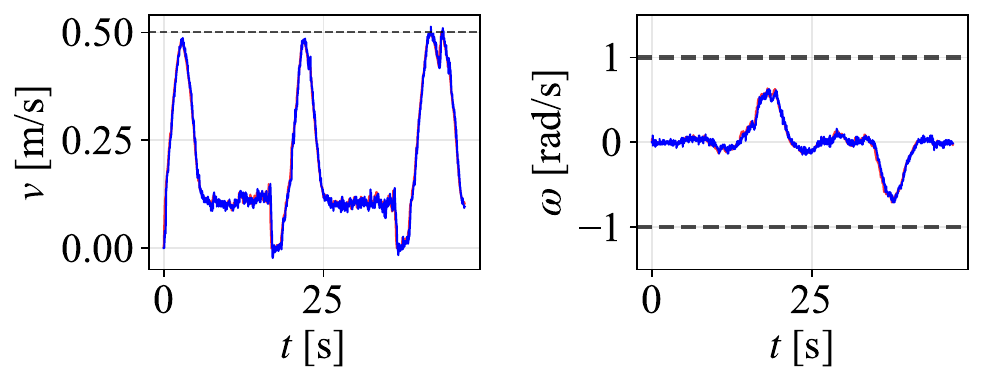}
        \caption{MPPI}
        \label{fig:vel_mppi_pathC}
    \end{subfigure}

    \caption{\rev{Experimental results in Path C: (a) paths tracked in all five trials (left: PP, APP, RPP, and DWPP, right: MPPI), and (b)--(f) velocity profiles of a representative trial of each controller, where black dashed lines show the velocity limits.}}
    \label{fig:results_all_pathC}
\end{figure*}

Tables~\ref{tab:violation}--\ref{tab:travel_time} summarize the quantitative results for each method; the reported values represent the means and standard deviations computed over five runs. Specifically, Table~\ref{tab:violation} lists the percentage of velocity and acceleration constraint violations, whereas Tables~\ref{tab:mean_error} and \ref{tab:max_error} present the means and maximum cross-track errors, respectively. The travel times required to reach the goal are listed in Table~\ref{tab:travel_time}. Furthermore, Figs.~\ref{fig:results_all_pathA}--\ref{fig:results_all_pathC} illustrate the path tracking performance and velocity profiles obtained with each controller. The paths are plotted for all trials, whereas the velocity profiles show a representative trial for each controller. The experimental behavior is demonstrated in the supplementary video.\footnote{\rev{\url{https://youtu.be/1H5yHKX0EAg}}}

\begin{table}[tbp]
\centering

% --- Table 1: Constraint Violation Ratio ---
\caption{Constraint violation ratio [\%] (Mean $\pm$ SD)}
\label{tab:violation}
\begin{tabular}{l c c c c c}
\hline
Path  & PP   & APP  & RPP  & DWPP & \rev{MPPI} \\
\hline
Path A & $3.3 \pm 1.4$  & $2.0 \pm 1.1$  & $1.4 \pm 0.5$  & $0.0 \pm 0.0$ & \rev{$0.1 \pm 0.1$} \\
Path B & $22.4 \pm 2.7$ & $13.9 \pm 1.7$ & $10.1 \pm 0.8$ & $0.0 \pm 0.0$ & \rev{$0.2 \pm 0.1$} \\
Path C & $55.0 \pm 2.8$ & $34.6 \pm 0.6$ & $32.1 \pm 2.1$ & $0.0 \pm 0.0$ & \rev{$0.4 \pm 0.1$} \\
\hline
\end{tabular}

\vspace{1.5em}

% --- Table 2: Mean Cross Track Error ---
\caption{Mean cross track error [m] (Mean $\pm$ SD)}
\label{tab:mean_error}
\begin{tabular}{l c c c c c}
\hline
Path  & PP   & APP  & RPP  & DWPP & \rev{MPPI} \\
\hline
Path A & $0.03 \pm \rev{0.00}$ & $0.03 \pm \rev{0.00}$ & $0.03 \pm 0.00$ & $0.03 \pm \rev{0.00}$ & \rev{$0.04 \pm 0.00$} \\
Path B & $0.05 \pm \rev{0.00}$ & $0.06 \pm \rev{0.00}$ & $0.04 \pm \rev{0.00}$ & $0.03 \pm 0.01$ & \rev{$0.03 \pm 0.00$} \\
Path C & $0.20 \pm \rev{0.02}$ & $0.14 \pm \rev{0.00}$ & $0.05 \pm \rev{0.00}$ & $0.03 \pm \rev{0.00}$ & \rev{$0.03 \pm 0.00$} \\
\hline
\end{tabular}

\vspace{1.5em}

% --- Table 3: Max Cross Track Error ---
\caption{Max cross track error [m] (Mean $\pm$ SD)}
\label{tab:max_error}
\begin{tabular}{l c c c c c}
\hline
Path  & PP   & APP  & RPP  & DWPP & \rev{MPPI} \\
\hline
Path A & $0.11 \pm 0.01$ & $0.11 \pm 0.01$ & $0.10 \pm 0.00$ & $0.10 \pm 0.01$ & \rev{$0.12 \pm 0.01$} \\
Path B & $0.24 \pm 0.01$ & $0.19 \pm 0.01$ & $0.15 \pm 0.02$ & $0.12 \pm 0.01$ & \rev{$0.11 \pm 0.01$} \\
Path C & $0.74 \pm 0.12$ & $0.60 \pm 0.02$ & $0.23 \pm 0.03$ & $0.13 \pm 0.02$ & \rev{$0.09 \pm 0.01$} \\
\hline
\end{tabular}

\vspace{1.5em}

% --- Table 4: Travel Time ---
\caption{Travel time [s] (Mean $\pm$ SD)}
\label{tab:travel_time}
\begin{tabular}{l c c c c c}
\hline
Path  & PP   & APP  & RPP  & DWPP & \rev{MPPI} \\
\hline
Path A & $19.6 \pm 0.3$ & $19.2 \pm 0.2$ & $19.4 \pm 0.3$ & $19.4 \pm 0.2$ & \rev{$20.5 \pm 0.3$} \\
Path B & $19.6 \pm 0.1$ & $19.2 \pm 0.1$ & $21.1 \pm 0.2$ & $22.8 \pm 0.9$ & \rev{$34.2 \pm 0.4$} \\
Path C & $24.0 \pm 0.9$ & $21.3 \pm 0.1$ & $23.8 \pm 0.7$ & $26.1 \pm 0.9$ & \rev{$47.1 \pm 0.5$} \\
\hline
\end{tabular}

\end{table}

% --- Figure Path A ---

% --- Figure Path B ---

% --- Figure Path C ---

As shown in Table~\ref{tab:violation}, \rev{PP, APP, and RPP} resulted in non-zero velocity and acceleration constraint violations; in contrast, DWPP produced no constraint violations for any path. \rev{MPPI also kept the violation ratio close to zero.}

\rev{Tables~\ref{tab:mean_error} and \ref{tab:max_error} indicate that the differences among the controllers grew as the path corner angle $\theta_{\mathrm{path}}$ increased. On Path~A, all controllers achieved similar small errors. On Paths~B and C, DWPP achieved the smallest mean and maximum errors in the PP family, with the largest margin on Path~C, and MPPI achieved accuracy comparable to DWPP. These results indicate that DWPP reduced the overall tracking error and the overshoot compared with PP, APP, and RPP on the paths that demand sharp turning.}

\rev{Table~\ref{tab:travel_time} shows that the travel time of DWPP was comparable to those of the other PP-family controllers on Path~A and longer on Paths~B and C.} \rev{MPPI required the longest travel time of all controllers because it reduced the velocity strongly while negotiating the corners (Figs.~\ref{fig:vel_mppi_pathA}--\ref{fig:vel_mppi_pathC}). The comparison with MPPI therefore illustrates the trade-off between the tracking accuracy and the travel time.}

\rev{Figures~\ref{fig:path_comparison_pathA}--\ref{fig:path_comparison_pathC} visualize this trend.} In particular, for Path~C with $\theta_{\mathrm{path}} = 135^\circ$, DWPP exhibited the smallest overshoot \rev{in the PP family}, as shown in Fig.~\ref{fig:path_comparison_pathC}. \rev{The tracked paths of MPPI also stayed close to the reference on every path.} These observations were consistent with the quantitative results reported in Tables~\ref{tab:mean_error} and \ref{tab:max_error}.

Figures~\ref{fig:vel_pp_pathA}--\ref{fig:vel_dwpp_pathC} show that PP, APP, and RPP generated velocity commands that were not always feasible, resulting in discrepancies between the commanded and executed velocities. In contrast, DWPP produced feasible velocity commands, and the commanded and executed velocities closely matched. \rev{The velocity commands of MPPI were feasible in almost all cycles, and its commanded and executed velocities also matched closely (Figs.~\ref{fig:vel_mppi_pathA}--\ref{fig:vel_mppi_pathC}).} \rev{These observations were consistent with the results shown in Table~\ref{tab:violation}.} 

\begin{revblock}
The per-cycle computation time measured on the same platform, pooled over all control cycles of five trials per path, was $0.94 \pm 0.52$~ms for PP, $0.99 \pm 0.51$~ms for APP, $1.00 \pm 0.50$~ms for RPP, $1.00 \pm 0.50$~ms for DWPP, and $7.60 \pm 1.07$~ms for MPPI (mean $\pm$ SD). The computation time of DWPP matched that of RPP and was within $0.06$~ms of that of PP, confirming that the closed-form $O(1)$ projection adds practically no overhead.
\end{revblock}

In addition, the velocity commands of PP and APP remained constant at the maximum velocity, whereas RPP and DWPP reduced the velocity at corners. Among these methods, DWPP exhibited a greater decrease in velocity. This behavior corresponded to the longer travel times observed for DWPP \rev{on Paths~B and C} in Table~\ref{tab:travel_time}. We hypothesized that this increase in travel time could be mitigated by using a longer lookahead distance, as a longer lookahead distance leads to a smaller curvature of the circular arc toward the lookahead point and thus makes velocity reduction less likely. However, using a longer lookahead distance is also expected to degrade the path-tracking accuracy. Therefore, \rev{Section~\ref{sec:sim_lookahead} analyzes} the effect of the lookahead distance on the path-tracking error and travel time of DWPP.

\subsection{\texorpdfstring{\rev{Confined Corridor Experiment}}{Confined Corridor Experiment}}
\label{sec:corridor}

\begin{figure}[tbp]
    \centering
    \color{red}
    \captionsetup{font+=revred}
    \captionsetup[sub]{font+=revred}
    \begin{subfigure}[b]{\linewidth}
        \centering
        \includegraphics[width=0.78\linewidth]{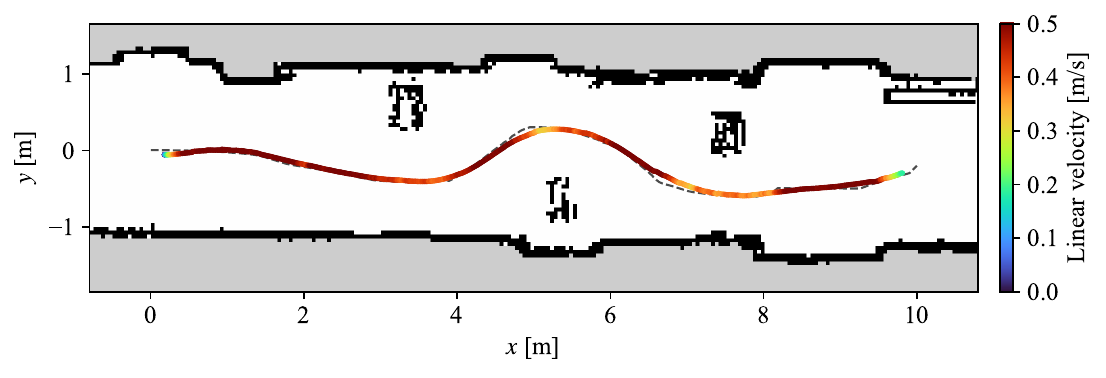}
        \caption{RPP}
        \label{fig:corridor_speed_rpp}
    \end{subfigure}\\[4pt]
    \begin{subfigure}[b]{\linewidth}
        \centering
        \includegraphics[width=0.78\linewidth]{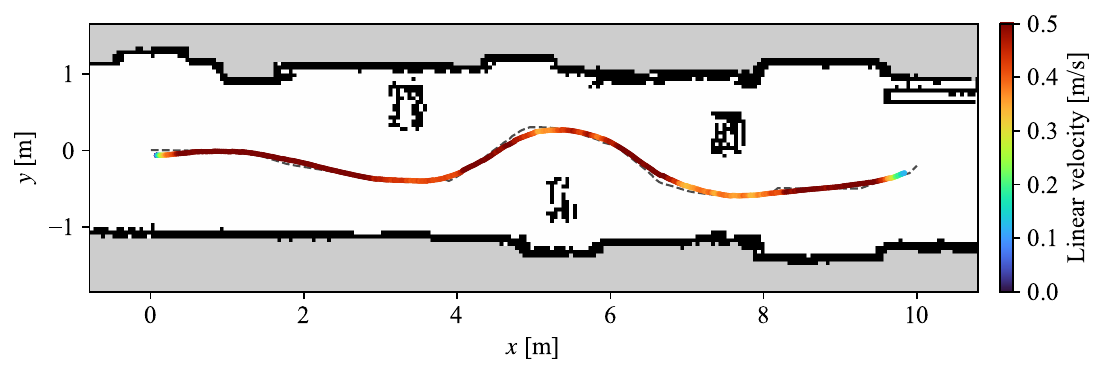}
        \caption{DWPP}
        \label{fig:corridor_speed_dwpp}
    \end{subfigure}
    \caption{Paths tracked in the confined corridor experiment for a representative trial (color bar from 0 to 0.5~m/s, red for fast and blue for slow).}
    \label{fig:corridor_speed}
\end{figure}

\begin{revblock}
\subsubsection{Setting}
\label{sec:corridor_setting}

The second experiment evaluates the behavior of RPP and DWPP in an obstacle environment in which the proximity heuristic of RPP is active, following the confined corridor experiment of the original RPP work \cite{macenski2023regulated}. As shown in Figure~\ref{fig:corridor_speed}, a corridor of about 2.0~m width was built, containing obstacles of approximately 1.0~m size placed on alternating sides, and the robot tracked a fixed reference path of 10~m length that weaves between the obstacles. The reference path was planned once by a Dijkstra-based global planner (NavFn of Nav2) and was sent to the controller in every trial so that all runs shared an identical tracking task. Its waypoints are spaced about $0.025$~m apart, following the grid resolution of the map. Both RPP and DWPP were run with the curvature and proximity heuristics enabled. The heuristic parameters were determined experimentally so that both heuristics function effectively in this environment, giving the curvature threshold $R_{\min} = 1.5$~m and the proximity threshold $d_{\mathrm{prox}} = 0.7$~m (gain $g_d = 1.0$). All other parameters were identical to Table~\ref{tab:exp_parameters}. The AMCL parameters were configured separately for this environment, and the per-update corrections had root-mean-square magnitudes of about $0.007$~m in position and $0.0048$~rad in heading at about 3~Hz. Each controller was tested in five trials. The complete parameter file used in this experiment is available online.\footnote{\rev{\url{https://github.com/decwest/dwpp_test_simulation/blob/main/params/confined_corridor_experiment.yaml}}}

\subsubsection{Results}
\label{sec:corridor_result}

\begin{table}[tbp]
\centering
\color{red}
\captionsetup{font+=revred}
\captionsetup[sub]{font+=revred}
\caption{Results of the confined corridor experiment (Mean $\pm$ SD)}
\label{tab:corridor}
\begin{tabular}{l c c}
\hline
Metric & RPP & DWPP \\
\hline
Constraint violation ratio [\%] & $6.5 \pm 1.6$ & $0.0 \pm 0.0$ \\
Mean cross track error [m] & $0.03 \pm 0.00$ & $0.03 \pm 0.00$ \\
Max cross track error [m] & $0.11 \pm 0.00$ & $0.11 \pm 0.01$ \\
Travel time [s] & $24.0 \pm 0.2$ & $24.1 \pm 0.1$ \\
\hline
\end{tabular}
\end{table}

\begin{figure}[tbp]
    \centering
    \color{red}
    \captionsetup{font+=revred}
    \captionsetup[sub]{font+=revred}
    \includegraphics[width=0.95\linewidth]{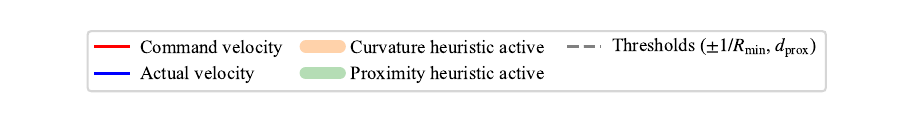}\\[2pt]
    \begin{subfigure}[b]{\linewidth}
        \centering
        \begin{subsubfigure}[b]{0.32\linewidth}
            \centering
            \includegraphics[width=\linewidth]{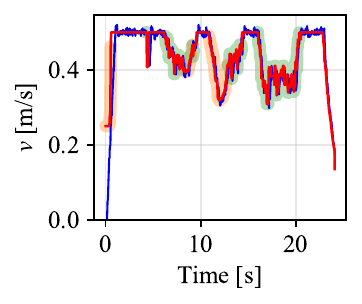}
            \caption{Velocity profile}
        \end{subsubfigure}
        \hfill
        \begin{subsubfigure}[b]{0.32\linewidth}
            \centering
            \includegraphics[width=\linewidth]{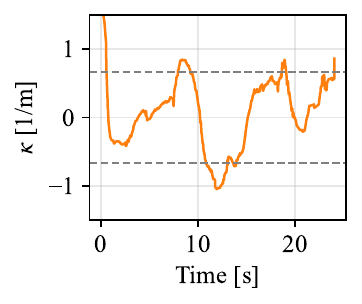}
            \caption{Commanded curvature}
        \end{subsubfigure}
        \hfill
        \begin{subsubfigure}[b]{0.32\linewidth}
            \centering
            \includegraphics[width=\linewidth]{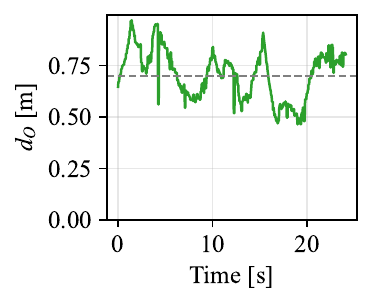}
            \caption{Obstacle distance}
        \end{subsubfigure}
        \caption{RPP}
        \label{fig:corridor_vcmd_rpp}
    \end{subfigure}\\[6pt]
    \begin{subfigure}[b]{\linewidth}
        \centering
        \begin{subsubfigure}[b]{0.32\linewidth}
            \centering
            \includegraphics[width=\linewidth]{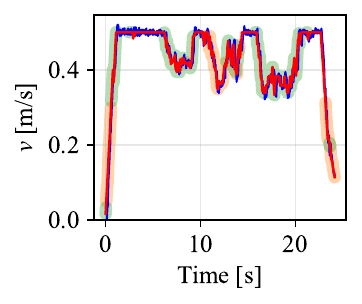}
            \caption{Velocity profile}
        \end{subsubfigure}
        \hfill
        \begin{subsubfigure}[b]{0.32\linewidth}
            \centering
            \includegraphics[width=\linewidth]{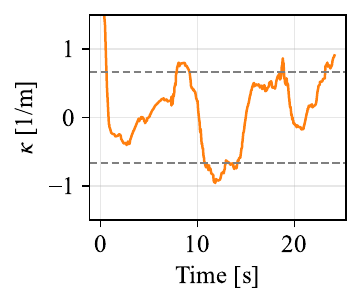}
            \caption{Commanded curvature}
        \end{subsubfigure}
        \hfill
        \begin{subsubfigure}[b]{0.32\linewidth}
            \centering
            \includegraphics[width=\linewidth]{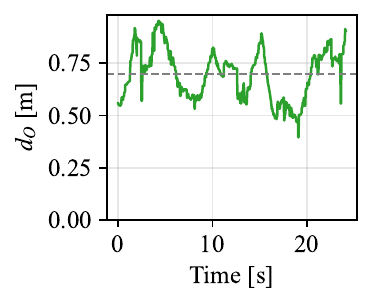}
            \caption{Obstacle distance}
        \end{subsubfigure}
        \caption{DWPP}
        \label{fig:corridor_vcmd_dwpp}
    \end{subfigure}
    \caption{Time series of a representative trial in the confined corridor experiment.}
    \label{fig:corridor_vcmd}
\end{figure}

Table~\ref{tab:corridor} summarizes the results. For a representative trial of each controller, Fig.~\ref{fig:corridor_speed} shows the tracked path colored by the measured speed, and Fig.~\ref{fig:corridor_vcmd} shows the time series of the linear velocity, the commanded curvature, and the obstacle distance. The experimental behavior is demonstrated in the supplementary video.\footnote{\rev{\url{https://youtu.be/INkE8k3Xw-U}}}

The central observation is that DWPP fully inherits the function of the RPP’s heuristics. As the robot approached each obstacle and negotiated the sharp curves between the obstacles, both controllers reduced the linear velocity according to the regulated velocity of the proximity and curvature heuristics (Fig.~\ref{fig:corridor_speed}), and the differences between the two controllers in the tracking errors and the travel time were within the trial-to-trial variability (Table~\ref{tab:corridor}). DWPP therefore provides the same protective slowdown behavior near obstacles and on sharp curves as RPP.

The two controllers differed only in the feasibility of the commands. RPP violated the acceleration constraints in a small fraction of the control cycles, concentrated at the onset and release of the heuristic regulation, as visible in Fig.~\ref{fig:corridor_vcmd}. RPP applies the regulated velocity directly as the command, regardless of the velocity change required from the current velocity. DWPP, by contrast, tracked the same regulated targets through feasible ramps within the velocity and acceleration limits and produced exactly zero violations, while the differences in the tracking errors and the travel time stayed within the trial-to-trial variability.

\end{revblock}

\section{\texorpdfstring{\rev{Simulation}}{Simulation}}
\label{sec:sim}

\begin{revblock}
This section complements the real-robot experiments with simulation analyses that would be impractical to perform exhaustively on hardware. The analyses cover parameter sweeps of the baseline controllers, a comparison under matched travel time, and the effects of the acceleration limits, localization noise, and the lookahead distance.

\subsection{Common Setting}
\label{sec:sim_common}

The simulation studies use a lightweight kinematic simulator that implements exactly the model of Section~\ref{subsec:problem_formulation}. The robot follows the unicycle model of Eq.~\eqref{eq:unicycle}, integrated with exact circular-arc steps at the control period, and the controller receives the true robot pose. Commands inside the admissible set of Eqs.~\eqref{eq:vel_limits} and \eqref{eq:acc_limits} are realized exactly within one control period. Commands outside the admissible set (which only PP, APP, and RPP produce) are clipped to the configured velocity and acceleration limits before execution, as on the real platform. Localization noise is injected only where explicitly stated (Sections~\ref{sec:sim_noise} and \ref{sec:sim_lookahead}). Unless otherwise stated, the velocity and acceleration limits, the control period, and the controller parameters are identical to those of the real-robot experiments (Table~\ref{tab:exp_parameters}). The simulation studies use only Path~C (Fig.~\ref{fig:pathC}), on which the differences among the controllers were the largest in the real-robot experiments. The simulator implementation is publicly available.\footnote{\rev{\url{https://github.com/decwest/dwpp}}}

\subsection{Effect of APP Parameter}
\label{sec:sim_app}

\begin{table}[!tbp]
\centering
\color{red}
\captionsetup{font+=revred}
\captionsetup[sub]{font+=revred}
\caption{Effect of the APP lookahead-time gain $l_t$ on Path~C.}
\label{tab:sim_app}
\scriptsize
\setlength{\tabcolsep}{3pt}
\begin{subtable}[t]{0.48\linewidth}
\centering
\caption{APP}
\begin{tabular}{rrrrr}
\toprule
$l_t$ [s] & Viol.\ [\%] & Mean [m] & Max [m] & Time [s] \\
\midrule
0.6 & 31.3 & 1.04 & 3.37 & 42.4 \\
0.8 & 68.2 & 0.29 & 0.76 & 25.3 \\
1.0 & 74.8 & 0.29 & 0.77 & 24.8 \\
1.2 & 49.1 & 0.17 & 0.62 & 21.5 \\
1.4 & 31.3 & 0.13 & 0.55 & 20.0 \\
1.6 & 26.6 & 0.14 & 0.50 & 19.3 \\
1.8 & 23.1 & 0.16 & 0.52 & 18.6 \\
2.0 & 0.6 & 0.19 & 0.57 & 18.1 \\
2.2 & 0.6 & 0.23 & 0.69 & 17.8 \\
2.4 & 0.6 & 0.27 & 0.79 & 17.5 \\
2.6 & 0.6 & 0.29 & 0.87 & 17.2 \\
2.8 & 0.6 & 0.31 & 0.90 & 16.7 \\
3.0 & 0.6 & 0.30 & 0.87 & 16.0 \\
\bottomrule
\end{tabular}
\end{subtable}
\hfill
\begin{subtable}[t]{0.48\linewidth}
\centering
\caption{APP+DW}
\begin{tabular}{rrrrr}
\toprule
$l_t$ [s] & Viol.\ [\%] & Mean [m] & Max [m] & Time [s] \\
\midrule
0.6 & 0.0 & 0.04 & 0.20 & 26.3 \\
0.8 & 0.0 & 0.02 & 0.14 & 24.2 \\
1.0 & 0.0 & 0.02 & 0.12 & 24.1 \\
1.2 & 0.0 & 0.02 & 0.12 & 23.8 \\
1.4 & 0.0 & 0.03 & 0.12 & 23.3 \\
1.6 & 0.0 & 0.04 & 0.17 & 23.6 \\
1.8 & 0.0 & 0.07 & 0.24 & 23.9 \\
2.0 & 0.0 & 0.11 & 0.36 & 24.2 \\
2.2 & 0.0 & 0.18 & 0.66 & 25.3 \\
2.4 & 0.0 & 0.20 & 0.66 & 26.8 \\
2.6 & 0.0 & 0.25 & 0.70 & 21.6 \\
2.8 & 0.0 & 0.22 & 0.77 & 19.8 \\
3.0 & 0.0 & 0.19 & 0.86 & 18.7 \\
\bottomrule
\end{tabular}
\end{subtable}
\end{table}

\begin{figure}[!tbp]
    \centering
    \color{red}
    \captionsetup{font+=revred}
    \captionsetup[sub]{font+=revred}
    \begin{subfigure}[b]{0.42\linewidth}
        \centering
        \includegraphics[width=\linewidth]{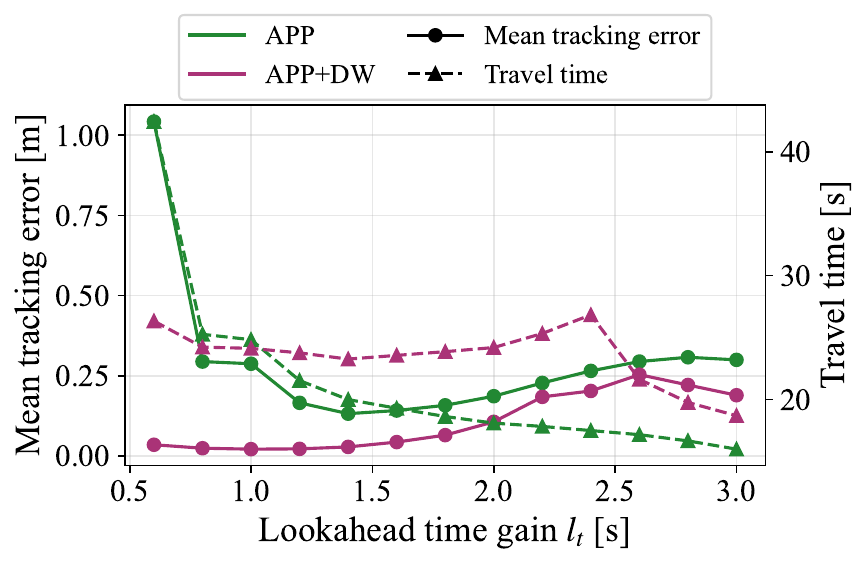}
        \caption{Mean tracking error and travel time vs.\ $l_t$}
    \end{subfigure}
    \hfill
    \begin{subfigure}[b]{0.20\linewidth}
        \centering
        \includegraphics[width=\linewidth]{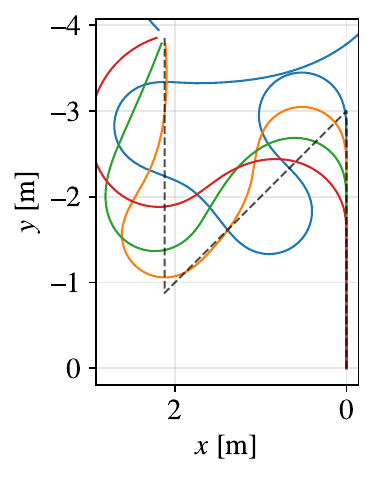}
        \caption{Tracked paths (APP)}
    \end{subfigure}
    \hfill
    \begin{subfigure}[b]{0.20\linewidth}
        \centering
        \includegraphics[width=\linewidth]{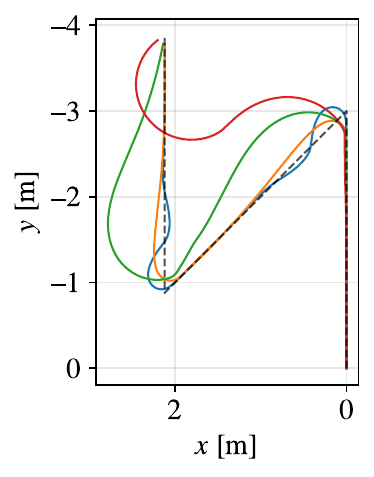}
        \caption{Tracked paths (APP+DW)}
    \end{subfigure}\\[0.2em]
    \includegraphics[width=0.75\linewidth]{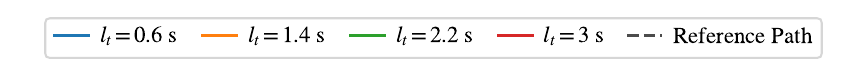}
    \caption{Effect of the APP lookahead-time gain $l_t$ on Path~C.}
    \label{fig:sim_app_sweep}
\end{figure}

To examine whether the conclusions depend on the tuning of the baselines, the lookahead-time gain $l_t$ of APP was swept over $0.6$--$3.0$~s in steps of $0.2$~s on Path~C, comparing plain APP with the same configuration under the dynamic-window command selection (denoted APP+DW). The curvature heuristic remains disabled so that only the effect of the dynamic window is isolated. In this study, the upper lookahead bound is relaxed to $L_{\max} = 2.0$~m so that the velocity-scaled lookahead is not saturated within the swept range, with the lower bound kept at $L_{\min} = 0.3$~m. Table~\ref{tab:sim_app} reports the four metrics of Section~\ref{sec:exp_common} for every setting, and Fig.~\ref{fig:sim_app_sweep} shows the mean tracking error and the travel time as functions of $l_t$, together with the tracked paths for representative settings.

Plain APP violated the acceleration constraints at every examined $l_t$, whereas APP+DW produced exactly zero violations in all 13 settings.

APP+DW also tracked the path more accurately throughout the sweep. At every value of $l_t$, both the mean and the maximum tracking errors of APP+DW were smaller than those of plain APP, and the smallest errors attainable over the sweep were likewise smaller for APP+DW.

This accuracy comes at the cost of travel time. Compared at the same $l_t$, APP+DW tended to take longer, because the dynamic-window command selection reduces the velocity where the commanded curvature is sharp. A striking exception is $l_t = 0.6$~s, where plain APP took an extremely long time because its tracking overshot severely around the sharp corner. APP+DW instead decelerated at the high-curvature sections, mitigated the overshoot, and followed the path smoothly, as directly visible in Fig.~\ref{fig:sim_app_sweep}.

\FloatBarrier
\subsection{Effect of RPP Parameter}
\label{sec:sim_rpp}

\begin{table}[!tbp]
\centering
\color{red}
\captionsetup{font+=revred}
\captionsetup[sub]{font+=revred}
\caption{Effect of the curvature-heuristic threshold $R_{\min}$ on Path~C.}
\label{tab:sim_rpp}
\scriptsize
\setlength{\tabcolsep}{2pt}
\begin{subtable}[t]{0.48\linewidth}
\centering
\caption{RPP}
\begin{tabular}{rrrrr}
\toprule
$R_{\min}$ [m] & Viol.\ [\%] & Mean [m] & Max [m] & Time [s] \\
\midrule
0.2 & 31.3 & 0.13 & 0.55 & 20.0 \\
0.4 & 29.2 & 0.05 & 0.20 & 20.7 \\
0.6 & 28.5 & 0.04 & 0.18 & 21.4 \\
0.8 & 28.0 & 0.04 & 0.18 & 21.6 \\
1.0 & 27.3 & 0.04 & 0.18 & 21.8 \\
1.2 & 26.7 & 0.04 & 0.18 & 21.9 \\
1.4 & 26.8 & 0.04 & 0.19 & 22.1 \\
1.6 & 26.9 & 0.05 & 0.20 & 22.4 \\
1.8 & 27.2 & 0.05 & 0.22 & 22.9 \\
2.0 & 27.8 & 0.05 & 0.22 & 23.1 \\
\bottomrule
\end{tabular}
\end{subtable}
\hfill
\begin{subtable}[t]{0.48\linewidth}
\centering
\caption{RPP+DW}
\begin{tabular}{rrrrr}
\toprule
$R_{\min}$ [m] & Viol.\ [\%] & Mean [m] & Max [m] & Time [s] \\
\midrule
0.2 & 0.0 & 0.03 & 0.12 & 23.3 \\
0.4 & 0.0 & 0.03 & 0.12 & 23.3 \\
0.6 & 0.0 & 0.03 & 0.12 & 23.4 \\
0.8 & 0.0 & 0.03 & 0.11 & 23.6 \\
1.0 & 0.0 & 0.02 & 0.11 & 23.7 \\
1.2 & 0.0 & 0.02 & 0.11 & 23.8 \\
1.4 & 0.0 & 0.02 & 0.11 & 24.0 \\
1.6 & 0.0 & 0.02 & 0.11 & 24.1 \\
1.8 & 0.0 & 0.02 & 0.11 & 24.2 \\
2.0 & 0.0 & 0.02 & 0.11 & 24.2 \\
\bottomrule
\end{tabular}
\end{subtable}
\end{table}

\begin{figure}[!tbp]
    \centering
    \color{red}
    \captionsetup{font+=revred}
    \captionsetup[sub]{font+=revred}
    \begin{subfigure}[b]{0.42\linewidth}
        \centering
        \includegraphics[width=\linewidth]{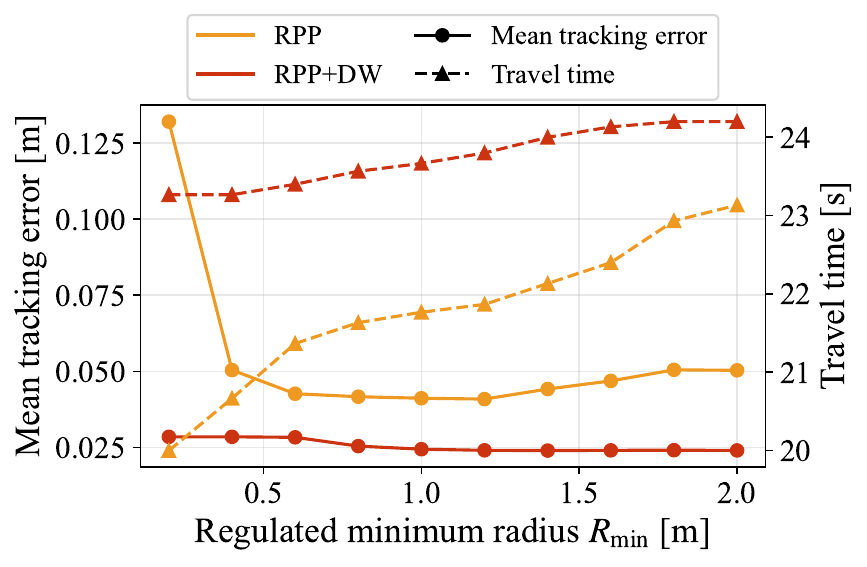}
        \caption{Mean tracking error and travel time vs.\ $R_{\min}$}
    \end{subfigure}
    \hfill
    \begin{subfigure}[b]{0.20\linewidth}
        \centering
        \includegraphics[width=\linewidth]{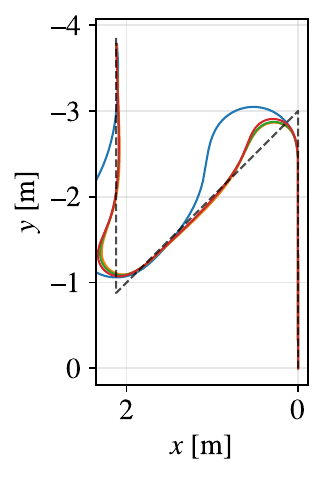}
        \caption{Tracked paths (RPP)}
    \end{subfigure}
    \hfill
    \begin{subfigure}[b]{0.20\linewidth}
        \centering
        \includegraphics[width=\linewidth]{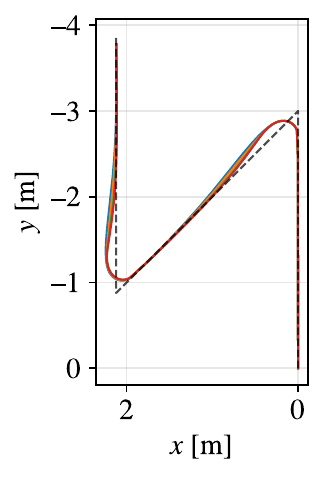}
        \caption{Tracked paths (RPP+DW)}
    \end{subfigure}\\[0.2em]
    \includegraphics[width=0.75\linewidth]{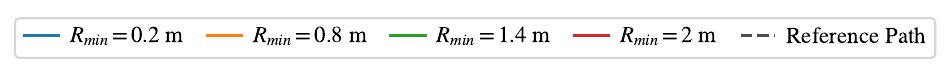}
    \caption{Effect of the curvature-heuristic threshold $R_{\min}$ of RPP and RPP+DW on Path~C.}
    \label{fig:sim_rpp_sweep}
\end{figure}

Analogously, the curvature-heuristic threshold $R_{\min}$ was swept over $0.2$--$2.0$~m in steps of $0.2$~m on Path~C, comparing plain RPP with the same configuration under the dynamic-window command selection (denoted RPP+DW), which corresponds to the proposed DWPP. Note that both controllers use the adaptive lookahead distance with the parameters of Table~\ref{tab:exp_parameters}. Table~\ref{tab:sim_rpp} reports the four metrics for every setting, and Fig.~\ref{fig:sim_rpp_sweep} shows the mean tracking error and the travel time as functions of $R_{\min}$, together with the tracked paths for representative settings.

RPP violated the acceleration constraints at every examined $R_{\min}$, whereas RPP+DW produced exactly zero violations in all 10 settings.

RPP+DW also tracked the path more accurately throughout the sweep. At every value of $R_{\min}$, both the mean and the maximum tracking errors of RPP+DW were smaller than those of RPP, and the smallest errors attainable over the sweep were likewise smaller for RPP+DW.

This accuracy comes at the cost of travel time. Compared at the same $R_{\min}$, RPP+DW tended to take longer than RPP.

\end{revblock}

\begin{revblock}
\subsection{Comparison under Same Travel Time}
\label{sec:sim_isotime}

\begin{table}[!tbp]
\centering
\color{red}
\captionsetup{font+=revred}
\captionsetup[sub]{font+=revred}
\caption{Comparison under matched travel time on Path~C.}
\label{tab:sim_isotime}
\footnotesize
\setlength{\tabcolsep}{3pt}
\begin{tabular}{lrrrrrr}
\toprule
Controller & $v_{\max}$ [m/s] & $R_{\min}$ [m] & Viol.\ [\%] & Mean [m] & Max [m] & Time [s] \\
\midrule
PP & 0.424 & -- & 26.0 & 0.10 & 0.46 & 23.2 \\
APP & 0.426 & -- & 27.8 & 0.10 & 0.47 & 23.3 \\
RPP & 0.500 & 2.0 & 27.8 & 0.05 & 0.22 & 23.1 \\
DWPP & 0.500 & 0.4 & 0.0 & 0.03 & 0.12 & 23.3 \\
\bottomrule
\end{tabular}
\end{table}

The travel-time increase of DWPP observed in Section~\ref{sec:experiment} raises the question of how the methods compare when the time--accuracy trade-off is fixed. Among the sweep conditions of Section~\ref{sec:sim_rpp}, the pair of RPP and DWPP settings with the closest travel times was selected, and $v_{\max}$ of PP and APP was reduced until their travel time on Path~C matched that of the selected DWPP setting within $\pm 1\%$. Table~\ref{tab:sim_isotime} summarizes the comparison.

In summary, even under the matched travel time, DWPP performed best in the PP family in both the violation ratio and the tracking error. PP, APP, and RPP still violated the acceleration constraints, whereas DWPP maintained zero violations. The mean and maximum errors of DWPP also remained smaller than those of PP, APP, and RPP.
\end{revblock}

\FloatBarrier
\begin{revblock}
\subsection{Effect of the Acceleration Limits}
\label{sec:sim_accel}

\begin{figure}[!tbp]
    \centering
    \color{red}
    \captionsetup{font+=revred}
    \includegraphics[width=0.5\linewidth]{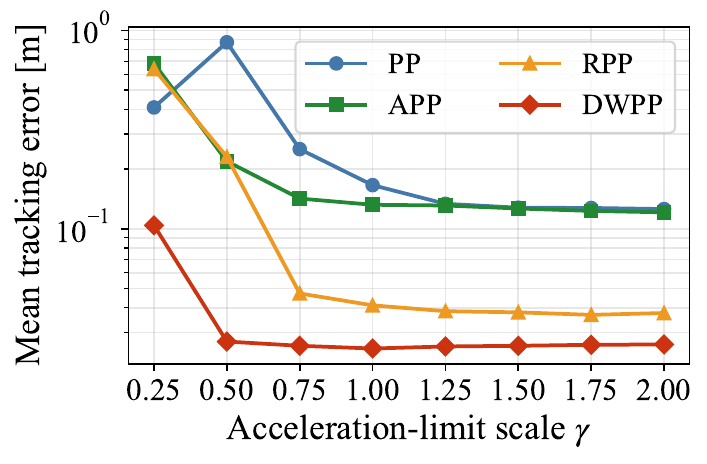}
    \caption{Effect of scaling the acceleration limits by $\gamma$ on Path~C.}
    \label{fig:sim_accel}
\end{figure}

To examine the sensitivity of the controllers to the configured acceleration limits, the four acceleration and deceleration limits of Table~\ref{tab:exp_parameters} were scaled by a common factor $\gamma$ swept over $0.25$--$2.0$ in steps of $0.25$ on Path~C, with $\gamma = 1$ corresponding to the configuration of the preceding sections and all other parameters unchanged. Figure~\ref{fig:sim_accel} shows the mean tracking error of every controller as a function of $\gamma$. Whereas the errors of PP, APP, and RPP grew sharply as the limits shrank, the error of DWPP stayed nearly unchanged for $\gamma \ge 0.5$ and remained the smallest at every scale, which shows that DWPP is the least sensitive to the acceleration limits.
\end{revblock}

\FloatBarrier
\begin{revblock}
\subsection{Effect of Localization Noise}
\label{sec:sim_noise}

\begin{figure}[!tbp]
    \centering
    \color{red}
    \captionsetup{font+=revred}
    \captionsetup[sub]{font+=revred}
    \includegraphics[width=0.65\linewidth]{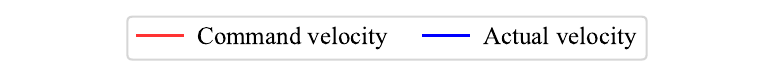}\\[0.2em]
    \begin{subfigure}[b]{0.45\linewidth}
        \centering
        \includegraphics[width=\linewidth]{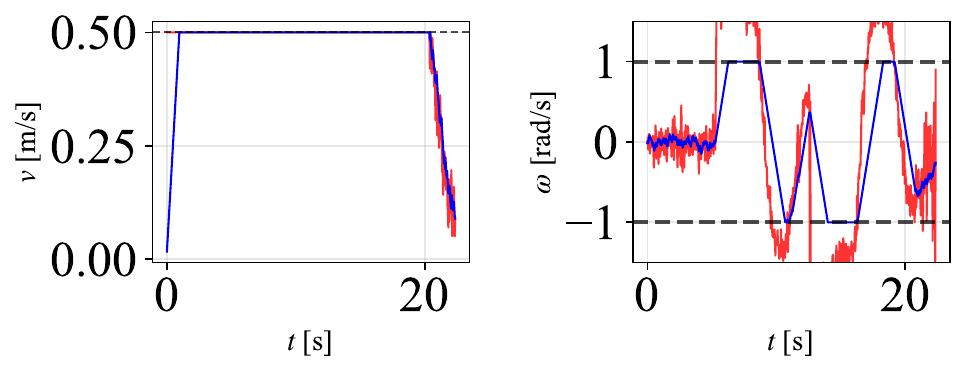}
        \caption{PP}
    \end{subfigure}
    \hfill
    \begin{subfigure}[b]{0.45\linewidth}
        \centering
        \includegraphics[width=\linewidth]{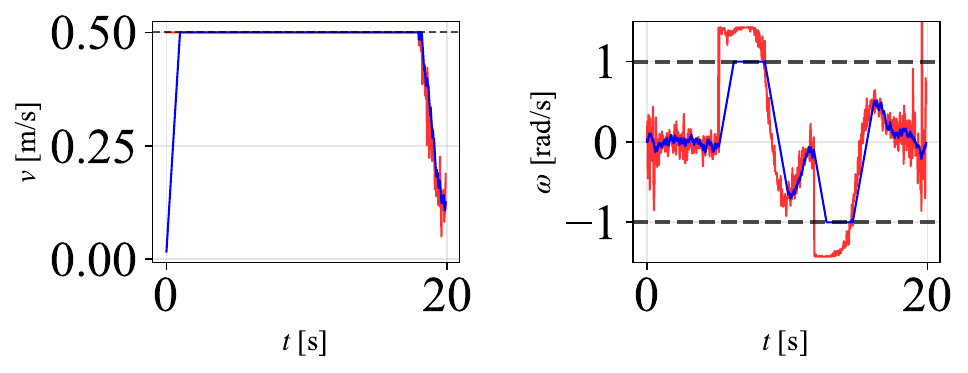}
        \caption{APP}
    \end{subfigure}\\[0.2em]
    \begin{subfigure}[b]{0.45\linewidth}
        \centering
        \includegraphics[width=\linewidth]{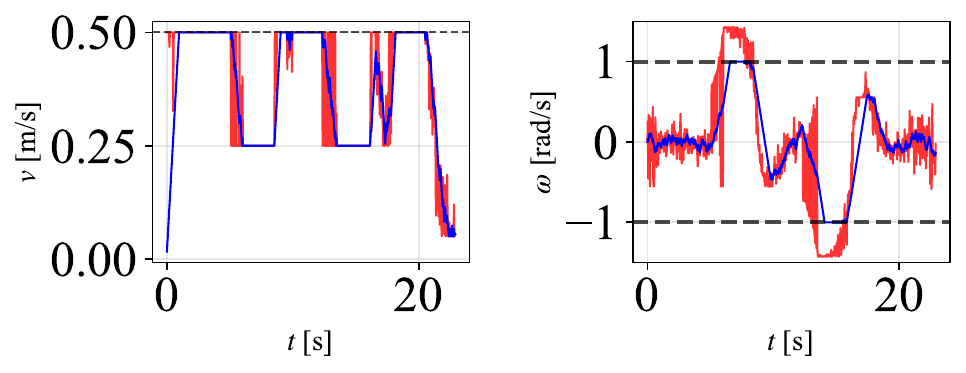}
        \caption{RPP}
    \end{subfigure}
    \hfill
    \begin{subfigure}[b]{0.45\linewidth}
        \centering
        \includegraphics[width=\linewidth]{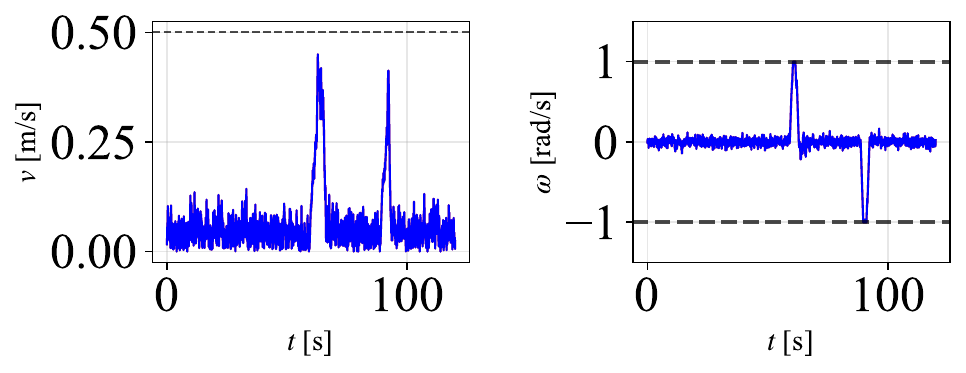}
        \caption{DWPP}
    \end{subfigure}
    \caption{Velocity profiles of each controller on Path~C at $\sigma_{xy} = 0.05$~m (representative seed), where the black dashed lines show the velocity limits.}
    \label{fig:sim_noise}
\end{figure}

\begin{table}[!tbp]
\centering
\color{red}
\captionsetup{font+=revred}
\captionsetup[sub]{font+=revred}
\caption{Effect of localization noise on Path~C. Nonzero-noise entries are mean $\pm$ SD over 20 seeds, and the noise level $\sigma_{xy}$ is in meters. Dashes mark settings that never reached the goal within the 120~s timeout, whose error metrics cover the truncated run.}
\label{tab:sim_noise}
\scriptsize
\setlength{\tabcolsep}{1pt}
\begin{subtable}[t]{0.49\linewidth}
\centering
\caption{PP}
\begin{tabular}{rrrrrr}
\toprule
$\sigma_{xy}$ & Viol.\ [\%] & Mean [m] & Max [m] & Time [s] & Reach \\
\midrule
0 & 49.1 & 0.17 & 0.62 & 21.5 & Yes \\
0.01 & 68.3$\pm$1.7 & 0.17$\pm$0.01 & 0.62$\pm$0.01 & 21.8$\pm$0.2 & Yes \\
0.02 & 81.7$\pm$1.9 & 0.18$\pm$0.01 & 0.63$\pm$0.03 & 22.0$\pm$0.5 & Yes \\
0.05 & 92.8$\pm$1.2 & 0.19$\pm$0.01 & 0.64$\pm$0.04 & 21.7$\pm$0.6 & Yes \\
0.1 & 96.3$\pm$0.7 & 0.18$\pm$0.03 & 0.63$\pm$0.06 & 20.8$\pm$1.2 & Yes \\
\bottomrule
\end{tabular}
\end{subtable}
\hfill
\begin{subtable}[t]{0.49\linewidth}
\centering
\caption{APP}
\begin{tabular}{rrrrrr}
\toprule
$\sigma_{xy}$ & Viol.\ [\%] & Mean [m] & Max [m] & Time [s] & Reach \\
\midrule
0 & 31.3 & 0.13 & 0.55 & 20.0 & Yes \\
0.01 & 51.8$\pm$1.5 & 0.13$\pm$0.00 & 0.55$\pm$0.01 & 20.0$\pm$0.1 & Yes \\
0.02 & 70.6$\pm$1.7 & 0.13$\pm$0.00 & 0.55$\pm$0.02 & 20.0$\pm$0.1 & Yes \\
0.05 & 87.5$\pm$1.2 & 0.13$\pm$0.00 & 0.57$\pm$0.02 & 19.9$\pm$0.2 & Yes \\
0.1 & 93.4$\pm$0.8 & 0.14$\pm$0.01 & 0.58$\pm$0.04 & 19.8$\pm$0.4 & Yes \\
\bottomrule
\end{tabular}
\end{subtable}

\vspace{0.8em}

\begin{subtable}[t]{0.49\linewidth}
\centering
\caption{RPP}
\begin{tabular}{rrrrrr}
\toprule
$\sigma_{xy}$ & Viol.\ [\%] & Mean [m] & Max [m] & Time [s] & Reach \\
\midrule
0 & 27.5 & 0.04 & 0.18 & 21.7 & Yes \\
0.01 & 47.9$\pm$1.3 & 0.04$\pm$0.00 & 0.18$\pm$0.01 & 21.7$\pm$0.1 & Yes \\
0.02 & 67.8$\pm$1.7 & 0.04$\pm$0.00 & 0.19$\pm$0.01 & 21.9$\pm$0.1 & Yes \\
0.05 & 85.4$\pm$1.1 & 0.05$\pm$0.00 & 0.20$\pm$0.02 & 22.6$\pm$0.5 & Yes \\
0.1 & 91.4$\pm$0.7 & 0.05$\pm$0.01 & 0.19$\pm$0.04 & 22.4$\pm$0.8 & Yes \\
\bottomrule
\end{tabular}
\end{subtable}
\hfill
\begin{subtable}[t]{0.49\linewidth}
\centering
\caption{DWPP}
\begin{tabular}{rrrrrr}
\toprule
$\sigma_{xy}$ & Viol.\ [\%] & Mean [m] & Max [m] & Time [s] & Reach \\
\midrule
0 & 0.0 & 0.02 & 0.11 & 23.7 & Yes \\
0.01 & 0.0$\pm$0.0 & 0.02$\pm$0.00 & 0.11$\pm$0.00 & 26.4$\pm$1.3 & Yes \\
0.02 & 0.0$\pm$0.0 & 0.01$\pm$0.00 & 0.10$\pm$0.01 & 48.8$\pm$5.5 & Yes \\
0.05 & 0.0$\pm$0.0 & 0.01$\pm$0.00 & 0.09$\pm$0.01 & -- & No \\
0.1 & 0.0$\pm$0.0 & 0.01$\pm$0.00 & 0.20$\pm$0.04 & -- & No \\
\bottomrule
\end{tabular}
\end{subtable}
\end{table}

To analyze the effect of localization noise, zero-mean Gaussian noise was added to the pose fed to the controller at every control step. The position-noise standard deviation was swept over $\sigma_{xy} \in \{0, 0.01, 0.02, 0.05, 0.10\}$~m, and the heading-noise standard deviation followed the fixed ratio $\sigma_{\theta}/\sigma_{xy} = 0.65$~rad/m, taken from the AMCL corrections measured in Section~\ref{sec:experiment_setting}. For each noise level, 20 trials were performed on Path~C, and each trial generated its noise sequence from a different random seed. The same 20 seeds were shared by every controller and noise level, so all controllers experienced identical noise sequences. The tracking error is always evaluated with the true pose. Note that the noise is redrawn independently at every 30~Hz control step, whereas the corrections of the actual localization module arrived at about 7~Hz and are temporally correlated. Table~\ref{tab:sim_noise} reports the four metrics and whether the goal was reached within the timeout, and Fig.~\ref{fig:sim_noise} shows the velocity profiles of every controller for a representative seed at $\sigma_{xy} = 0.05$~m.

The violation ratios of PP, APP, and RPP increased with the noise level, whereas DWPP maintained exactly zero violations at every noise level (Table~\ref{tab:sim_noise}).

In the tracking error and the travel time, PP, APP, and RPP were nearly insensitive to the noise, whereas DWPP stopped reaching the goal in some settings. The mean and maximum errors of PP, APP, and RPP varied only marginally across the noise levels, and their travel times stayed close to the noiseless values. DWPP achieved the smallest tracking error among the four controllers in the realistic range ($\sigma_{xy} \le 0.02$~m, covering the correction magnitudes measured on the real robot), responding to the noise by slowing down at the cost of a longer travel time. However, at $\sigma_{xy} \ge 0.05$~m, this slowdown grew until DWPP no longer reached the goal within the 120~s timeout.

The velocity profiles in Fig.~\ref{fig:sim_noise} make the reason for this difference visible. The linear velocity commands of PP and APP stay at the maximum velocity, and that of RPP is kept at or above $v^{\mathrm{min}}_{\mathrm{reg}}$ even when the regulation heuristic reduces it. The linear velocity command of DWPP, by contrast, cannot accelerate, and the run consequently does not reach the goal. Section~\ref{sec:discussion_noise} discusses the mechanism behind this behavior, and Section~\ref{sec:sim_lookahead} quantifies a practical mitigation, a larger lookahead distance.

\end{revblock}

\FloatBarrier
\subsection{Effect of Lookahead Distance}
\label{sec:simulation}
\label{sec:sim_lookahead}

\begin{table}[!tbp]
\centering
\color{red}
\captionsetup{font+=revred}
\captionsetup[sub]{font+=revred}
\caption{Effect of the lookahead distance $L$ of DWPP on Path~C. The $\sigma_{xy}=0.05$~m entries are mean $\pm$ standard deviation over 20 seeds. Dashes mark settings that never reached the goal within the 120~s timeout.}
\label{tab:sim_lookahead}
\footnotesize
\setlength{\tabcolsep}{3pt}
\begin{subtable}[t]{0.44\linewidth}
\centering
\caption{$\sigma_{xy}=0$~m}
\begin{tabular}{rrrr}
\toprule
$L$ [m] & Mean [m] & Max [m] & Time [s] \\
\midrule
0.2 & 0.04 & 0.21 & 30.0 \\
0.3 & 0.04 & 0.20 & 25.2 \\
0.4 & 0.04 & 0.21 & 23.1 \\
0.5 & 0.05 & 0.23 & 22.1 \\
0.6 & 0.06 & 0.25 & 21.9 \\
0.7 & 0.08 & 0.28 & 21.7 \\
0.8 & 0.10 & 0.34 & 21.4 \\
0.9 & 0.13 & 0.42 & 21.1 \\
1.0 & 0.17 & 0.52 & 20.4 \\
1.1 & 0.21 & 0.62 & 19.7 \\
1.2 & 0.24 & 0.71 & 19.0 \\
\bottomrule
\end{tabular}
\end{subtable}
\hfill
\begin{subtable}[t]{0.52\linewidth}
\centering
\caption{$\sigma_{xy}=0.05$~m}
\begin{tabular}{rrrrr}
\toprule
$L$ [m] & Mean [m] & Max [m] & Time [s] & Reach \\
\midrule
0.2 & 0.00$\pm$0.00 & 0.08$\pm$0.01 & -- & No \\
0.3 & 0.01$\pm$0.00 & 0.09$\pm$0.01 & -- & No \\
0.4 & 0.01$\pm$0.00 & 0.12$\pm$0.01 & 92.4$\pm$4.4 & Yes \\
0.5 & 0.02$\pm$0.00 & 0.15$\pm$0.01 & 61.0$\pm$3.3 & Yes \\
0.6 & 0.03$\pm$0.00 & 0.19$\pm$0.01 & 45.1$\pm$2.8 & Yes \\
0.7 & 0.04$\pm$0.00 & 0.26$\pm$0.01 & 37.0$\pm$1.9 & Yes \\
0.8 & 0.06$\pm$0.00 & 0.32$\pm$0.02 & 33.1$\pm$1.5 & Yes \\
0.9 & 0.09$\pm$0.00 & 0.41$\pm$0.01 & 31.2$\pm$1.3 & Yes \\
1.0 & 0.13$\pm$0.01 & 0.52$\pm$0.02 & 29.9$\pm$2.1 & Yes \\
1.1 & 0.16$\pm$0.01 & 0.62$\pm$0.01 & 29.9$\pm$1.8 & Yes \\
1.2 & 0.19$\pm$0.01 & 0.69$\pm$0.03 & 26.3$\pm$1.9 & Yes \\
\bottomrule
\end{tabular}
\end{subtable}
\end{table}

\begin{figure}[!tbp]
    \centering
    \color{red}
    \captionsetup{font+=revred}
    \captionsetup[sub]{font+=revred}
    \begin{subfigure}[b]{0.57\linewidth}
        \centering
        \includegraphics[width=\linewidth]{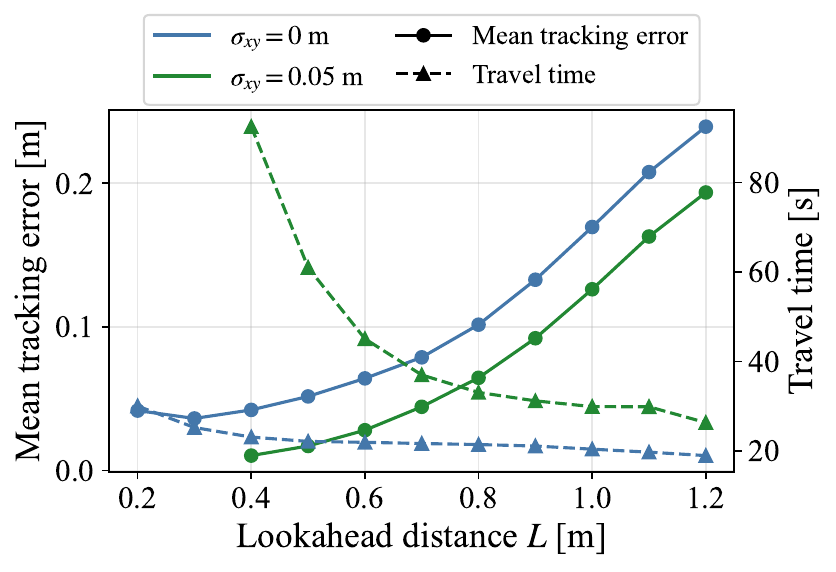}
        \caption{Mean tracking error and travel time vs.\ $L$}
        \label{fig:sim_lookahead_tradeoff}
    \end{subfigure}
    \hfill
    \begin{subfigure}[b]{0.36\linewidth}
        \centering
        \includegraphics[width=\linewidth]{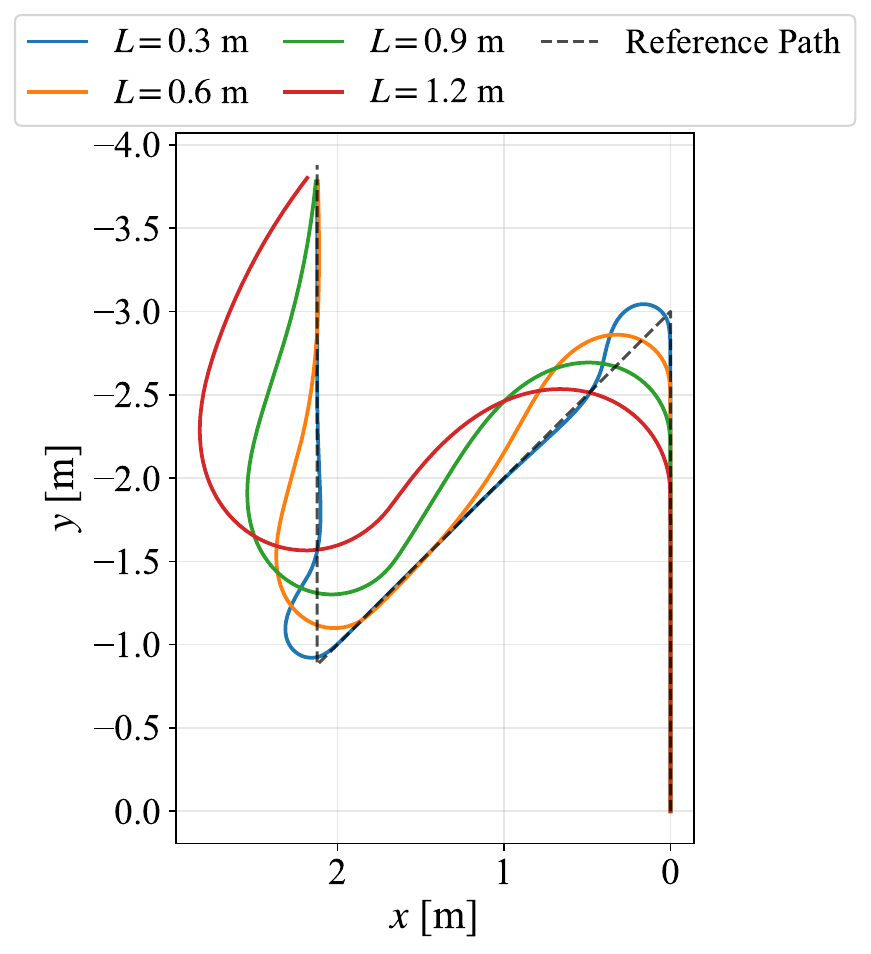}
        \caption{Tracked paths for representative $L$ (noiseless)}
        \label{fig:sim_lookahead_traj}
    \end{subfigure}
    \caption{Effect of the lookahead distance $L$ of DWPP on Path~C under localization noise. The settings with $L \le 0.3$~m at $\sigma_{xy}=0.05$~m did not reach the goal within the timeout and are excluded from the curves.}
    \label{fig:sim_lookahead}
\end{figure}

\begin{revblock}
Finally, the effect of the lookahead distance on the path-tracking error and the travel time of DWPP was investigated, including its interaction with localization noise. DWPP was run with a fixed lookahead distance $L$ swept over $0.2$--$1.2$~m in steps of $0.1$~m on Path~C, with the localization noise of Section~\ref{sec:sim_noise} switched off and on at the severe level $\sigma_{xy} = 0.05$~m (20 seeds for the noisy setting). DWPP satisfied the acceleration constraints in every setting. Table~\ref{tab:sim_lookahead} reports the remaining metrics, and Fig.~\ref{fig:sim_lookahead} shows the mean tracking error and the travel time as functions of $L$, together with the tracked paths for representative $L$.

Without noise, the classical trade-off is recovered. A small $L$ yields a small tracking error at the cost of a longer travel time, because the tighter arc curvature triggers stronger velocity reduction, whereas a large $L$ shortens the travel time but increases corner cutting and hence the tracking error.

Under localization noise, a lower bound on the usable lookahead distance appears. Every setting reached the goal without noise, whereas at $\sigma_{xy} = 0.05$~m, $L \le 0.3$~m never reached the goal within the timeout, $L = 0.4$~m reached it in every seed, and larger $L$ progressively traded accuracy for speed. Section~\ref{sec:discussion_noise} discusses why a larger $L$ mitigates the noise effect.

\end{revblock}

\section{Discussion}
\label{sec:discussion}

\subsection{Why DWPP Outperforms RPP}

\begin{figure*}[tbp]
    \centering

    % ---- 凡例（横長） ----
    \begin{subfigure}[b]{0.25\textwidth}
        \centering
        \includegraphics[width=\textwidth]{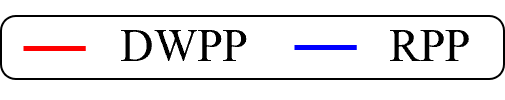}
    \end{subfigure}
    
    % \vspace{0.5em}
    
    % --- (a) 速度プロファイル (元 fig6.png) ---
    \begin{subfigure}{\textwidth}
        \centering
        % 縦長になりすぎないよう幅を調整 (例: 0.8\linewidth)
        \includegraphics[width=0.8\linewidth]{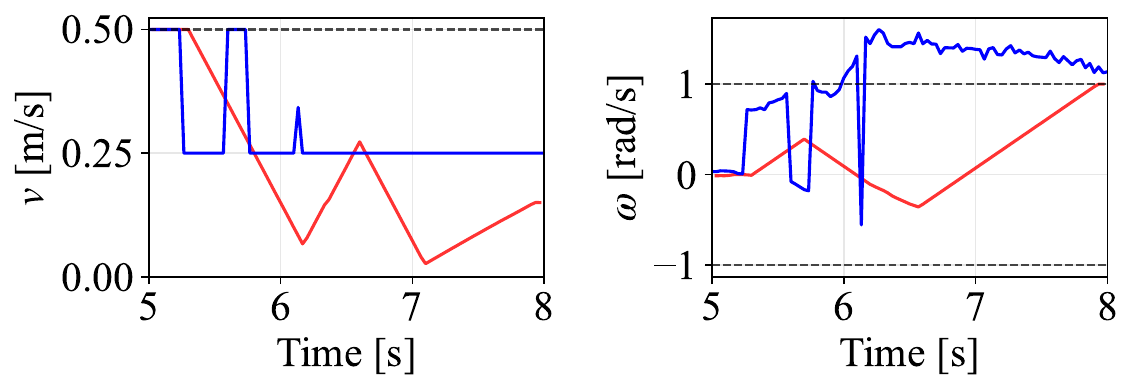}
        \caption{Command velocity profile of RPP and DWPP during tracking Path C}
        \label{fig:vel_profile}
    \end{subfigure}
    
    \vspace{1.5em}

    % ---- 凡例（横長） ----
    \begin{subfigure}[b]{0.7\textwidth}
        \centering
        \includegraphics[width=\textwidth]{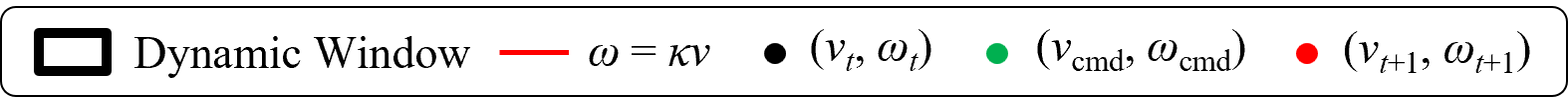}
    \end{subfigure}
    
    \vspace{0.5em}

    % --- (b) RPP の時系列 DW ---
    \begin{subfigure}{\textwidth}
        \centering
        \begin{minipage}{0.19\textwidth}
            \centering \scriptsize $t=7.033$ \\
            \includegraphics[width=\textwidth]{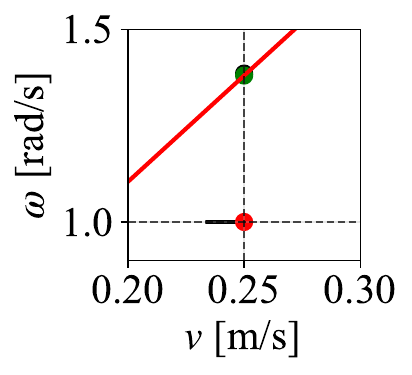}
        \end{minipage}
        \hfill
        \begin{minipage}{0.19\textwidth}
            \centering \scriptsize $t=7.067$ \\
            \includegraphics[width=\textwidth]{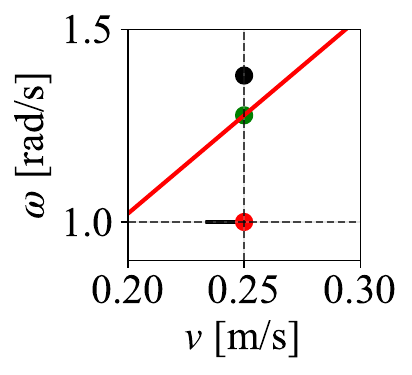}
        \end{minipage}
        \hfill
        \begin{minipage}{0.19\textwidth}
            \centering \scriptsize $t=7.100$ \\
            \includegraphics[width=\textwidth]{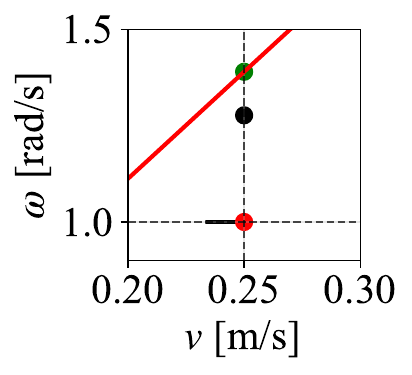}
        \end{minipage}
        \hfill
        \begin{minipage}{0.19\textwidth}
            \centering \scriptsize $t=7.133$ \\
            \includegraphics[width=\textwidth]{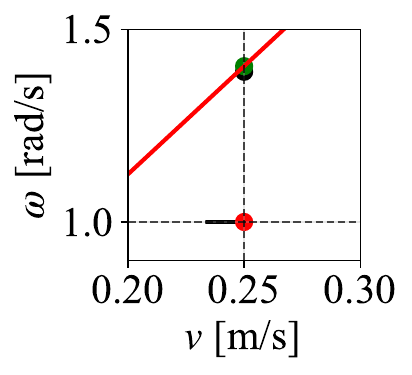}
        \end{minipage}
        \hfill
        \begin{minipage}{0.19\textwidth}
            \centering \scriptsize $t=7.166$ \\
            \includegraphics[width=\textwidth]{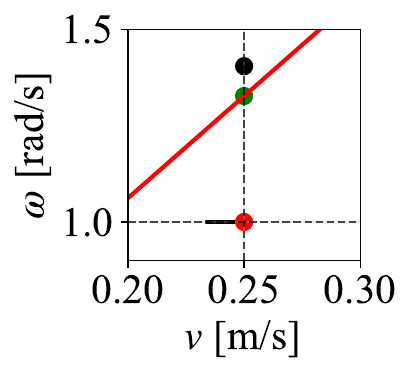}
        \end{minipage}
        \caption{Transition of the dynamic window in RPP.}
        \label{fig:dw_transition_rpp}
    \end{subfigure}

    \vspace{1.5em}

    % --- (c) DWPP の時系列 DW ---
    \begin{subfigure}{\textwidth}
        \centering
        \begin{minipage}{0.19\textwidth}
            \centering \scriptsize $t=7.033$ \\
            \includegraphics[width=\textwidth]{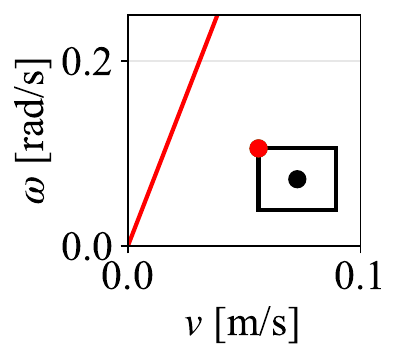}
        \end{minipage}
        \hfill
        \begin{minipage}{0.19\textwidth}
            \centering \scriptsize $t=7.067$ \\
            \includegraphics[width=\textwidth]{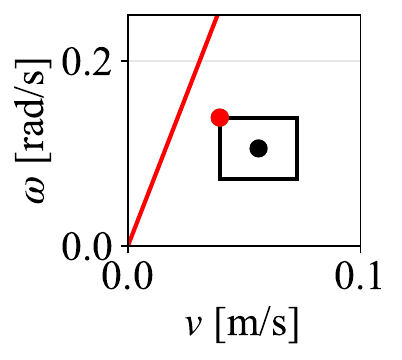}
        \end{minipage}
        \hfill
        \begin{minipage}{0.19\textwidth}
            \centering \scriptsize $t=7.100$ \\
            \includegraphics[width=\textwidth]{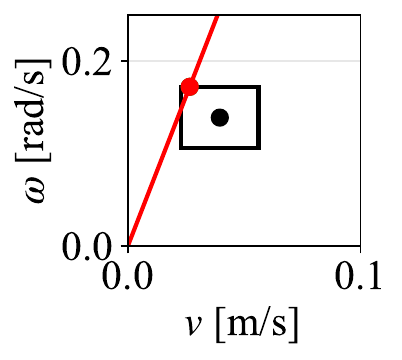}
        \end{minipage}
        \hfill
        \begin{minipage}{0.19\textwidth}
            \centering \scriptsize $t=7.133$ \\
            \includegraphics[width=\textwidth]{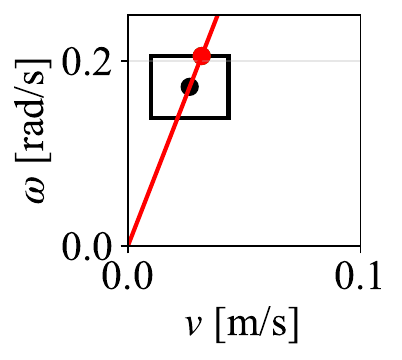}
        \end{minipage}
        \hfill
        \begin{minipage}{0.19\textwidth}
            \centering \scriptsize $t=7.166$ \\
            \includegraphics[width=\textwidth]{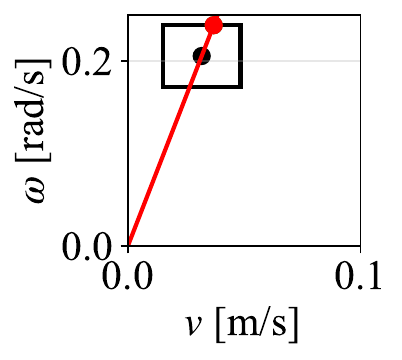}
        \end{minipage}
        \caption{Transition of the dynamic window in the proposed DWPP.}
        \label{fig:dw_transition_dwpp}
    \end{subfigure}

    % --- 図全体の共通キャプション ---
    \caption{Analysis of command velocity computation. (a) Command velocity profile; (b) and (c) dynamic window transitions at the specific time frames indicated above each plot.}
    \label{fig:integrated_analysis}
\end{figure*}

Figure \ref{fig:vel_profile} shows the velocity command profiles of DWPP and RPP during the first corner of Path C ($t = 5.0$~s to $t = 8.0$~s), where a performance gap between the two methods becomes evident. To further analyze the underlying cause, Figs. \ref{fig:dw_transition_rpp} and \ref{fig:dw_transition_dwpp} illustrate the transitions of velocity command calculation in the $v$-$\omega$ space for the interval from $t = 7.033$~s to $t = 7.166$~s. During this period, DWPP underwent deceleration, whereas RPP maintained a constant linear velocity.

These results provide the following insights. In RPP, the linear velocity is fixed at $v^{\min}_{\mathrm{reg}}$, and the angular velocity is determined based on Eq. \eqref{eq:pp_angular_velocity} without considering velocity and acceleration constraints. Consequently, the velocity commands are clipped by these constraints during execution. Consequently, although the left vertex of the dynamic window is the optimal feasible point for path tracking, the actual realized velocity corresponds to the right vertex.

In contrast, DWPP selects the point within the dynamic window that is closest to the line defined by Eq. \eqref{eq:pp_angular_velocity} (i.e., $\omega = \kappa v$). This approach ensures that the calculated velocity commands are directly executable without being saturated. Moreover, DWPP effectively manages deceleration from $t=7.033$ to $7.100$ to maintain tracking and accelerates from $t=7.133$ to $7.166$. This demonstrates that DWPP can calculate velocity commands that \rev{best satisfy the path-tracking condition} while strictly adhering to velocity and acceleration constraints. These observations confirm that by explicitly incorporating these constraints into the calculation phase, DWPP achieves a more ideal and superior path-tracking performance compared with RPP.

\subsection{Key Advantages of DWPP over Conventional Pure Pursuit Variants}

The primary advantages of DWPP are summarized into the following three points:

\begin{itemize}
    \item \textbf{Direct compliance with velocity and acceleration constraints:} DWPP generates velocity commands that inherently satisfy velocity and acceleration constraints. This eliminates the need for external clipping mechanisms or post-processing layers, such as the \texttt{velocity\-\_smoother} in Nav2, which were previously required to ensure feasibility before command execution.
    
    \item \textbf{\rev{Accurate} tracking within velocity and acceleration constraints:} By incorporating velocity and acceleration constraints directly into the command velocity calculation phase, DWPP can determine the velocity commands \rev{that best satisfy the path-tracking condition $\omega = \kappa v$ within the constraints}, for instance, by automatically decelerating at sharp corners. Consequently, DWPP can track challenging paths with discontinuous or high curvature (e.g., Path C) with \rev{substantially} reduced overshoot compared with conventional \rev{pure pursuit variants}.

    \item \textbf{Parameter-free deceleration:} DWPP \rev{derives its deceleration directly from the constraints} without the need for manual parameter tuning. While RPP can also decelerate at high-curvature sections using its curvature heuristic, it requires the manual design of parameters such as $R_{\min}$ and $v^\mathrm{min}_{\mathrm{reg}}$. In contrast, DWPP automatically calculates \rev{the deceleration} based solely on the provided velocity and acceleration limits. Furthermore, DWPP remains compatible with manually designed heuristics, enabling it to inherit their benefits if additional specific tuning is desired.
\end{itemize}

\subsection{Key Advantages of DWPP over Dynamic Window Approach}
DWA~\cite{fox2002dynamic} is another prominent local planner that utilizes the dynamic window. In this section, we discuss the comparative advantages of DWPP over DWA, specifically when DWA is employed solely for the purpose of path tracking. These advantages are primarily categorized into the following two aspects:

\begin{itemize}
    \item \textbf{Analytical optimality in a continuous solution space:} In DWA, candidate velocity commands are obtained through discrete sampling within the dynamic window. This results in a set of discrete solutions, which may result in velocity commands that slightly deviate from \rev{the exact optimum within the dynamic window}. In contrast, DWPP analytically computes the command velocity within a continuous solution space by directly exploiting the geometric relationship between the line $\omega = \kappa v$ and the dynamic window in the $v$-$\omega$ space. This ensures that the optimal velocity command is always obtained within the feasible constraints.

    \item \textbf{Computational efficiency:} The computational complexity of DWA is typically $O(NM)$, where $N$ denotes the number of velocity samples and $M$ represents the number of simulated time steps in the prediction horizon. On the other hand, as described in Section \ref{sec:3}, the computational cost of DWPP is $O(1)$ because it derives the velocity commands through a closed-form analytical solution.
\end{itemize}

Consequently, when the primary objective of the local planner is pure path tracking without obstacle avoidance, DWPP can calculate velocity commands \rev{closer to the path-tracking condition} with \rev{much} lower computational overhead compared with DWA.

\begin{revblock}
\subsection{Mechanism of the Localization-Noise Effect}
\label{sec:discussion_noise}

Sections~\ref{sec:sim_noise} and \ref{sec:sim_lookahead} showed a sharp contrast under localization noise. The tracking error and the travel time of PP, APP, and RPP barely changed, whereas DWPP under severe noise stopped accelerating and no longer reached the goal when the lookahead distance was small. This subsection explains both behaviors from the command-generation mechanism.

PP, APP, and RPP keep progressing because their linear velocity stays high under the noise and their angular velocity does not change sharply owing to the clipping by the velocity and acceleration limits. Regarding the linear velocity, PP and APP always command the maximum velocity, and RPP keeps at least $v^{\mathrm{min}}_{\mathrm{reg}}$, so a situation in which these controllers decelerate to the point of no longer progressing rarely arises. Regarding the angular velocity, the command follows $\omega = \kappa v$, and the curvature $\kappa$ fluctuates with the position and heading noise, so $\omega$ also fluctuates, which causes the high violation ratios in Table~\ref{tab:sim_noise}. However, the executed $\omega$ is clipped by the velocity and acceleration limits, so its sharp changes are suppressed, the resulting motion stays smooth, and the tracking error is therefore hardly affected by the noise.

In DWPP, by contrast, the noise perturbs $\kappa$ and can make it larger than the true value, and a short lookahead distance enlarges $\kappa$ further, so the computed command can reduce the linear velocity and increase the angular velocity. Moreover, on the straight segments the sign of $\kappa$ can flip at every control period, in which case the angular command is held near zero. In addition, when the velocity-scaled lookahead mechanism of APP is adopted, a low linear velocity shortens the lookahead distance, which makes this cycle even more likely. Consequently, DWPP can become unable to accelerate when the noise is large and the lookahead distance is short. \ref{app:noise_bound} quantifies this mechanism and derives a closed-form estimate of the smallest usable lookahead distance.

In practice, this issue is addressed by improving the localization accuracy or by increasing the lookahead distance. Indeed, Section~\ref{sec:sim_lookahead} shows that DWPP reaches the goal without difficulty when the localization is accurate or the lookahead distance is long. Extending the method to schedule the lookahead distance by explicitly accounting for the level of the localization noise is left for future work.
\end{revblock}

\subsection{\texorpdfstring{\rev{Extension to Actuator-Level Constraints}}{Extension to Actuator-Level Constraints}}
\label{sec:discussion_nonholonomic}

\begin{revblock}

The velocity and acceleration limits considered in this study are operational design constraints on the body-velocity commands, chosen for example for ride comfort and safety, and Section~\ref{subsec:problem_formulation} assumes that they lie within the physical capabilities of the platform. This subsection discusses how the geometric framework of DWPP extends to the velocity and acceleration limits of the individual actuators.

First, consider a differential-drive robot with wheel radius $r_{\mathrm{w}}$, wheel separation $b$, and wheel angular velocities $\dot{\phi}_{\mathrm{R}}$ and $\dot{\phi}_{\mathrm{L}}$. The inverse kinematics
\begin{equation}
    \dot{\phi}_{\mathrm{R}} = \frac{1}{r_{\mathrm{w}}}\left(v + \frac{b}{2}\omega\right),
    \qquad
    \dot{\phi}_{\mathrm{L}} = \frac{1}{r_{\mathrm{w}}}\left(v - \frac{b}{2}\omega\right)
    \label{eq:differential_drive_inverse_kinematics}
\end{equation}
maps symmetric wheel-velocity limits $|\dot{\phi}_{\mathrm{R}}|, |\dot{\phi}_{\mathrm{L}}| \leq \dot{\phi}_{\max}$ into the $v$-$\omega$ plane as
\begin{equation}
    \left|v + \frac{b}{2}\omega\right| \leq r_{\mathrm{w}}\dot{\phi}_{\max},
    \qquad
    \left|v - \frac{b}{2}\omega\right| \leq r_{\mathrm{w}}\dot{\phi}_{\max},
    \label{eq:differential_drive_velocity_region}
\end{equation}
which is the feasible region representing the velocity limits and takes a diamond shape (a parallelogram under asymmetric or wheel-dependent limits). The wheel-acceleration limits $|\ddot{\phi}_{\mathrm{R}}|, |\ddot{\phi}_{\mathrm{L}}| \leq \ddot{\phi}_{\max}$ map in the same manner. Specifically, differentiating the inverse kinematics of Eq.~\eqref{eq:differential_drive_inverse_kinematics}, imposing the wheel-acceleration limits, and multiplying by $\Delta t$ yields another diamond centered at the current body velocity $(v_t, \omega_t)$. The intersection of the two diamonds is the actuator-level dynamic window, as shown in Fig.~\ref{fig:actuator_window_diffdrive}. The same closest-point selection of Section~\ref{sec:3} applies, with the candidate set extended to the vertices of this polygon and the intersections of its edges with the line $\omega = \kappa v$.

\begin{figure}[tbp]
    \centering
    \captionsetup{font+=revred}
    \captionsetup[sub]{font+=revred}
    \begin{subfigure}[b]{0.48\linewidth}
        \centering
        \includegraphics[width=\linewidth]{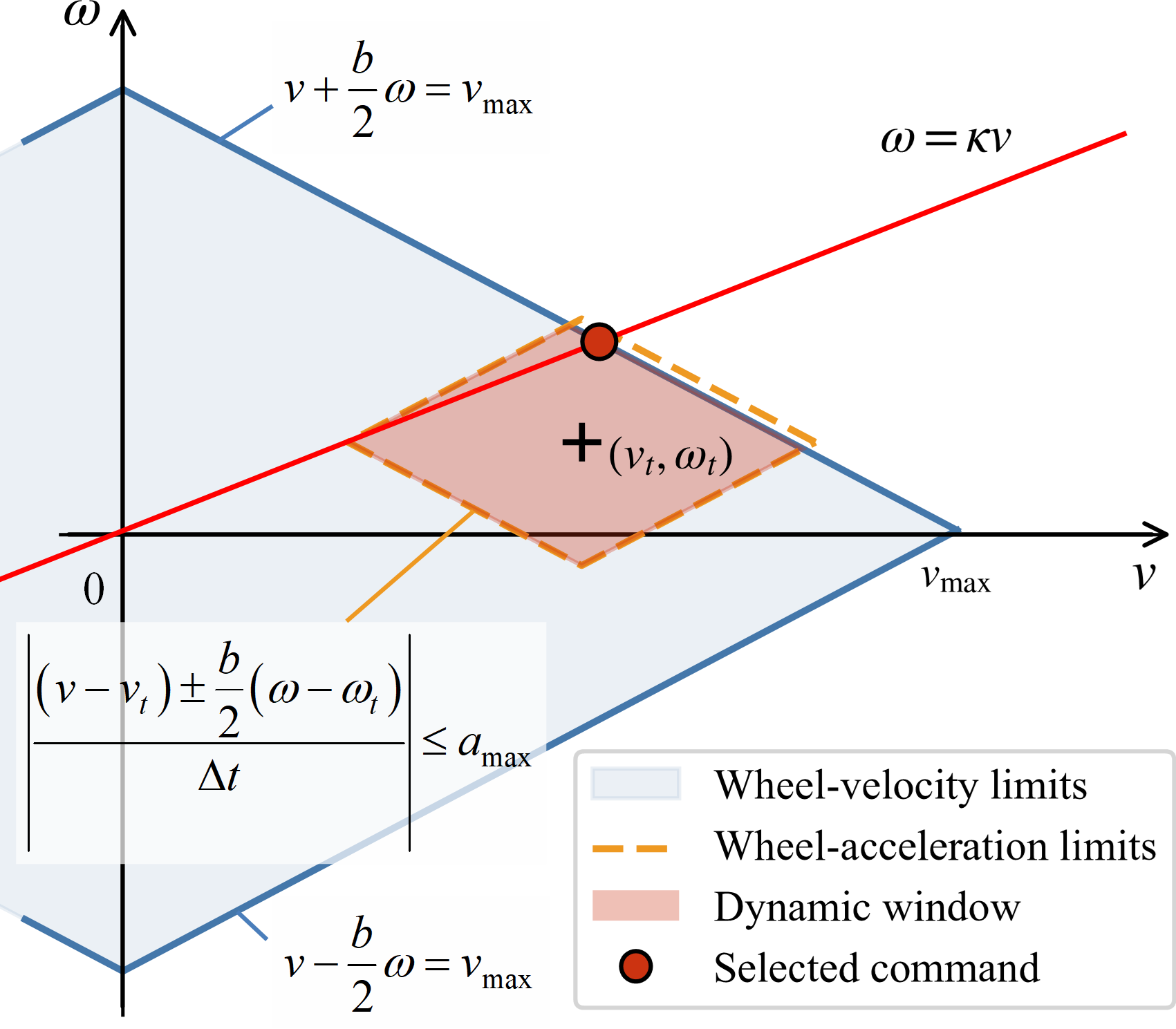}
        \caption{Differential drive}
        \label{fig:actuator_window_diffdrive}
    \end{subfigure}
    \hfill
    \begin{subfigure}[b]{0.48\linewidth}
        \centering
        \includegraphics[width=\linewidth]{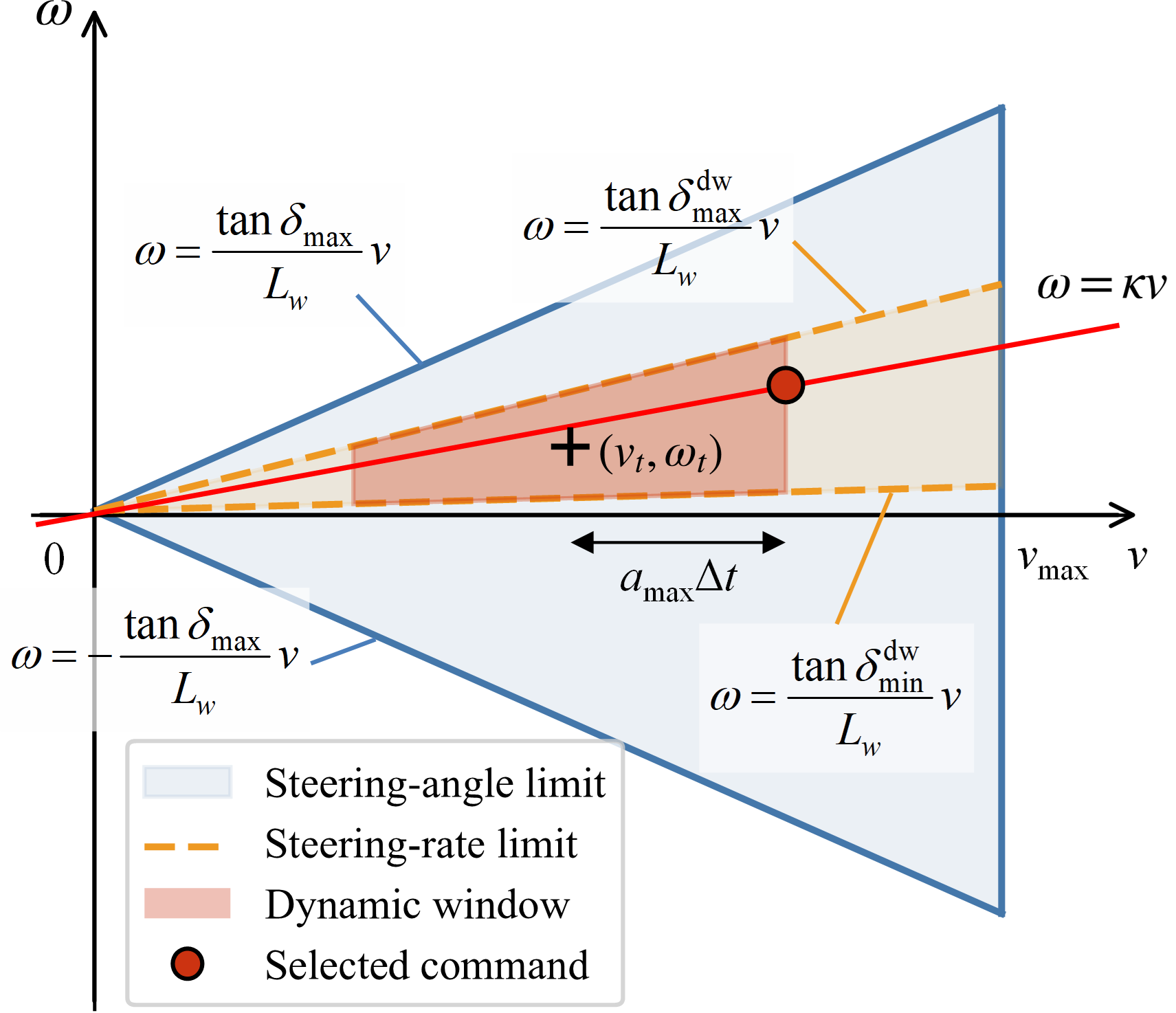}
        \caption{Ackermann steering}
        \label{fig:actuator_window_ackermann}
    \end{subfigure}
    \caption{Actuator-level dynamic windows in the $v$--$\omega$ plane.}
    \label{fig:actuator_windows}
\end{figure}

Next, consider an Ackermann-steered vehicle. The kinematic bicycle model with the rear-axle center as the reference point gives $\omega = v\tan\delta/L_{\mathrm{w}}$, where $\delta$ is the steering angle and $L_{\mathrm{w}}$ is the wheelbase. A steering-angle limit $|\delta| \leq \delta_{\max}$ therefore maps to the triangular region $|\omega| \leq |v|\tan\delta_{\max}/L_{\mathrm{w}}$ with its apex at the origin, which reflects that the vehicle cannot rotate in place. Next, consider a steering-rate limit $|\dot{\delta}| \leq \dot{\delta}_{\max}$. For forward motion, the steering interval reachable within one control period, $[\delta^{\mathrm{dw}}_{\min}, \delta^{\mathrm{dw}}_{\max}] = [\max(-\delta_{\max}, \delta_t - \dot{\delta}_{\max}\Delta t),\ \min(\delta_{\max}, \delta_t + \dot{\delta}_{\max}\Delta t)]$, spans
\begin{equation}
    \frac{\tan\delta^{\mathrm{dw}}_{\min}}{L_{\mathrm{w}}}v
    \leq \omega \leq
    \frac{\tan\delta^{\mathrm{dw}}_{\max}}{L_{\mathrm{w}}}v,
    \label{eq:ackermann_reachable_region}
\end{equation}
which forms a triangular region depending on the linear velocity $v$. The intersection of this region with the admissible interval of the linear velocity is a triangular or trapezoidal window, as shown in Fig.~\ref{fig:actuator_window_ackermann}. The same closest-point selection again applies to this polygonal window.

In general, let $\mathcal{U}(t)$ denote the set of actuator states reachable within one control period and $\Psi_t$ the kinematic mapping from the actuator space to the body-velocity space. The dynamic window then becomes $\Psi_t(\mathcal{U}(t))$, and the command selection of Section~\ref{sec:3} again applies to it. When the mapped set is a convex polygon with a fixed number of edges, the command is still obtained from a fixed number of line--edge intersections and projections, preserving the closed-form $O(1)$ computation. For nonlinear or nonconvex sets, the same formulation remains applicable, although the closed form is not necessarily preserved. The detailed implementation and experimental evaluation of these extensions are left for future work.
\end{revblock}

\section{Conclusion}
\label{sec:conclusion}

This paper presents dynamic window pure pursuit (DWPP), a novel pure pursuit method designed to explicitly incorporate velocity and acceleration constraints into the velocity command computation process. By selecting the optimal point within the dynamic window closest to the line $\omega = \kappa v$, the proposed method analytically derives commands that \rev{best satisfy the path-tracking condition} while ensuring strict adherence to these constraints.  \rev{Real-robot experiments integrated with the Nav2 framework showed that the generated commands remained within the specified velocity and acceleration limits and that DWPP reduced the path-tracking errors compared with conventional pure pursuit variants on the paths that demand sharp turning. Simulation studies confirmed that these results persist across the parameter sweeps of the baselines and under matched travel time.} The DWPP implementation has been integrated into the Nav2 framework and is publicly available~\footnote{\url{https://github.com/ros-navigation/navigation2}}, enabling engineers to readily evaluate and adopt the proposed method.

Future research will focus \rev{on scheduling the lookahead distance by explicitly accounting for the level of the localization noise and on extending the dynamic-window formulation to actuator-level constraints such as those of Ackermann-steered vehicles}.

\section*{Acknowledgment}
The authors would like to express their sincere gratitude to Steve Macenski and Maurice Alexander Purnawan (Open Navigation LLC) for their thorough and insightful code reviews during the integration of this work into the Nav2 framework.

\section*{Declaration of competing interest}
The authors declare that they have no known competing financial interests or personal relationships that could have appeared to influence the work reported in this paper.

%% If you have bib database file and want bibtex to generate the
%% bibitems, please use
%%
 \appendix

\begin{revblock}
\section{Finite-Time Recovery of the Intersection}
\label{app:stability}
\setcounter{equation}{0}
\renewcommand{\theequation}{A.\arabic{equation}}

This appendix states and proves the finite-time recovery result summarized in Section~\ref{sec:stability}.
% First, the notation and assumptions used in the proof are introduced.
In the $v$--$\omega$ plane, let $\mathcal{D}_t$ denote the regulated dynamic window at control step $t$, i.e., the rectangle of Eqs.~\eqref{eq:dw_v_max}--\eqref{eq:dw_w_min} intersected with $v \in [0, v_{\mathrm{reg}}]$, and let $\mathcal{G} = [\max(v_{\min},0),\, \min(v_{\max}, v_{\mathrm{reg}})] \times [\omega_{\min}, \omega_{\max}]$ be the fixed admissible set, so that $\mathcal{D}_t \subseteq \mathcal{G}$. Let $L_\kappa = \{(v,\omega) : \omega = \kappa v\}$ denote the line of Eq.~\eqref{eq:pp_angular_velocity} and $\eta(v,\omega) = \omega - \kappa v$ the signed deviation. Since the distance from a point to $L_\kappa$ is $|\eta|/\sqrt{1+\kappa^2}$, DWPP minimizes $|\eta|$ over $\mathcal{D}_t$, and $L_\kappa \cap \mathcal{D}_t = \emptyset$ exactly when $\eta$ has a constant nonzero sign on $\mathcal{D}_t$. Three assumptions are used:
\begin{itemize}
    \item[(A1)] the curvature command $\kappa$ of Eq.~\eqref{eq:pp_curvature} and the regulated velocity $v_{\mathrm{reg}}$ are held constant over the steps considered (the reference path consists of straight and constant-curvature segments as in \cite{ollero1995stability}, where $\kappa$ is constant at the tracking equilibrium);
    \item[(A2)] commands are realized within one control period; that is, $(v_t, \omega_t)$ is the previous command and lies in $\mathcal{D}_{t-1} \subseteq \mathcal{G}$, and hence also in $\mathcal{D}_t$, so the dynamic window is never empty;
    \item[(A3)] $L_\kappa \cap \mathcal{G} \neq \emptyset$, i.e., the curvature is realizable within the limits, which holds for every $\kappa$ whenever $v_{\min} \le 0$ and $\omega_{\min} \le 0 \le \omega_{\max}$, since the origin then belongs to $\mathcal{G}$.
\end{itemize}

\begin{proposition}[Finite-time recovery of the intersection]
\label{prop:recovery}
Let (A1)--(A3) hold and suppose $L_\kappa \cap \mathcal{D}_t = \emptyset$. Let $\eta^*_t = \min_{\mathbf{p} \in \mathcal{D}_t} |\eta(\mathbf{p})| > 0$ and define the recovery rate
\begin{equation}
\delta = \Delta t \cdot
\begin{cases}
\min\!\left(\alpha^{\mathrm{acc}}_{\max},\,
\alpha^{\mathrm{dcc}}_{\max},\,
|\kappa|\, a^{\mathrm{acc}}_{\max},\,
|\kappa|\, a^{\mathrm{dcc}}_{\max}\right) & (\kappa \neq 0), \\[2pt]
\min\!\left(\alpha^{\mathrm{acc}}_{\max},\,
\alpha^{\mathrm{dcc}}_{\max}\right) & (\kappa = 0).
\end{cases}
\label{eq:recovery_rate}
\end{equation}
Then the selected command $\mathbf{q}_t = (v_{t+1}, \omega_{t+1})$ attains $|\eta(\mathbf{q}_t)| = \eta^*_t$, and either $L_\kappa \cap \mathcal{D}_{t+1} \neq \emptyset$ or $\eta^*_{t+1} \le \eta^*_t - \delta$. Consequently, the non-intersection condition arising at step $t_0$ lasts at most $\lceil \eta^*_{t_0} / \delta \rceil$ steps.
\end{proposition}
\begin{pf}
Since DWPP selects the point of $\mathcal{D}_t$ closest to $L_\kappa$, the command $\mathbf{q}_t$ minimizes $|\eta|$ and attains $|\eta(\mathbf{q}_t)| = \eta^*_t$. Assume $\eta > 0$ on $\mathcal{D}_t$; the case $\eta < 0$ is proved in the same way after replacing the lower bounds below with the upper bounds and interchanging the acceleration and deceleration limits.

First, it is shown that at each step either the intersection is recovered or the deviation decreases. Under (A2), $(v_{t+1}, \omega_{t+1}) = \mathbf{q}_t$, so $\mathcal{D}_{t+1}$ is the regulated dynamic window built around $\mathbf{q}_t$. Because $\mathcal{D}_{t+1}$ contains $\mathbf{q}_t$ with $\eta(\mathbf{q}_t) = \eta^*_t > 0$, the line $L_\kappa$ intersects $\mathcal{D}_{t+1}$ if and only if $\min_{\mathcal{D}_{t+1}} \eta \le 0$: when the minimum is nonpositive, the intermediate value theorem yields a zero of $\eta$ on the convex set $\mathcal{D}_{t+1}$, and otherwise $\eta > 0$ on $\mathcal{D}_{t+1}$. Since $\eta = \omega - \kappa v$ is increasing in $\omega$ and, for $\kappa > 0$ (resp.\ $\kappa < 0$), decreasing (resp.\ increasing) in $v$, this minimum is attained at the vertex $\mathbf{p}' = (v', \omega')$ with $\omega' = \max(\omega_{\min},\, \omega_{t+1} - \alpha^{\mathrm{dcc}}_{\max} \Delta t)$ and $v' = \min(v_{\max},\, v_{\mathrm{reg}},\, v_{t+1} + a^{\mathrm{acc}}_{\max} \Delta t)$ for $\kappa \ge 0$ (resp.\ $v' = \max(v_{\min},\, 0,\, v_{t+1} - a^{\mathrm{dcc}}_{\max} \Delta t)$ for $\kappa < 0$); for $\kappa = 0$, $\eta$ does not depend on $v$ and the same $v'$ is used. With the displacements $\Delta\omega = \omega_{t+1} - \omega' \ge 0$ and $\Delta v = |v' - v_{t+1}| \ge 0$, substituting $\mathbf{p}'$ into $\eta$ and using $\eta(\mathbf{q}_t) = \eta^*_t$ yields
\begin{equation}
\min_{\mathcal{D}_{t+1}} \eta = \eta(\mathbf{p}')
= \eta^*_t - \Delta\omega - |\kappa|\, \Delta v,
\label{eq:probe_deviation}
\end{equation}
where the linear-velocity term equals $-|\kappa|\, \Delta v$ in both cases: for $\kappa \ge 0$, $v$ increases by $\Delta v$ and the term is $-\kappa\, \Delta v$, and for $\kappa < 0$, $v$ decreases by $\Delta v$ and the term is $+\kappa\, \Delta v = -|\kappa|\, \Delta v$. Therefore, if $\eta(\mathbf{p}') \le 0$, the intersection is recovered at step $t+1$ and DWPP selects a command on the line; otherwise $\eta > 0$ on $\mathcal{D}_{t+1}$, the intersection is not recovered, $\eta^*_{t+1} = \eta(\mathbf{p}') = \eta^*_t - (\Delta\omega + |\kappa|\, \Delta v)$, and $\mathbf{p}'$ is exactly the command that DWPP selects at step $t+1$.

Next, the decrease $\Delta\omega + |\kappa|\, \Delta v$ in the latter case is bounded from below by $\delta$. Each displacement equals its full step, $\Delta\omega = \alpha^{\mathrm{dcc}}_{\max} \Delta t$ and $\Delta v = a^{\mathrm{acc}}_{\max} \Delta t$ (resp.\ $a^{\mathrm{dcc}}_{\max} \Delta t$), unless the corresponding update is clipped, i.e., stopped at a bound of $\mathcal{G}$. It is shown by contradiction that the two updates cannot both be clipped. A clipped angular update gives $\omega' = \omega_{\min}$, and a clipped linear update gives $v' = \min(v_{\max}, v_{\mathrm{reg}})$ (resp.\ $v' = \max(v_{\min}, 0)$); if both held, $\mathbf{p}'$ would be the vertex minimizing $\eta$ over $\mathcal{G}$, so (A3) would give $\eta(\mathbf{p}') = \min_{\mathcal{G}} \eta \le 0$, contradicting $\eta(\mathbf{p}') > 0$. Hence at least one update realizes its full step, and the decrease is at least $\alpha^{\mathrm{dcc}}_{\max} \Delta t$ or $|\kappa|\, a^{\mathrm{acc}}_{\max} \Delta t$ (resp.\ $|\kappa|\, a^{\mathrm{dcc}}_{\max} \Delta t$); the symmetric case $\eta < 0$ involves $\alpha^{\mathrm{acc}}_{\max}$ and $a^{\mathrm{dcc/acc}}_{\max}$ instead. For $\kappa = 0$, the linear velocity does not change $\eta$, so the angular update itself must be unclipped; indeed, a clipped one would give $\eta(\mathbf{p}') = \omega_{\min} \le 0$, the same contradiction. The decrease is then at least $\alpha^{\mathrm{dcc}}_{\max} \Delta t$ ($\alpha^{\mathrm{acc}}_{\max} \Delta t$ for $\eta < 0$). The rate $\delta$ of Eq.~\eqref{eq:recovery_rate} is the minimum of these full-step contributions and therefore bounds the decrease from below in every case.

Combining the two parts, at every step of the non-intersection condition either the intersection is recovered or $\eta^*_{t+1} \le \eta^*_t - \delta$; since the positive quantity $\eta^*_t$ can decrease by $\delta$ at most $\lceil \eta^*_{t_0} / \delta \rceil$ times, the intersection is recovered within $\lceil \eta^*_{t_0} / \delta \rceil$ steps. \qed
\end{pf}

Note that assumption (A1) treats $\kappa$ and $v_{\mathrm{reg}}$ as constant while the non-intersection transient elapses, whereas both are recomputed at every control step. However, the proof tolerates this variation quantitatively. Let $\Delta\kappa$ and $\Delta v_{\mathrm{reg}}$ bound their per-step changes, and let $\underline{\delta}$ be the smallest recovery rate of Eq.~\eqref{eq:recovery_rate} during the transient. The deviation of the selected command $\mathbf{q}_t$ moves by at most $\Delta\kappa\, v_{\max} + |\kappa|\, \Delta v_{\mathrm{reg}}$ per step, so $\eta^*_t$ still decreases by at least $\underline{\delta}$ minus this amount, and the recovery occurs within $\lceil \eta^*_{t_0} / \varepsilon \rceil$ steps whenever the margin stays above a positive constant $\varepsilon$. On piecewise-constant-curvature paths, a jump of $\kappa$ at a segment junction initializes the transient, and during the recovery $\Delta\kappa$ and $\Delta v_{\mathrm{reg}}$ remain small for a sufficiently long lookahead distance, in line with the classical results on the lookahead distance required for stable tracking \cite{coulter1992implementation, ollero1995stability}.
\end{revblock}

\begin{revblock}
\section{An Estimate of the Smallest Usable Lookahead Distance under Localization Noise}
\label{app:noise_bound}
\setcounter{equation}{0}
\renewcommand{\theequation}{B.\arabic{equation}}

This appendix quantifies the mechanism of Section~\ref{sec:discussion_noise} and estimates the smallest usable lookahead distance under localization noise. Consider a straight segment tracked under the noise model of Section~\ref{sec:sim_noise}. In this model, the pose fed to the controller carries independent zero-mean position and heading noise with standard deviations $\sigma_{xy}$ and $\sigma_{\theta}$ at every control step. A position error shifts the lookahead point laterally by its lateral component, and a heading error $\delta\theta$ shifts it by approximately $l\,\delta\theta$ at the lookahead distance $l$, so the lateral perturbation of the lookahead point has the standard deviation
\begin{equation}
  \sigma_{\mathrm{eff}} = \sqrt{\sigma_{xy}^2 + l^2 \sigma_{\theta}^2}.
  \label{eq:sigma_eff}
\end{equation}
Write this perturbation as $\varepsilon$, which is zero-mean Gaussian under the noise model. On the straight segment the nominal curvature command is zero, and the perturbation turns the curvature command of Eq.~\eqref{eq:pp_curvature} into $\kappa = 2\sin\varphi/l \approx 2\varepsilon/l^2$, since $\sin\varphi \approx \varepsilon/l$ for the small angles involved. The curvature command therefore fluctuates around zero, and its magnitude has the median $2 \times 0.674\,\sigma_{\mathrm{eff}}/l^2 \approx 1.35\,\sigma_{\mathrm{eff}}/l^2$, where $0.674$ is the median magnitude of the standard normal distribution. Because the sign of this fluctuation flips from step to step, the angular command cannot accumulate beyond the per-step change $\alpha^{\mathrm{acc}}_{\max}\Delta t$. The selected command still satisfies $\omega = \kappa v$, so dividing this angular-command bound by the median curvature magnitude gives the sustainable linear velocity
\begin{equation}
  v(l) \approx \frac{\alpha^{\mathrm{acc}}_{\max}\Delta t\, l^2}{1.35\,\sigma_{\mathrm{eff}}}.
  \label{eq:noise_vcap}
\end{equation}
Requiring $v(l)$ to exceed the mean speed $v_{\mathrm{req}}$ that the task demands, and using $\sigma_{\mathrm{eff}} \approx \sigma_{xy}$ because the position term dominates in practical settings, gives the closed-form estimate
\begin{equation}
  l_{\min} \approx \sqrt{\frac{1.35\, v_{\mathrm{req}}\, \sigma_{xy}}{\alpha^{\mathrm{acc}}_{\max}\Delta t}},
  \label{eq:l_min}
\end{equation}
which is obtained directly from the noise level of the pose estimate, the angular acceleration limit, and the control period. For the noise level and the timeout of Sections~\ref{sec:sim_noise} and \ref{sec:sim_lookahead}, with $v_{\mathrm{req}} = 9.0~\mathrm{m}/120~\mathrm{s}$, the bound evaluates to about $0.39$~m, consistent with the result of Section~\ref{sec:sim_lookahead} that $L = 0.4$~m reached the goal whereas $L = 0.3$~m did not. Note that the estimate neglects the intervals in which the sampled noise stays small and the velocity builds up, visible as the brief speed bursts in Fig.~\ref{fig:sim_noise}, and that it assumes the independent redraw of the noise at every control step, whereas the corrections of the real localization module are slower and temporally correlated.
\end{revblock}

\bibliographystyle{elsarticle-num} 
 \bibliography{bibliography}

%% else use the following coding to input the bibitems directly in the
%% TeX file.

%% Refer following link for more details about bibliography and citations.
%% https://en.wikibooks.org/wiki/LaTeX/Bibliography_Management

\end{document}